%% file: main.tex
\documentclass[10pt,twocolumn,letterpaper]{article}

\usepackage[pagenumbers]{cvpr} 


\input{sec/defn}

\input{plots/utils}
\usepackage{nicefrac}

\definecolor{cvprblue}{rgb}{0.21,0.49,0.74}
\usepackage[pagebackref,breaklinks,colorlinks,allcolors=cvprblue]{hyperref}

\title{Retrieve and Segment: Are a Few Examples Enough \\ to Bridge the Supervision Gap in Open-Vocabulary Segmentation?}

\author{
Tilemachos Aravanis\textsuperscript{1} \quad
Vladan Stojnić\textsuperscript{1} \quad
Bill Psomas\textsuperscript{1} \quad
Nikos Komodakis\textsuperscript{2 3 4} \quad
Giorgos Tolias\textsuperscript{1} \\
\vspace{2pt}
\textsuperscript{1}VRG, FEE, Czech Technical University in Prague \\
\textsuperscript{2}University of Crete
\textsuperscript{3}Archimedes, Athena RC
\textsuperscript{4}IACM-FORTH
\vspace{2pt}
}

\begin{document}
\input{sec/teaser}
\maketitle
\input{sec/abstract}

\input{sec/intro}
\input{sec/related}
\input{sec/method}
\input{sec/experiments}
\input{sec/conclusion}

\input{sec/acknowledgements}
{
    \small
    \bibliographystyle{ieeenat_fullname}
    \bibliography{main}
}

\input{sec/supp}

\end{document}

%% file: sec/defn.tex
\newcommand{\ours}{\textsc{RnS}\xspace}
\newcommand{\knnclip}{\ensuremath{\text{kNN-}}\textsc{CLIP}\xspace}
\newcommand{\freeda}{\textsc{FreeDA}\xspace}

\definecolor{personalgreen}{RGB}{0, 128, 1}
\definecolor{personalred}{RGB}{128, 0, 0}

\definecolor{OrangeFrame}{HTML}{D79B00}
\definecolor{OliveGreen}{RGB}{85, 107, 47}
\definecolor{BrickRed}{RGB}{156, 32, 32}
\definecolor{LightSteelBlue1}{RGB}{202, 225, 255}

\definecolor{appleblue}{RGB}{80,122,255}  
\definecolor{appleteal}{RGB}{85,163,152}  
\definecolor{applepurple}{RGB}{138,107,221}  
\definecolor{applemustard}{RGB}{234,179,77}  
\definecolor{appleorange}{RGB}{255,127,80}  
\definecolor{applebrown}{RGB}{162,121,92}

\definecolor{appletealD1}{RGB}{68,130,122}
\definecolor{appletealL1}{RGB}{119,181,173} 
\definecolor{appletealL2}{RGB}{153,200,193} 
\definecolor{appletealL3}{RGB}{187,218,214} 
\definecolor{appletealL35}{RGB}{204,228,224}
\definecolor{appletealL4}{RGB}{221,237,234} 
\definecolor{appletealD2}{RGB}{51,98,91} 
\definecolor{appletealD4}{RGB}{17,33,30}
\definecolor{appletealD3}{RGB}{34,65,61}

\definecolor{appleorangeL2}{RGB}{255,178,150} 
\definecolor{appleorangeD2}{RGB}{153,76,48}   

\definecolor{appleorangeL1}{RGB}{255,153,115} 

\definecolor{appleorangeD1}{RGB}{204,102,64}  

\definecolor{applepink}{RGB}{217,127,174}
\definecolor{applebrownlight}{RGB}{193,153,121} 
\definecolor{applebrowndark}{RGB}{118,88,67}

\definecolor{applegreen}{RGB}{52,199,89}
\definecolor{appledeepgreen}{RGB}{34,135,74}

\definecolor{appleyellow}{RGB}{255,204,0}

\definecolor{applegreen-dark}{RGB}{34,135,74}
\definecolor{applegreen-light}{RGB}{190,237,204}
\definecolor{appleblue-dark}{RGB}{58,103,214}
\definecolor{appleblue-light}{RGB}{202,219,255}

\definecolor{teasergreen}{HTML}{25A825}
\definecolor{teaserorange}{HTML}{F0C131}
\definecolor{teaserblue}{HTML}{3D85C6}
\makeatletter
\newcommand\minuscule{\@setfontsize\minuscule{6}{6.2}} 
\makeatother

\definecolor{imagequerycolor}{RGB}{255,140,0}

\definecolor{textquerycolor}{rgb}{0.21,0.49,0.74}


%% file: plots/utils.tex
\usepackage{tikz}
\usepackage[table]{xcolor}
\usepackage{pgfplots}
\pgfplotsset{compat=1.9}
\usepackage{xparse}
\usepackage{pgfplotstable}
\usepackage{xstring}
\usepackage{makecell}
\usepackage[export]{adjustbox}
\usepackage{multirow}
\usepackage{booktabs,pifont}
\usetikzlibrary{plotmarks}
\usepackage{booktabs}
\usepackage{arydshln}

\newcommand{\xmark}{\textcolor{red}{\ding{55}}}

\definecolor{oursgreen}{HTML}{1F7A1F}      
\definecolor{ourslight}{HTML}{7FC97F}      
\definecolor{kclipblue}{HTML}{1F77B4}      
\definecolor{kcliplight}{HTML}{9ED0E6}     
\definecolor{freedaorange}{HTML}{FFA600}   
\definecolor{freedayellow}{HTML}{FFD84D}   

\definecolor{ourscyan}{HTML}{5DD6E6}
\definecolor{offlinepink}{HTML}{E98EAE}
\definecolor{offlineorange}{HTML}{F3A532} 
\definecolor{offlinepurple}{HTML}{6F2DBD} 

\NewDocumentEnvironment{ShotMiouAxis}{ O{} }{
  \begin{tikzpicture}
  \begin{axis}[
    xtick={1,2,3,5,10,20},
    xmin=0.7, xmax=20.3,
    ymin=26, ymax=51,
    width=\linewidth, height=6.5cm,
    grid=both,
    grid style={dotted,gray!80},
    xlabel={Support images per class},
    ylabel={mIoU (\%)},
    legend pos=south east, 
    legend style={
      draw=black!60,
      fill=white,
      fill opacity=0.8,
      rounded corners=5pt,
      font=\tiny,
      row sep=-2pt,
    },
    legend cell align=left,
    xlabel style={font=\scriptsize},
    ylabel style={font=\scriptsize},
    x tick label style={font=\scriptsize},
    y tick label style={font=\scriptsize},
    #1 
  ]
}{
    \end{axis}
  \end{tikzpicture}
}

\NewDocumentEnvironment{ShotMiouDinoAxis}{ O{} }{
  \begin{tikzpicture}
  \begin{axis}[
    xtick={1,2,3,5,10,20},
    xmin=0.7, xmax=20.3,
    ymin=26, ymax=51,
    width=\linewidth, height=6.5cm,
    grid=both,
    grid style={dotted,gray!80},
    xlabel={Support images per class},
    ylabel={mIoU (\%)},
    legend pos=south east, 
    legend style={
      draw=black!60,
      fill=white,
      rounded corners=5pt,
      font=\tiny
    },
    legend cell align=left,
    xlabel style={font=\scriptsize},
    ylabel style={font=\scriptsize},
    x tick label style={font=\scriptsize},
    y tick label style={font=\scriptsize},
    #1 
  ]
}{
    \end{axis}
  \end{tikzpicture}
}

\NewDocumentCommand{\AddShotMiouCurve}{ m m m O{} }{%
  \def\templegendname{#1}%
  \ifx\templegendname\empty
    \addplot+[
      color=#2,
      line width=1.2pt,
      mark=*,
      mark size=2.0pt,
      mark options={solid, fill=#2, draw=black!80, line width=0.4pt},
      solid,
      forget plot,
      #4
    ] coordinates {#3};
  \else
    \addplot+[
      color=#2,
      line width=1.2pt,
      mark=*,
      mark size=2.0pt,
      mark options={solid, fill=#2, draw=black!80, line width=0.4pt},
      solid,
      #4
    ] coordinates {#3};
    \addlegendentry{#1}%
  \fi%
}

\NewDocumentCommand{\AddShotMiouDashedCurve}{ m m m O{} }{%
  \def\templegendname{#1}%
  \ifx\templegendname\empty
    \addplot+[
      color=#2,
      line width=1.2pt,
      mark=*,
      mark size=2.0pt,
      mark options={solid, fill=#2, draw=black!80, line width=0.4pt},
      dashed,
      forget plot,
      #4
    ] coordinates {#3};
  \else
    \addplot+[
      color=#2,
      line width=1.2pt,
      mark=*,
      mark size=2.0pt,
      mark options={solid, fill=#2, draw=black!80, line width=0.4pt},
      dashed,
      #4
    ] coordinates {#3};
    \addlegendentry{#1}%
  \fi%
}

\NewDocumentCommand{\AddZeroShotLine}{ m m m }{%
  \def\templegendname{#1}%
  \ifx\templegendname\empty
    \addplot+[
      domain=\pgfkeysvalueof{/pgfplots/xmin}:\pgfkeysvalueof{/pgfplots/xmax},
      samples=2,
      line width=1.2pt,
      color=#2,
      mark=none,
      mark size=2.0pt,
      mark options={solid, fill=#2, draw=black!80, line width=0.4pt},
      solid,
      forget plot
    ] {#3};
  \else
    \addplot+[
      domain=\pgfkeysvalueof{/pgfplots/xmin}:\pgfkeysvalueof{/pgfplots/xmax},
      samples=2,
      line width=1.2pt,
      color=#2,
      mark=none,
      mark size=2.0pt,
      mark options={solid, fill=#2, draw=black!80, line width=0.4pt},
      solid
    ] {#3};
    \addlegendentry{#1}%
  \fi%
}

\NewDocumentCommand{\AddZeroShotDashedLine}{ m m m }{%
  \def\templegendname{#1}%
  \ifx\templegendname\empty
    \addplot+[
      domain=\pgfkeysvalueof{/pgfplots/xmin}:\pgfkeysvalueof{/pgfplots/xmax},
      samples=2,
      line width=1.2pt,
      color=#2,
      mark=none,
      mark size=2.0pt,
      mark options={solid, fill=#2, draw=black!80, line width=0.4pt},
      dashed,
      forget plot
    ] {#3};
  \else
    \addplot+[
      domain=\pgfkeysvalueof{/pgfplots/xmin}:\pgfkeysvalueof{/pgfplots/xmax},
      samples=2,
      line width=1.2pt,
      color=#2,
      mark=none,
      mark size=2.0pt,
      mark options={solid, fill=#2, draw=black!80, line width=0.4pt},
      dashed
    ] {#3};
    \addlegendentry{#1}%
  \fi%
}

\NewDocumentEnvironment{DropFracAxis}{ O{} }{
  \begin{tikzpicture}
    \begin{axis}[
      xlabel={Fraction of classes w/o visual support},
      ylabel={mIoU (\%)},
      grid=both,
      grid style={dotted,gray!80},
      legend pos=south west,
      legend cell align=left,
      legend style={font=\tiny, row sep=-2pt},
      legend image post style={scale=0.9},
      mark size=2.0pt,
      legend style={
      draw=black!60,
      fill=white,
      fill opacity=0.8,
      rounded corners=5pt
      },
      xlabel style={font=\scriptsize},
      ylabel style={font=\scriptsize},
      x tick label style={font=\scriptsize},
      y tick label style={font=\scriptsize},
      xticklabel style={/pgf/number format/fixed},
      yticklabel style={/pgf/number format/fixed},
      #1
    ]
}{
    \end{axis}
  \end{tikzpicture}
}

\NewDocumentCommand{\AddMiouDropCurve}{ m m m O{} }{%
  \def\templegendname{#1}%
  \ifx\templegendname\empty
    \addplot+[
      color=#2,
      line width=1.2pt,
      mark=*,
      mark size=2.0pt,
      mark options={solid, fill=#2, draw=black!80, line width=0.4pt},
      solid,
      forget plot,
      #4
    ] coordinates {#3};
  \else
    \addplot+[
      color=#2,
      line width=1.2pt,
      mark=*,
      mark size=2.0pt,
      mark options={solid, fill=#2, draw=black!80, line width=0.4pt},
      solid,
      #4
    ] coordinates {#3};
    \addlegendentry{#1}%
  \fi%
}

\NewDocumentCommand{\AddMiouBaseline}{ m m m }{%
  \AddZeroShotLine{#1}{#2}{#3}%
}

\NewDocumentEnvironment{DropFracTextAxis}{ O{} }{
  \begin{tikzpicture}
    \begin{axis}[
      xlabel={Fraction of classes w/o textual support},
      ylabel={mIoU (\%)},
      grid=both,
      grid style={dotted,gray!80},
      legend pos=south west,
      legend cell align=left,
      legend style={font=\tiny, row sep=-2pt},
      legend image post style={scale=0.9},
      mark size=2.0pt,
      legend style={
      draw=black!60,
      fill=white,
      fill opacity=0.8,
      rounded corners=5pt
      },
      xlabel style={font=\scriptsize},
      ylabel style={font=\scriptsize},
      x tick label style={font=\scriptsize},
      y tick label style={font=\scriptsize},
      xticklabel style={/pgf/number format/fixed},
      yticklabel style={/pgf/number format/fixed},
      #1
    ]
}{
    \end{axis}
  \end{tikzpicture}
}

\NewDocumentCommand{\AddMiouBaselineWithMarks}{ m m m O{1,2,3,5,10,20} }{%
  \AddZeroShotLine{#1}{#2}{#3}%
  \addplot+[
    only marks,
    samples at={#4},
    mark=+, mark size=2.4pt,              
    mark options={
      rotate=45, line cap=round, line join=round,
      preaction={draw=black, line width=1.1pt}, 
      draw=#2, line width=0.7pt
    },
    forget plot
  ] {#3};
}

%% file: sec/teaser.tex
\newcommand{\teaserfig}[1]{\includegraphics[trim={0cm 0cm 0cm 0cm},width=1.0\textwidth,valign=c]{#1}}
\makeatletter
\apptocmd\@maketitle{{\teaser{}}}{}{}
\makeatother
\newcommand{\teaser}{%
\vspace{-20pt}
\centering
\teaserfig{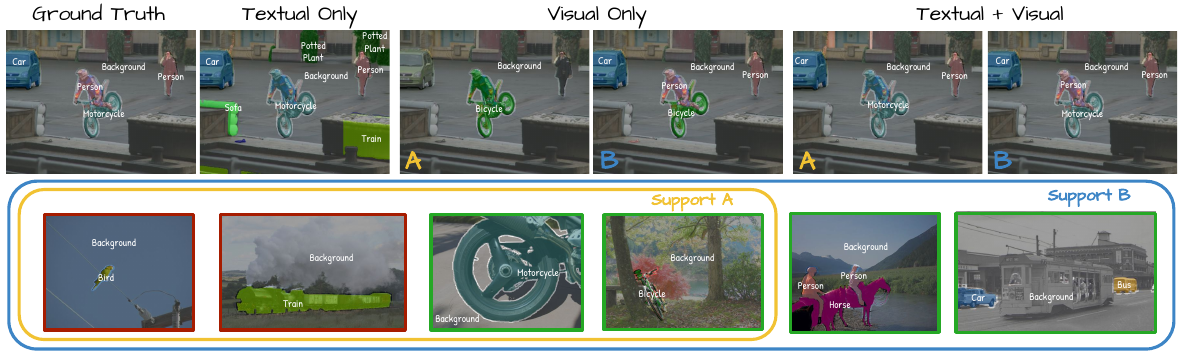}\\
\vspace{-10pt}
\captionof{figure}{\textbf{Open-vocabulary segmentation (OVS) results.} We compare three settings: (i) \emph{textual-only} support (zero-shot OVS), (ii) a simplified version of \texttt{\ours} using \emph{visual-only} support, and (iii) the full \texttt{\ours} combining \emph{textual+visual} support. 
Textual support: class name or description. Visual support: a small set of pixel-annotated images for some classes. Initially, visual support includes images in \textcolor{teaserorange}{$A$} and is later expanded to \textcolor{teaserblue}{$B$}, with $A \subseteq B$. Text-only support often yields ambiguous predictions (rider as motorcycle, background hallucinations). Visual-only support struggles when some classes lack support (person, car) and can confuse similar objects (motorcycle, bicycle) even when all classes have support. 
By \underline{\emph{retrieving}} information from images \textcolor{teasergreen}{relevant to the test image} and combining with the textual support, \texttt{\ours} is robust under missing visual support for some classes and achieves accurate \underline{\emph{segmentation}}.
\label{fig:teaser}}
\par\vspace{12pt}
}

%% file: sec/abstract.tex
\begin{abstract}
Open-vocabulary segmentation (OVS) extends the zero-shot recognition capabilities of vision–language models (VLMs) to pixel-level prediction, enabling segmentation of arbitrary categories specified by text prompts. Despite recent progress, OVS lags behind fully supervised approaches due to two challenges: the coarse image-level supervision used to train VLMs and the semantic ambiguity of natural language. We address these limitations by introducing a few-shot setting that augments textual prompts with a support set of pixel-annotated images. Building on this, we propose a retrieval-augmented test-time adapter that learns a lightweight, per-image classifier by fusing textual and visual support features. Unlike prior methods relying on late, hand-crafted fusion, our approach performs learned, per-query fusion, achieving stronger synergy between modalities. The method supports continually expanding support sets, and applies to fine-grained tasks such as personalized segmentation. 
Experiments show that we significantly narrow the gap between zero-shot and supervised segmentation while preserving open-vocabulary ability. \href{https://github.com/TilemahosAravanis/Retrieve-and-Segment}{\textcolor{purple}{\emph{Code}}} 
\end{abstract}

%% file: sec/intro.tex
\section{Introduction}
\label{sec:intro}
Semantic segmentation traditionally relies on fully supervised models trained on dense pixel-level annotations within a fixed set of categories~\cite{long2015fully, zhao2017pyramid, chen2018encoder}. While this approach yields accurate, well-localized masks, it does not scale; collecting pixel-level annotations is costly, and models cannot recognize categories unseen during training.

Open-vocabulary segmentation (OVS) builds on the zero-shot recognition capabilities of contrastively trained vision–language models (VLMs)~\cite{radford2021learning,jia2021scaling,zhai2023sigmoid}. Trained on large image–text datasets~\cite{changpinyo2021conceptual, schuhmann2022laion}, VLMs learn a shared embedding space where images and text are directly comparable, enabling recognition of arbitrary categories specified at test time via text prompts or class names. Extending this ability from image-level to pixel-level prediction has driven rapid progress in OVS~\cite{clipdinoiser,lan2024proxyclip,clearclip,stojnic2025lposs}. However, a substantial gap remains to fully supervised models~\cite{chen2018encoder}; recent improvements show signs of plateauing~\cite{vsaric2025holds}.

Two key challenges underlie this gap: (i) the mismatch between image-level supervision used to train VLMs and the fine-grained predictions required for segmentation, and (ii) the semantic ambiguity of natural language supervision. While language enables open-vocabulary recognition, it often lacks the precision needed for pixel-level tasks.

We address these challenges by introducing a few-shot setting that supplements the textual support of class names with a small support set of visual examples, \ie images with pixel-level annotation. We aim to \emph{bridge the gap} between zero-shot OVS and supervised segmentation, while preserving the open-vocabulary predictions. We propose a retrieval-augmented test-time adapter that  trains a lightweight classifier per test image. Inspired by retrieval-based methods~\cite{khandelwalgeneralization,wumemorizing}, our approach retrieves relevant visual support examples and fuses them with textual support to construct test-time training data. Unlike previous work~\cite{gui2024knnclip,freeda} relying on hand-crafted fusion, our method performs a learned per-image fusion of textual and visual prototypes, enabling strong synergy between modalities, as shown in~\autoref{fig:teaser}. Importantly, we store only a compact set of visual prototypes from the support images, keeping the memory footprint minimal, while our test-time training requires less than a second on an NVIDIA $A100$ GPU\footnote{See supplementary material for runtime details.}.

This design is enabled by the strong, generalizable features of modern VLMs trained at large-scale~\cite{schuhmann2022laion,changpinyo2021conceptual}. These rich features allow us to avoid retraining the backbone and instead steer predictions with a lightweight classifier trained on only a few visual examples. By leveraging these robust embeddings, our approach achieves efficient adaptation while preserving the open-vocabulary nature of the task.

Our method, called \emph{\underline{R}etrieve a\underline{n}d \underline{S}egment (\ours)}, can handle diverse real world settings, from textual and visual support for all classes, to partial support for some classes, where either a textual description is not straightforward to obtain or visual examples are not yet available. 
Moreover, it is compatible with a dynamic continually changing setting where new visual examples can be added to the support set at any time, without sacrificing the open-vocabulary nature. Due to its simplicity and its dynamic adaptability, our approach is easily applicable to segmentation of particular objects, \ie the so called personalized segmentation~\cite{zhang2024personalize,sundaram2025personalized}.
In summary, our contributions are:
\begin{itemize}
\item We investigate multiple few-shot settings for open-vocabulary segmentation, enriching textual prompts with pixel-annotated visual examples.
\item We introduce \ours, a retrieval-augmented, test-time adapter that learns to fuse textual and visual support more effectively than prior approaches.
\item \ours significantly reduces the performance gap between zero-shot and fully supervised segmentation, while maintaining open-vocabulary generalization.
\item \ours supports dynamic support expansion in continually evolving environments, and adapts seamlessly to fine-grained tasks such as personalized segmentation.
\end{itemize}

%% file: sec/related.tex
\section{Related work}
\label{sec:related}

\textbf{Open-vocabulary segmentation} 
is performed by leveraging vision–language models (VLMs) that align images and text in a shared space~\cite{radford2021learning, zhai2023sigmoid, jia2021scaling}. The fixed classifier is replaced by a text encoder and matches patch features to text features, but vanilla VLMs pre-trained with image-level supervision struggle with dense localization~\cite{bousselham2024grounding,maskclip}. There are three lines of work: (i) \emph{Training VLMs for segmentation}: either weakly supervised with masks without labels~\cite{ding2022decoupling, ghiasi2022scaling} or image–caption pairs~\cite{cha2023learning,luo2023segclip,mukhoti2023open,ranasinghe2023perceptual,xu2022groupvit,xu2023learning}, or fully supervised with pixel annotations~\cite{liang2023open,cho2024catseg,yu2023convolutions,jiao2024collaborative,li2025maskadapter,xie2024sed,zhao2025dpseg,li2025images}. However, this approach hinders the open-vocabulary performance outside of the training domain~\cite{stojnic2025lposs, cho2024catseg}, as shown in datasets like MESS~\cite{blumenstiel2023mess}. (ii) \emph{Training-free VLM tweaks}: modified inference processes to boost spatial sensitivity, \eg removing the final attention layer~\cite{maskclip} or residual connections and feedforward networks~\cite{clearclip}, with further variants~\cite{clearclip,sclip,bousselham2024grounding}. (iii) \emph{VLM+VM hybrids}: combining VLM semantics with the localization of vision models (VMs)~\cite{wysoczanska2024clip,clipdinoiser,lan2024proxyclip,kang2024defense,shi2025harnessing,kim2025distilling,stojnic2025lposs,zhang2025corrclip}. Methods typically localize objects with DINO~\cite{caron2021emerging,oquab2024dinov2} or SAM~\cite{kirillov2023segment,ravi2024sam2segmentimages}, then classify the localized regions in an open-vocabulary manner using VLMs. Despite progress, OVS still trails task-specific, fully supervised models. 


\textbf{Few-shot segmentation} learns on base classes and adapts to novel classes from few (support) labeled examples. Meta-learning and prototypical methods perform episodic $1-$ or $N-$way training, create per-class prototypes and classify via similarity~\cite{shaban2017one,wang2019panet,siam2019amp,tian2020prior,min2021hypercorrelation}, typically assuming a closed world.
A recent generalized setting evaluates on both base and novel classes~\cite{tian2022generalized,hajimiri2023strong,hossain2024visual,liu2023learning}, but still requires abundant pixel-level annotations for base classes and does not leverage VLMs. 
Close to our setting, a concurrent work Power-of-One~\cite{hossain2025power} introduces one-shot per class fine-tuning of text embeddings and specific backbone layers. They require access to raw images, while we operate on pre-extracted features, and fine-tune internal VLM~\cite{li2022blip} layers for each new class set, which is not as light-weight as our test-time adapter. Beyond VLMs, CAT-SAM~\cite{xiao2024cat} explores few-shot adaptation of SAM~\cite{kirillov2023segment} via conditional tuning with lightweight adapters, but does not combine visual and textual support. 
COSINE~\cite{liu2025unified} unifies open-vocabulary (text-prompted) and in-context (image-prompted) segmentation by training a decoder on top of frozen foundation models to handle multi-modal prompts. For open vocabulary semantic segmentation with many categories the model is only evaluated with unimodal prompts though.

\looseness=-1


\textbf{Retrieval augmentation in segmentation.} Retrieval-augmented models for prediction and generation have shown strong performance by dynamically expanding their knowledge base~\cite{ram2023incontextretrievalaugmentedlanguagemodels, yuan2024ragdriver, iscen2024retrievalenhancedcontrastivevisiontextmodels, liu2023learningcustomizedvisualmodels}.
Following this paradigm in semantic segmentation, recent work enhances in-context scene understanding of vision encoders \cite{balazevic2023towardsincontextsceneunderstanding, pariza2025nearfar, sirkogalouchenko2025dipunsuperviseddenseincontext}.
%
\freeda \cite{freeda} is conceptually related to \ours, but relies on generated visual examples. At test time, textual class features are expanded into visual counterparts via retrieval, forming a non-parametric visual classifier, which is then combined with a standard zero-shot textual classifier. For fair comparison, we adapt this method to use real support images instead of generated ones.
Closely related to our work is \knnclip~\cite{gui2024knnclip}, which enhances open-vocabulary semantic segmentation by leveraging a memory-efficient support set of class vectors derived from pixel-annotated images. At test time, it assigns labels to image regions based on the similarity to their $k$ nearest neighbors in the support set. This approach outperforms continual learning baselines, demonstrating that a dynamic support set can incorporate visual examples of new classes without forgetting previously learned ones. However, while \knnclip claims to expand the model’s vocabulary to arbitrary class sets, its performance remains limited to classes for which annotated examples are available.


\textbf{Test-time adaptation (TTA)} for VLMs has grown rapidly, but many methods operate in batch/transductive or streaming modes~\cite{karmanov2024efficient,fan2025test,lee2025ratta} and carry unrealistic assumptions, such as class-complete batches or i.i.d. streams, which compromises zero-shot robustness~\cite{zanella2024boosting}. In contrast, single-image TTA adapts per sample: TPT~\cite{shu2022test} and ZERO~\cite{farina2024frustratingly} perform optimization or predictions over augmentations. For segmentation, single-image TTA is explored with self-supervised objectives \cite{janouskova2024single}, and OVS-specific TTA is just emerging, proposing adaptation layers for VLM-based segmenters at test time \cite{noori2025test}.

%% file: sec/method.tex
\section{Method}
\label{sec:method}

\newcommand{\argmax}{\operatorname{argmax}}
\newcommand{\I}{\mathcal{I}}
\newcommand{\RR}{\mathbb{R}}
\newcommand{\CC}{\mathcal{C}}
\newcommand{\Ic}{\mathcal{I}_c}
\newcommand{\Ci}{\mathcal{C}_i}
\newcommand{\kNN}{\operatorname{k\text{-}NN}}
\newcommand{\KL}{\operatorname{KL}}
\newcommand{\CE}{\operatorname{CE}}

\begin{figure*}[t]
\vspace{-10pt}
  \centering
  \includegraphics[width=\linewidth]{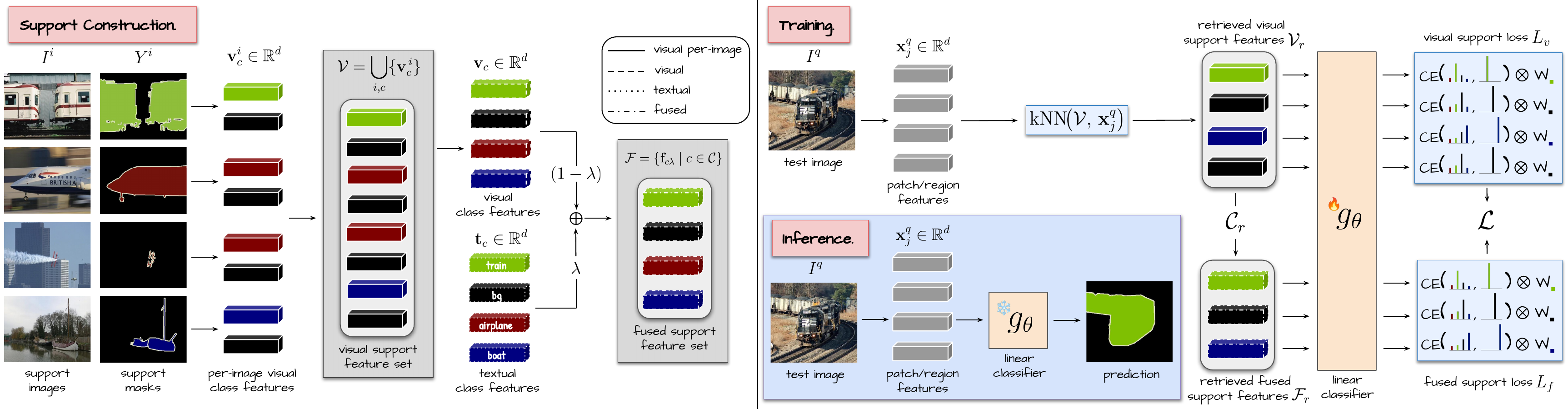}
  \vspace{-20pt}
  \caption{\textbf{Overview of \ours when \emph{full textual and visual support} is available.} Having access to a set of pixel-level annotated images, \emph{per-image visual class features} $\mathbf{v}^{i}_{c}$ are extracted. These features are then aggregated by class to form \emph{visual class features} $\mathbf{v}_{c}$, which are combined with \emph{textual class features} $\mathbf{t}_c$, through a mixing coefficient $\lambda$, to produce \emph{fused class features} $\mathbf{f}_{c\lambda}$. During test-time training, a test-image-relevant subset of \emph{visual support features} and \emph{fused class features}, along with their class labels, are used to train a lightweight linear classifier $g_{\theta}$ using cross-entropy loss. Each training sample is weighted with a class relevance weight $w_{c}$ (\eg $w_{\scalebox{0.4}{$\blacksquare$}}$ for \texttt{bg}). At inference, this classifier, trained per test image, is applied to patch-level features $\mathbf{x}_j^q$ to generate segmentation predictions. When SAM is available, patch-level features are replaced by region-level features $\mathbf{x}^{\,q}_r$ for improved accuracy.
  \label{fig:support-construction}
  \vspace{-10pt}}
\end{figure*}

\subsection{Task formulation}
Given image $I \in \RR^{H\times W \times 3}$ and set $\CC$ of $C$ semantic classes, we aim to assign each pixel of $I$ to one of the classes.
The class set is arbitrary and defined at test time. Therefore, the task is referred to as \emph{open-vocabulary segmentation (OVS)}.

Each class is specified either by a textual example, \eg a class name, or by a small set of visual examples in the form of raw images with pixel-level annotations, like in a few-shot setting.
We refer to these as the \emph{textual support set} and \emph{visual support set}, respectively.
Typically, a textual example is available for every class.
However, visual examples may be missing initially, but their number and diversity can increase over inference time, reflecting a continually expanding open-world scenario.
In rare cases, textual examples may also be absent, \eg for novel categories with unknown names or in specialized domains such as medical imaging or remote sensing, where naming is non-trivial.

We consider the following settings: (i) \emph{full-support}: every class in $\CC$ has a class name and at least one annotated support image. (ii) \emph{partial-visual-support}: some classes lack visual examples, but all have class names. (iii) \emph{partial-textual-support}: some classes lack class names, but all have visual examples.
(iv) \emph{only-textual-support}: class names are available for all classes in $\CC$, but no visual support images are provided, \ie the commonly studied \emph{zero-shot} segmentation setting.

In contrast to zero-shot segmentation, all other settings have visual support examples available, and our goal is to leverage them to improve segmentation, while preserving the ability to operate with an open-vocabulary.

We denote a support image by $I^{i}$, a test (query) image by $I^{q}$, while we drop the superscript when referring to images in general and we adopt consistent notation for other variables as introduced in the following sections.

\subsection{Zero-shot segmentation with VLMs}
\label{zero-shot}
Image $I \in \RR^{H \times W \times 3}$, processed by the vision encoder of a VLM, is mapped to a \emph{patch-level feature matrix} $X \in \mathbb{R}^{n \times d}$. Here $n = h \times w$ corresponds to the flattened $h \times w$ patch grid produced by a vision transformer (ViT), and $d$ denotes the feature dimension.
The \emph{patch feature} at position $j \in 1, \ldots, n$ is denoted by $\mathbf{x}_{j} \in \RR^{d}$.

Each class name $c \in \CC$, processed by the VLM's text encoder, is mapped to a \emph{textual class feature} $\mathbf{t}_c \in \RR^{d}$. 
Both textual and visual features are normalized to unit length.

The patch-level prediction map is denoted by $\hat{P} \in [0,1]^{n \times C}$ with elements $\hat{P}_{jc}$ denoting the probability of patch $j$ for class $c$ and computed by the dot product similarity between patch feature and textual class feature as

\begin{equation}
\label{eq:zero-shot-pred}
\hat{P}_{jc} \;=\; s_{\CC}(\mathbf{x}_{j}^{\top}\mathbf{t}_c),
\end{equation}
where $s_\CC(\cdot)$ is the softmax over class set $\CC$.
Then, low-resolution prediction $\hat{P}$ is reshaped and upsampled to the full-resolution prediction $\hat{Y} \in [0,1]^{H \times W \times C}$, and segmentation is obtained by the $\argmax$ of $\hat{Y}$ over classes.

\subsection{Full textual and visual support}
\label{sec:support_for_all}
In the following, we describe how to process examples in the textual and visual support sets to construct two support features sets, namely the \emph{visual support feature set} and the \emph{fused support feature set} combining visual and textual information.
During inference, the elements of the support feature sets that are the most relevant to a test image are used to train a lightweight patch-level linear classifier, which is then applied to the features of the same test image. Overview of this process is shown in~\autoref{fig:support-construction}.

\textbf{Visual support features.} 
Given a support image $I^i$ and its ground-truth segmentation labels $Y^{i} \in \{0,1\}^{H \times W \times C}$ at full resolution, its patch-level feature matrix $X^{i} \in \RR^{n \times d}$ is extracted.
Then, the ground-truth pixel-level labels $Y^{i}$ are down-sampled and reshaped to obtain patch-level labels, which are not binary anymore due to interpolation and are subsequently $L_1$-normalized per class (column) to obtain $P^{i} \in [0,1]^{n \times C}$.
These labels are used to pool the patch features into \emph{per-image visual class features} $\mathbf{v}^{i}_{c}$ for classes $\Ci$ present in image $i$ by

\begin{equation}
\label{eq:z-vecs}
\mathbf{v}^{i}_{c} \;=\; \sum_{j=1}^{n} P^{\,i}_{jc}\, \mathbf{x}^{\,i}_{j}.
\end{equation}

Assuming $M$ available support images, the union of per-image visual class features is given by
\begin{equation}
\label{eq:z-union}
\mathcal{V} \;=\; \bigcup_{i=1}^{M}  \{\mathbf{v}^{\,i}_{c} : c \in \Ci \},
\end{equation}
and referred to as \emph{visual support feature set}, which is used to support the test-time adaptation process.

\textbf{Fused support features.} 
We aim to leverage the semantic information in the textual class features. However, due to the modality gap between visual and textual features in VLMs~\cite{liang2022mind}, and because our goal is to classify image patch-level features, combining them directly with visual support features does not perform well, as confirmed by our experiments.
Thus, for class $c$, we create a \emph{fused class feature}
\begin{equation}
\label{eq:blend}
\mathbf{f}_{c\lambda} \;=\; \lambda\, \mathbf{t}_{c} \;+\; (1-\lambda)\, \mathbf{v}_{c},
\qquad \lambda \in [0,1],
\end{equation}
by combining textual class feature $\mathbf{t}_{c}$ and \emph{visual class feature} $\mathbf{v}_{c}$ obtained by aggregating all per-image visual class features for class $c$ across the visual support set by
\begin{equation}
\label{eq:z-class-agg}
\mathbf{v}_{c} \;=\; \sum_{i \in \Ic} \mathbf{v}^{\,i}_{c},
\end{equation}
where $\Ic$ is the set of support images that contain class $c$.
The fusion process is performed for a set of mixing coefficients $\Lambda \subseteq [0,1]$ to  capture diverse and complementary information from both modalities, yielding 
multiple fused class features per class. We denote the \emph{fused support feature set} by $\mathcal{F} = \{\mathbf{f}_{c\lambda} | c\in \CC, \lambda \in \Lambda\}$, which is used to support the test-time adaptation process.

\textbf{Support feature set maintenance.}
For each support image, we extract its feature matrix and pool patch features into per-image visual class features (\ref{eq:z-vecs}).
If a new support image arrives, we update the visual support feature set $\mathcal{V}$ (\ref{eq:z-union}), the visual class features (\ref{eq:z-class-agg}), the fused class features (\ref{eq:blend}), and consequently the fused support feature set $\mathcal{F}$.
Therefore, our support sets are dynamically expandable in a straightforward and efficient manner, allowing to operate in a continually evolving open-world scenario.

\textbf{Test-time adaptation.} For test image $I^q$ with feature matrix $X^q \in \RR^{n \times d}$, and the feature of patch $j$ denoted by $\mathbf{x}_j^q$, we train linear classifier $g_{\theta}: \RR^d \to \RR^{C}$, specifically for $I^q$, to project features to class probabilities. 
We leverage the relevant elements of the two support feature sets. 

To this end, we retrieve the $k$ nearest neighbors of each test image patch features from the visual support feature set and unite them into the \emph{retrieved visual support feature set}
\begin{equation}
\label{eq:knn}
\mathcal{V}_{r} \;=\; \bigcup_{j=1}^{n} \operatorname{kNN}\!\bigl(\mathcal{V},\, \mathbf{x}_{j}^q\bigr).
\end{equation}
We define the \emph{visual support loss} by
\begin{equation}
\label{eq:lr-weighted}
L_{v} \;=\; \sum_{\mathbf{v}\,\in\,\mathcal{V}_{r}} w_{\,l(\mathbf{v})}\;
\CE\bigl( g_{\theta}(\mathbf{v}) , \mathbf{1}_{l(\mathbf{v})} \bigr),
\end{equation}
where $\CE$ is the cross entropy loss, $l(\mathbf{v})$ provides the label of feature $\mathbf{v}$ (a per-image visual class feature), $\mathbf{1}_{c}$ is the one-hot encoding of class $c$, and $w_c$ is a \emph{class relevance weight} for class $c$.
This loss encourages the classifier to assign high probability to the correct class for each retrieved support feature.
The retrieved set is expected to include features corresponding to classes present in the test image, particularly those originating from visually similar images.

The class relevance weights are used to suppress the impact of retrieved features irrelevant to the test image.
Weight $w_c$ is estimated via the similarity between the image-level feature $\mathbf{x}^q \in \RR^d$
and the corresponding textual class feature, followed by softmax:
\begin{equation}
w_c \;=\; s_{\CC}\left(({\mathbf{x}^q)}^\top \mathbf{t}_{c}\right),
\end{equation}
with $\mathbf{x}^q$ given by global average pooling
\begin{equation}
\label{eq:qbar-wc}
\mathbf{x}^q \;=\; \frac{1}{n}\sum_{j=1}^{n}\mathbf{x}_{j}^q.
\end{equation}

We utilize textual support, via the fused support set, by training the classifier on the fused class features (denoted by $\mathcal{F}_{r}$ and referred to as \emph{retrieved fused support feature set}) of the classes (denoted by $\mathcal{C}_{r}$) that appear in retrieved visual support feature set $\mathcal{V}_{r}$.
This is performed via the \emph{fused support loss}:

\begin{equation}
\label{eq:lt}
L_{f} \;=\; \sum_{c \in \mathcal{C}_{r}} w_{c}\;
\sum_{\lambda \in \Lambda}
\operatorname{CE}\bigl( g_{\theta}(\mathbf{f}_{c\lambda}) , \mathbf{1}_{c} \bigr).
\end{equation}
We observe that using multiple mixing coefficients $\Lambda$ improves performance compared to using a single coefficient.
The total loss is given by $\mathcal{L} \;=\; L_{v} \;+\; \beta_{f}\, L_{f}.$
After training, classifier $g_{\theta}$ is applied to test image patch features to obtain patch-level predictions.
Then, this low resolution prediction map is upsampled to the original image resolution to obtain the final segmentation map.

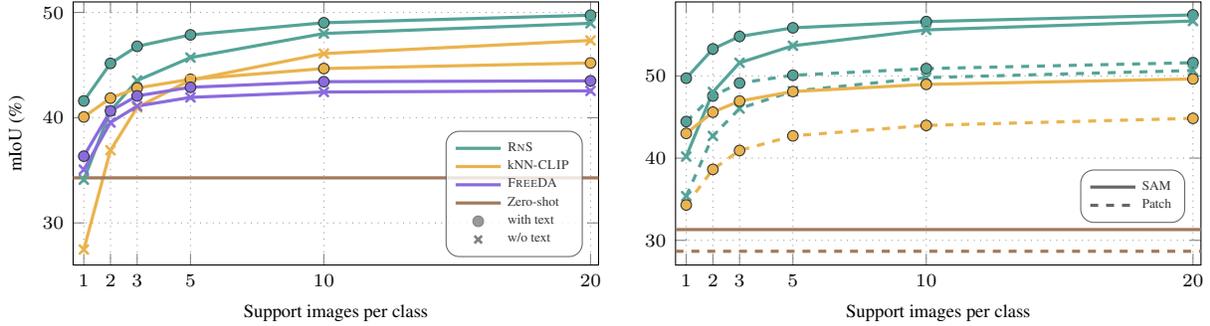
\begin{figure*}[t]
  \vspace{-5pt}
  \centering
  \begin{tabular}{@{\hspace{-10pt}}c@{\hspace{5pt}}c@{\hspace{-10pt}}}
      \input{plots/full_visual_support_varying_mem_avg} &
      \input{plots/full_visual_support_varying_mem_avg_2} \\
  \end{tabular}
  \vspace{-8pt}
  \caption{\textbf{Full textual and visual support.} 
  We compare zero-shot, \ours, \knnclip and \freeda and their variants without class name information  (w/o text) for increasing number of support images per class. 
  SAM 2.1 is used for region proposals.
  \emph{Left:} OpenCLIP (ViT-B/16) for region-level predictions.
  \emph{Right:} DINOv3.txt (ViT-L/16) for patch-level and region-level predictions.
  \label{fig:full_visual_vary_mem_two}}
\end{figure*}

\begin{figure*}[t]
  \centering
  \begin{tabular}{@{\hspace{-10pt}}c@{\hspace{5pt}}c@{\hspace{-10pt}}}
      \input{plots/partial_visual_support_mem} &
      \input{plots/partial_textual_support_mem} \\
  \end{tabular}
  \vspace{-8pt}
  \caption{\textbf{Partial visual (left) and textual (right) support settings.} 
  Results of zero-shot, \ours, \knnclip, \freeda and their variants without class name information (w/o text). 
  \ours evaluated w/o the pseudo-label loss in  (\ref{eq:proto-pseudo-loss}).
  OpenCLIP ViT-B/16 and SAM 2.1 are used. 
  Left: a fraction of classes lack visual examples, while $B=3$ for the rest.
  Right: a fraction of classes lack textual class names, and $B=1$.
  \label{fig:partial}}
\end{figure*}
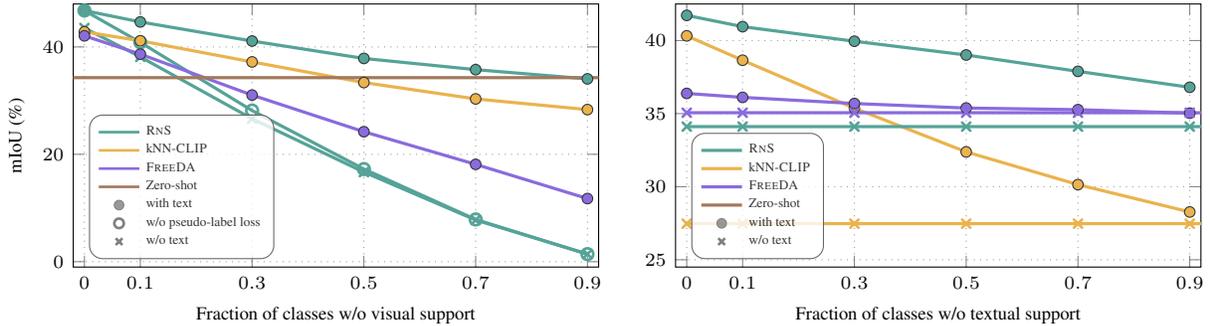

\subsection{Partial visual support}
\label{sec:support_for_some}
We present an additional loss, 
while other components remain as in Section \ref{sec:support_for_all}.

\textbf{Classes without visual support.}
The set of classes supported by their class name but with no image examples are denoted as $\CC_{d}$. 
Given the absence of visual support for these classes, we cannot compute their visual class feature $\mathbf{v}_{c}$ (\ref{eq:z-class-agg}), and consequently, their fused class feature $\mathbf{f}_{c\lambda}$ (\ref{eq:blend}).

To circumvent this, we exploit the test image to identify whether any of the classes in $\CC_{d}$ are present in it via zero-shot prediction $\hat{P}^q$ (\ref{eq:zero-shot-pred}).
Predictions in $\hat{P}^q$ are converted to one-hot vectors by assigning each patch to its most probable class, and the result is $L_1$-normalized per class (column) to obtain $\tilde{P}^q \in [0, 1]^{n \times C}$.
We denote by $\CC_{q} \subseteq \CC$ the set of classes assigned to at least one patch according to $\tilde{P}^q$.
Then, we define the visual class feature by pooling the patch features according to these pseudo-labels: 
\begin{equation}
\label{eq:zq}
\mathbf{v}_c \;=\; \sum_{j=1}^{n} \tilde{P}^{\,q}_{jc}\, \mathbf{x}_{j}^q, \quad c \in \CC_{d} \cap \CC_{q}.
\end{equation}
Note\footnote{Notation $\mathbf{v}_c$ used for both types of visual class features; (\ref{eq:z-class-agg}) used for classes with visual-textual support, while (\ref{eq:zq}) for those with only textual.} that we perform this only for classes in $\CC_{q} \cap \CC_{d}$ whose visual class feature cannot be obtained by (\ref{eq:z-class-agg}).
This allows us to perform the modality fusion by (\ref{eq:blend}) for all classes.

\textbf{Fused feature pseudo-labeling.}
Fused class feature $\mathbf{f}_{c\lambda}$ is derived through visual class features that use pseudo-labeling.
Therefore, its association with class $c$ is uncertain.
To circumvent this, we pseudo-label those fused class features.
The predicted probability distribution for $\mathbf{f}_{c\lambda}$ denoted by $\hat{\mathbf{p}}_{c\lambda} \in [0,1]^{C}$ has its
$c^\prime$ element, associated with class $c^\prime$, estimated by  $\mathbf{f}_{c\lambda} ^\top \mathbf{t}_{c^\prime}$ followed by softmax over all classes.

\textbf{Extended test-time adaptation loss.}
We introduce a \emph{pseudo-label loss} term exploiting such pseudo-labels:
\begin{equation}
\label{eq:proto-pseudo-loss}
L_{p} \;=\; \sum_{c \in \CC_d \cap \CC_q} w_c \sum_{\lambda \in \Lambda} \KL\bigl(   \hat{\mathbf{p}}_{c\lambda} \,\big\|\, g_{\theta}(\mathbf{f}_{c\lambda})  \bigr), 
\end{equation}
where $\KL$ is the Kullback-Leibler divergence. 

The total loss is given by $\mathcal{L} \;=\; L_{v} \;+\; \beta_{f}\,L_{f} \;+\; \beta_{p}\,L_{p}.$
Note that classes with visual support are handled in the second loss term, while the rest are either handled in the third loss term ($\CC_r \cap \CC_d = \emptyset$) or ignored if they are not predicted to be present in the test image.

\subsection{Partial textual support}
\label{partial-text}

We modify \emph{fused support features} of classes without textual 
support, while other components remain as in Section \ref{sec:support_for_all}.

\textbf{Classes without textual support.} When class names are absent, we cannot compute textual class features $\mathbf{t}_{c}$ or their fused counterparts $\mathbf{f}_{c\lambda}$ (\ref{eq:blend}). 
Excluding these classes from the loss introduces bias toward classes with both supports. 
Instead, we replace missing textual class features with the average textual class feature across classes with an available class name. 
This provides a neutral semantic prior, ensuring that all classes equivalently participate in the loss.

If textual support is missing for all classes, no textual class features can be formed. In this case, we simply set $\Lambda = \{0\}$ and class relevance weights $w_c$ are set equal to 1 for all classes,
effectively reducing to our \emph{w/o text} baseline.

\subsection{Region-proposal predictions}
\label{Proposals}

Our method so far assumes patch-level feature extraction and predictions, \ie $\mathbf{x}^q_j$ for patch $j$. 
When region proposals $S \in \{0,1\}^{H \times W \times R}$ for $R$ binary masks are available for the test image, \eg from SAM~\cite{kirillov2023segment,ravi2024sam2segmentimages}, we proceed as follows.
We downsample each mask to patch resolution and apply $L_1$ normalization per region, yielding $\bar{S} \in [0,1]^{n \times R}$.

We then pool patch features into region-level features by
\begin{equation}
\label{eq:region}
\mathbf{x}^{\,q}_r \;=\; \sum_{j=1}^{n} \bar{S}_{jr}\, \mathbf{x}^{\,q}_{j}, \quad r = 1, \ldots, R,
\end{equation}
where $r$ denotes the region index.
Finally, we assign labels at the region level and map them to the corresponding mask regions at the full image resolution to obtain the final segmentation map.

%% file: plots/full_visual_support_varying_mem_avg.tex
\begin{ShotMiouAxis}[
  width=.40\linewidth, height=3.5cm,
  scale only axis,
  xtick={1,2,3,5,10,20},
  xmin=0.6, xmax=20.3,
  ymin=26, ymax=51,
  xlabel={Support images per class},
]

\addplot[color=appleteal, line width=1.2pt] coordinates {(0,0)(1,1)};
\addlegendentry{\ours};

\addplot[color=applemustard, line width=1.2pt] coordinates {(0,0)(1,1)};
\addlegendentry{\knnclip};

\addplot[color=applepurple, line width=1.2pt] coordinates {(0,0)(1,1)};
\addlegendentry{\freeda};

\addplot[color=applebrown, line width=1.2pt] coordinates {(0,0)(1,1)};
\addlegendentry{Zero-shot};

\addplot[
  only marks,
  mark=*,
  mark options={fill=gray, draw=gray},
] coordinates {(0,0)};
\addlegendentry{with text};

\addplot[
  only marks,
  mark=x,
  mark options={draw=gray, line width=1pt},
] coordinates {(0,0)};
\addlegendentry{w/o text};

  \AddShotMiouCurve{}{appleteal}{
    (1,41.59) (2,45.16) (3,46.78) (5,47.87) (10,49.02) (20,49.74)
  }

  \AddShotMiouCurve{}{appleteal}{
    (1,34.11) (2,40.65) (3,43.53) (5,45.71) (10,48.00) (20,48.97)}[mark=x, mark size=2.2pt,    mark options={
     line cap=round,
     draw=appleteal, 
     line width=1pt,
     double=black, 
     double distance=0.55pt 
   }]

  \AddShotMiouCurve{}{applemustard}{
    (1,40.07) (2,41.86) (3,42.82) (5,43.65) (10,44.67) (20,45.20)
  }

  \AddShotMiouCurve{}{applemustard}{
    (1,27.47) (2,36.90) (3,40.97) (5,43.58) (10,46.09) (20,47.34)}[mark=x, mark size=2.2pt,    
    mark options={
     line cap=round,
     draw=applemustard,    
     line width=1pt,
     double=black,      
     double distance=0.55pt 
   }]
  \AddShotMiouCurve{}{applepurple}{
    (1,36.34) (2,40.64) (3,42.07) (5,42.89) (10,43.42) (20,43.50)
  }

  \AddShotMiouCurve{}{applepurple}{
    (1,35.06) (2,39.54) (3,41.10) (5,41.93) (10,42.44) (20,42.55)}[mark=x, mark size=2.2pt,    
     mark options={
     line cap=round,
     draw=applepurple,   
     line width=1pt,
     double=black,   
     double distance=0.55pt
   }]
   
  \AddZeroShotLine{}{applebrown}{34.28}
\end{ShotMiouAxis}

%% file: plots/full_visual_support_varying_mem_avg_2.tex
\begin{ShotMiouAxis}[
  width=.40\linewidth, height=3.5cm,
  scale only axis,
  xtick={1,2,3,5,10,20},
  xmin=0.6, xmax=20.3,
  ymin=27, ymax=59,
  ylabel={},
  legend style={
      draw=black!60,
      fill=white,
      rounded corners=5pt,
      font=\tiny,
    at={(rel axis cs:0.97,0.16)},
    anchor=south east,
  },
]

\AddShotMiouDashedCurve{}{appleteal}{
  (1,44.45) (2,47.56) (3,49.13) (5,50.08) (10,50.89) (20,51.61)
}

\AddShotMiouDashedCurve{}{appleteal}{
  (1,35.37) (2,42.70) (3,46.03) (5,48.14) (10,49.79) (20,50.66)
}[mark=x, mark size=2.2pt,    mark options={
     solid,
     line cap=round,
     draw=appleteal,
     line width=1pt,
     double=black,
     double distance=0.55pt 
   }]

\AddShotMiouDashedCurve{}{applemustard}{
  (1,34.30) (2,38.63) (3,40.93) (5,42.71) (10,43.98) (20,44.84)
}

\AddShotMiouCurve{}{appleteal}{
  (1,49.72) (2,53.28) (3,54.80) (5,55.85) (10,56.61) (20,57.42)
}

\AddShotMiouCurve{}{appleteal}{
  (1,40.19) (2,48.05) (3,51.60) (5,53.67) (10,55.60) (20,56.68)}[mark=x, mark size=2.2pt,    mark options={
     line cap=round,
     draw=appleteal, 
     line width=1pt,
     double=black,  
     double distance=0.55pt 
   }]

\AddShotMiouCurve{}{applemustard}{
  (1,43.01) (2,45.59) (3,46.94) (5,48.10) (10,48.97) (20,49.63)
}

\AddZeroShotLine{}{applebrown}{31.31}

\AddZeroShotDashedLine{}{applebrown}{28.67}

\addlegendimage{no markers, solid,  line width=1.2pt}
\addlegendentry{SAM}
\addlegendimage{no markers, dashed, line width=1.2pt}
\addlegendentry{Patch}

\end{ShotMiouAxis}

%% file: plots/partial_visual_support_mem.tex
\begin{DropFracAxis}[
  width=.40\linewidth, height=3.5cm,
  scale only axis,
  xtick={0.0,0.1,0.3,0.5,0.7,0.9},
  xmin=-0.02, xmax=0.92,
  ymin=-1.0, ymax=48
]

\addplot[color=appleteal, line width=1.0pt] coordinates {(-5.0,-5.0)(-5.0,-5.0)};
\addlegendentry{\ours};

\addplot[color=applemustard, line width=1.0pt] coordinates {(-5.0,-5.0)(-3.0,-3.0)};
\addlegendentry{\knnclip};

\addplot[color=applepurple, line width=1.0pt] coordinates {(-5.0,-5.0)(-5.0,-5.0)};
\addlegendentry{\freeda};

\addplot[color=applebrown, line width=1.0pt] coordinates {(-5.0,-5.0)(-5.0,-5.0)};
\addlegendentry{Zero-shot};

\addplot[
  only marks,
  mark=*,
  mark options={fill=gray, draw=gray},
] coordinates {(-5.0,-5.0)};
\addlegendentry{with text};

\addplot[
  only marks,
  mark=o,
  mark options={draw=gray, line width=1pt},
] coordinates {(-5.0,-5.0)};
\addlegendentry{w/o pseudo-label loss};

\addplot[
  only marks,
  mark=x,
  mark options={draw=gray, line width=1pt},
] coordinates {(-5.0,-5.0)};
\addlegendentry{w/o text};

\AddMiouDropCurve
  {}{appleteal}
  { (0.00,46.78) (0.10,44.66) (0.30,41.10) (0.50,37.86) (0.70,35.77) (0.90,34.05) };

\AddMiouDropCurve
  {}{appleteal}
  { (0.00,46.78) (0.10,40.85) (0.30,28.15) (0.50,17.26) (0.70,7.87) (0.90,1.41) }[mark=o, mark size=2.2pt,    mark options={
 solid,
 line cap=round,
 draw=appleteal, 
 line width=1pt,
 double=black, 
 double distance=0.55pt 
}];

\AddMiouDropCurve
  {}{appleteal}
  { (0.00,43.53) (0.10,38.15) (0.30,26.63) (0.50,16.69) (0.70,7.76) (0.90,1.42)}[mark=x, mark size=2.2pt,    mark options={
 solid,
 line cap=round,
 draw=appleteal,    
 line width=1pt,
 double=black,   
 double distance=0.55pt 
}];

\AddMiouDropCurve
  {}{applemustard}
  { (0.00,42.82) (0.10,41.15) (0.30,37.21) (0.50,33.37) (0.70,30.32) (0.90,28.33) };
  
\AddMiouDropCurve
  {}{applepurple}
  { (0.00,42.07) (0.10,38.66) (0.30,31.03) (0.50,24.21) (0.70,18.13) (0.90,11.76) };

\AddMiouBaseline{}{applebrown}{34.28}

\end{DropFracAxis}

%% file: plots/partial_textual_support_mem.tex
\begin{DropFracTextAxis}[
  width=.40\linewidth,
  height=3.5cm,
  scale only axis,
  xtick={0.0,0.1,0.3,0.5,0.7,0.9},
  xmin=-0.02, xmax=0.92,
  ymin=24.5, ymax=42.5,
  ylabel={},
]

\addplot[color=appleteal, line width=1.2pt] coordinates {(0,0)(1,1)};
\addlegendentry{\ours};

\addplot[color=applemustard, line width=1.2pt] coordinates {(0,0)(1,1)};
\addlegendentry{\knnclip};

\addplot[color=applepurple, line width=1.2pt] coordinates {(0,0)(1,1)};
\addlegendentry{\freeda};

\addplot[color=applebrown, line width=1.2pt] coordinates {(0,0)(1,1)};
\addlegendentry{Zero-shot};

\addplot[
  only marks,
  mark=*,
  mark options={fill=gray, draw=gray},
] coordinates {(0,0)};
\addlegendentry{with text};

\addplot[
  only marks,
  mark=x,
  mark options={draw=gray, line width=1pt},
] coordinates {(0,0)};
\addlegendentry{w/o text};

\AddMiouDropCurve{}{appleteal}{
  (0.00,41.72)
  (0.10,40.95)
  (0.30,39.95)
  (0.50,39.01)
  (0.70,37.89)
  (0.90,36.80)
};

\AddMiouDropCurve{}{applemustard}{
  (0.00,40.32)
  (0.10,38.65)
  (0.30,35.37)
  (0.50,32.38)
  (0.70,30.14)
  (0.90,28.27)
};

\AddMiouDropCurve{}{applepurple}{
  (0.00,36.38)
  (0.10,36.11)
  (0.30,35.69)
  (0.50,35.38)
  (0.70,35.27)
  (0.90,35.03)
};

\AddMiouDropCurve
  {}{appleteal}
  { (0.0,34.11) (0.1,34.11) (0.3,34.11) (0.5,34.11) (0.7,34.11) (0.9,34.11) (20,34.11)}[mark=x, mark size=2.2pt,    mark options={
 solid,
 line cap=round,
 draw=appleteal, 
 line width=1pt,
 double=black, 
 double distance=0.55pt 
}];

\AddMiouDropCurve
  {}{applemustard}
  {(0.0,27.47) (0.1,27.47) (0.3,27.47) (0.5,27.47) (0.7,27.47) (0.9,27.47) (20,27.47)}[mark=x, mark size=2.2pt,    mark options={
 solid,
 line cap=round,
 draw=applemustard, 
 line width=1pt,
 double=black,  
 double distance=0.55pt 
}];

\AddMiouDropCurve
  {}{applepurple}
  {(0.0,35.06) (0.1,35.06) (0.3,35.06) (0.5,35.06) (0.7,35.06) (0.9,35.06) (20,35.06)}[mark=x, mark size=2.2pt,    mark options={
 solid,
 line cap=round,
 draw=applepurple,   
 line width=1pt,
 double=black,   
 double distance=0.55pt 
}];

\end{DropFracTextAxis}

%% file: sec/experiments.tex
\section{Experiments}
\label{sec:experiments}

\subsection{Experimental setup}
\label{sec:setup}

\textbf{Datasets and evaluation.}
We evaluate on the validation split of six OVS benchmarks: PASCAL VOC~\cite{everingham2010pascal} (VOC), PASCAL Context~\cite{mottaghi2014context} (Context), COCO Object (Object), COCO-Stuff~\cite{caesar2018cocostuff} (Stuff), Cityscapes~\cite{cordts2016cityscapes} (City), and ADE20K~\cite{zhou2017ade20k} (ADE). We report the \emph{average mIoU} over all datasets unless otherwise noted. Per dataset results are presented in the supplementary. We also evaluate on PASCAL Context-59 (C-59)~\cite{mottaghi2014context}, FoodSeg103 (Food)~\cite{wu2021foodseg}, and CUB~\cite{welinder2010cub} to compare OVS and fully supervised methods. Details about sampling the $B$ support images per class for the visual support set, and the construction of the few-shot benchmark are shown in the supplementary.

\textbf{Implementation details.} 
We use OpenCLIP ViT-B/16~\cite{cherti2023reproducible} trained on LAION~\cite{schuhmann2022laion} and apply the MaskCLIP trick~\cite{maskclip}. 
For DINOv3~\cite{simeoni2025dinov3}, we use the public ViT-L/16 checkpoint adapted with dino.txt~\cite{jose2024dinov2meetstextunified} to derive text-aligned patch features, denoted by DINOv3.txt. For region proposals, we run SAM 2.1~\cite{ravi2024sam2segmentimages} Hiera-L on each query image with a $32{\times}32$ grid of points (one mask per point) and non-maximum suppression to ensure non-overlap. More implementation details are shown in the supplementary.

\textbf{Competitors.} We compare to zero-shot prediction (Section~\ref{zero-shot}), and to two retrieval-based OVS methods that leverage both visual and textual support sets: \knnclip~\cite{gui2024knnclip} , which we re-implement, and \freeda~\cite{freeda}, where we adapt the official code to use a real support set, \ie pixel-annotated images, rather than synthetic-only prototypes. Please refer to the supplementary material for details in the implementation of the aforementioned competitors. 
We also report performance for training a linear classifier or the full network ``offline'' on the entire support set in a closed-set manner. These variants lose the open-vocabulary ability as they cannot provide predictions for unseen classes, but provide useful reference baselines.
Specifically, we consider the following three offline training methods:
    (i) Linear classifier on per-image visual class features $\mathbf{v}_c^i$ and their corresponding labels. 
    (ii) Offline linear classifier on patch-level features $\mathbf{x}_j^i$  and pixel-level annotations. Predictions are upsampled to apply the loss at the full image resolution.
    (iii) Offline finetuning on images and pixel-level annotations. This configuration builds upon the previous setup but, in addition to training the linear classifier, it also finetunes the parameters of the vision encoder.
We also compare to SOTA fully supervised and different kinds of OVS methods in \autoref{tab:ovseg_comparison}.

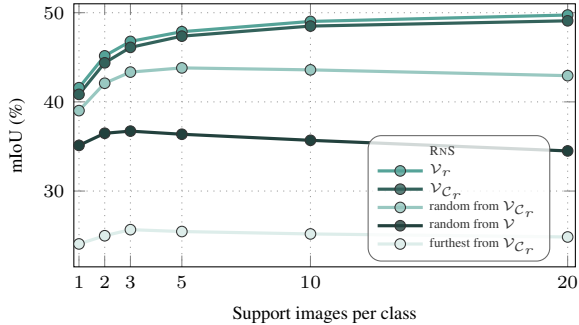
\begin{figure}
  \vspace{-1.5pt}
  \centering
  \input{plots/retrieval}
  \vspace{-6pt}
  \caption{\textbf{Impact of retrieval on \ours.} We replace the retrieved visual support feature set $\mathcal{V}_r$ of \ours with a random subset of the visual support feature set $\mathcal{V}$, or different variants of visual support features from the retrieved classes $\mathcal{V}_{\mathcal{C}_r}$.}
  \vspace{-2pt}
  \label{fig:ablations}
\end{figure}

\begin{table}[t]
    \centering
    \resizebox{\linewidth}{!}{%
    \input{tab/ablation_method}

    }
      \vspace{-3pt}
    \caption{\textbf{Ablations of \ours.} We report average mIoU across the considered datasets for three different numbers of available support images per class ($B=1$, $B=5$, $B=10$). Blue numbers denote difference to the number in the same column but first row.}
    \label{tab:ablations}
    \vspace{-8pt}
\end{table}

\subsection{Experimental results}

\textbf{Full textual and visual support.}
\autoref{fig:full_visual_vary_mem_two} reports mIoU as we vary the number of support images $B$ per class. 
\ours consistently outperforms all competitors on every $B$, both backbones, and input feature granularity, with large gains. Moreover, it yields significant improvement over zero-shot segmentation: $+7.3\%$ on OpenCLIP and $+18.4\%$ on DINOv3.txt with just one image per class. 
\ours effectively leverages text, achieving a performance gain at $B=1$ over the w/o-text variant while performing on par in the $B=20$ case, indicating that textual priors are valuable when support is sparse, while visuals dominate as support densifies.
Interestingly, \knnclip’s fusion heuristics help at $B=1$, but hinder after $B=5$, suggesting sensitivity to hand-tuned fusion as support grows. 
The variant of \knnclip that discards text is competitive when enough support images are available but scores low in the few-shot regime, where the diversity and quality of the support are minimal. 
\freeda does not benefit enough from combining textual with visual support; low gains with respect to w/o-text baseline. 
Moreover, on newer backbones with stronger localization, e.g., DINOv3, the gap between \ours and \knnclip widens, indicating that \ours scales better with better visual representations.
Region-proposals consistently boost performance compared to patch-level predictions, by leveraging the large-scale training of SAM, at the expense of higher computational cost at inference time.

\textbf{Partial visual support.}
In~\autoref{fig:partial} (left), we vary the fraction of classes without visual examples, while keeping text descriptions for all. 
\ours degrades smoothly, and manages to benefit from the available visual support even for small fractions of supported classes.
Removing pseudo-label loss (\ref{eq:proto-pseudo-loss}) leads to a steep drop in performance; this is our mechanism to compensate for the missing visual support. 
Further removing textual support (w/o text) degrades performance more, as the model cannot leverage neither fusion nor class relevance weights.
\knnclip and \freeda drop significantly, and soon perform below zero-shot, as they do not account for missing classes in their support set.
Note that missing visual support for some classes is inevitable in an open-world and dynamically expanding environment. 

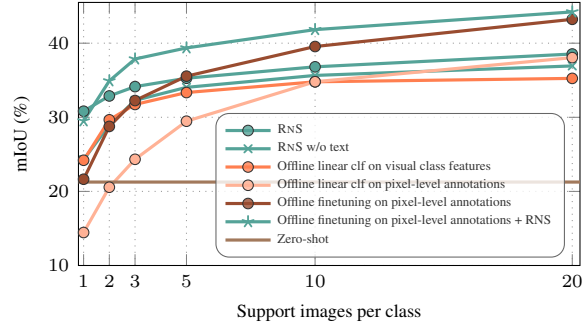
\begin{figure}
  \centering
  \input{plots/offline}
    \vspace{-6pt}
  \caption{\textbf{Comparison in a closed vocabulary setting.} We compare \ours to the offline baseline competitors. 
   To ensure a fair comparison we tune the learning rate, batch size, and number of iterations using a train-validation split from the available support images.
  No mask proposals are used. We report average performance on VOC, ADE, and Stuff.
  }
  \vspace{-12pt}
  \label{fig:offline_baselines}
\end{figure}

\textbf{Partial textual support.}
In \autoref{fig:partial} (right), we vary the fraction of classes without text descriptions, while keeping the visual support set intact. 
\ours remains the best across the entire range, dropping only mildly as text becomes scarce. 
Results suggest that training jointly on visual and textual support is superior to heuristic late fusion of independent predictions; \knnclip degrades steeply, while \freeda barely benefits even when text is available. \ours uses text as an auxiliary signal, used for fusion and relevance weighting, yielding a consistent boost when available but without making the model fragile when it disappears. 

\textbf{Ablations} are presented in \autoref{tab:ablations}. Removing class relevance weights ($w_c$) yields a reduction across all shots, confirming their role in suppressing irrelevant retrieved classes. Additionally, creating fused features (\ref{eq:blend}) with a single $\lambda=0.8$ harms especially the low-shot regime, where effective use of text fusion is more important. 
Overall, components of \ours provide complementary benefits, with lower but considerable impact as visual support increases.

In~\autoref{fig:ablations}, we analyze the impact of retrieval mechanism on the performance of \ours. When replacing retrieved visual support feature set $\mathcal{V}_r$ (\ref{eq:knn}) with a random subset of $\mathcal{V}$ performance degrades a lot, confirming that using similar image regions is crucial. 
Then, we perform the default retrieval (\ref{eq:knn}) but instead of keeping $\mathcal{V}_r$, we focus on retrieved classes $\mathcal{C}_r$, and form a set of visual support features from the retrieved classes $
\mathcal{V}_{\mathcal{C}_r}{=}\bigcup_{i=1}^{M}  \{\mathbf{v}^{\,i}_{c} : c \in \mathcal{C}_{r} \}$. 
We perform the test-time adaptation using different subsets of this set.
Using the full set $\mathcal{V}_{\mathcal{C}_r}$ yields slightly lower performance than the default approach, suggesting that $\mathcal{V}_{\mathcal{C}_r}$ includes irrelevant examples that are filtered out by \ours. Using a random subset of $\mathcal{V}_{\mathcal{C}_r}$ performs quite better than a random subset of $\mathcal{V}$, indicating that restricting adaptation to semantically relevant classes is beneficial. In contrast, selecting the furthest per query feature examples from $\mathcal{V}_{\mathcal{C}_r}$ results in the worst performance, even below random sampling from $\mathcal{V}$.
These demonstrate that retrieval in \ours identifies relevant support features, which improves performance significantly.

\begin{figure}[t]
  \vspace{-12pt}
  \centering
  \resizebox{\linewidth}{!}{\includegraphics{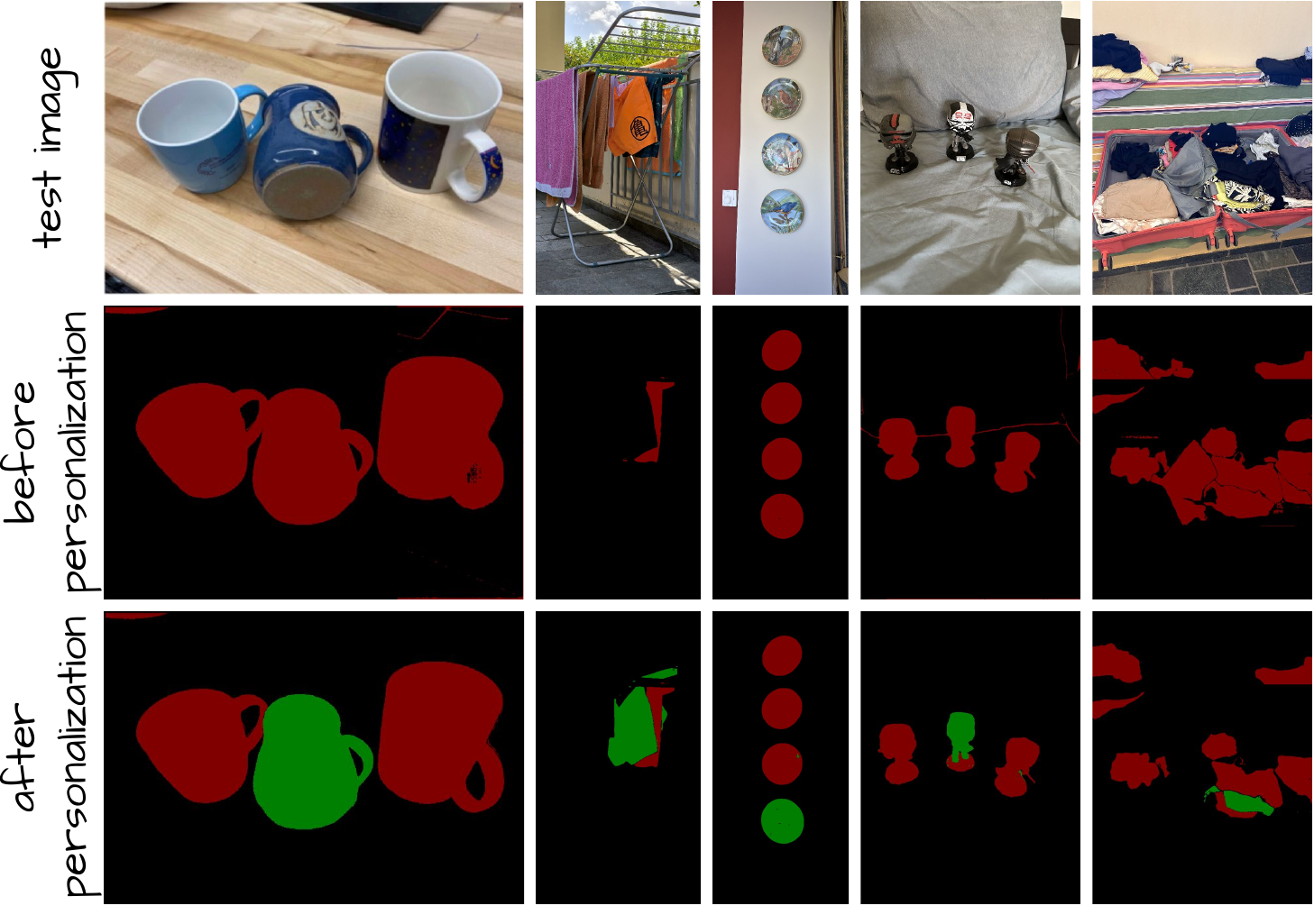}}
  \vspace{-15pt}
  \caption{\textbf{\ours for personalized segmentation.} We append examples of a specific instance to the support. \textbf{{\color{personalgreen}Green:}} personalized instance, \textbf{{\color{personalred}red:}} generic class, and \textbf{{\color{black}black:}} background.
  Support sets shown in the supplementary.}
  \vspace{-15pt}
  \label{fig:personalization}
\end{figure}

\textbf{Closed-set comparisons.}
To further evaluate the effectiveness of \ours, we compare it against fully supervised offline baselines and present results in~\autoref{fig:offline_baselines}. 
The offline classifier trained on visual class features performs comparably to \ours w/o text in the $B=1$ and $B=2$ scenarios, but its performance deteriorates as the number of images per class increases. This highlights the advantage of test-time retrieval, which dynamically selects the most relevant visual class features from the support set for each test image.
Offline methods trained on pixel-level annotations perform poorly in low-shot settings, indicating that such methods require larger amounts of data to avoid overfitting. Freezing the backbone and training only a linear classifier with pixel-level cross entropy loss underperforms \ours in all shots, suggesting that our method and objective are more effective in few-shot support utilization. On the other hand, training the backbone as well results in stronger performance when sufficient support is available, at the cost of additional training computational complexity. The best performance is obtained when combining the pretrained backbone weights from the latter experiment with \ours by adapting the linear classifier during test-time. This suggests again: (i) the complementary to the visual support gains from the use of textual supervision with our method (ii) the efficacy of our test-time training on test-image relevant support compared to offline training on the entire support.

\begin{table}[t]
\vspace{-10pt}
\centering
\renewcommand{\arraystretch}{1.15}
\setlength{\tabcolsep}{4pt}
\resizebox{\linewidth}{!}{%
\input{tab/soup_tmp}
}
\vspace{-2pt}
\caption{\textbf{OVS \vs fully supervised segmentation.} \emph{Fully supervised}: best method picked per dataset.
$^{*}$: self-evaluated, $^{\dag}$: from CAT-Seg. \emph{Domain}: using annotations from each training set. \emph{Annot:} the number of pixel-level semantic segmentation annotations that each method uses. For \emph{domain} we report the ADE  annotations. Full table and details in the supplementary.
\label{tab:ovseg_comparison}
\vspace{-12pt}}
\end{table}

\textbf{Personalized segmentation.}
In~\autoref{fig:personalization}, we qualitatively demonstrate that the dynamic expandability of the support set in \ours enables personalized segmentation of particular object instances within a class. Starting from a support set that allows \ours to segment pixels of a class, \eg, \texttt{plate}, appending only a few examples of a particular instance, \eg, \texttt{plate with a kingfisher / my plate}, enables the model to distinguish that instance from the broader class.
We observe that \ours accurately segments partially occluded personalized objects, such as the \texttt{skirt with a tropical pattern}. Failure cases remain; \eg, \ours incorrectly segments part of an \texttt{orange towel} as a \texttt{swimsuit}, likely due to insufficient contextual information and reliance on color cues.
Nevertheless, results illustrate that \ours is effectively employed for personalized segmentation without any modifications.

\textbf{Bridging the gap.} In~\autoref{tab:ovseg_comparison}, we compare different OVS methods, including \ours, and fully supervised ones (closed vocabulary). We consider methods from four categories (i) SAN and CAT-Seg as OVS methods that train on COCO-Stuff, (ii) LPOSS and CorrCLIP as training free OVS methods, (iii) \ours as an OVS method that uses support sets constructed from annotated images, (iv) fully supervised methods pre-trained offline on a closed set of categories. \ours with $B=20$, narrows down the gap to fully supervised segmentation to $11.5$ on average, improving the zero-shot baseline by $34$. \ours also surpasses the second best OVS method, CAT-Seg, by $14.1$, even though it uses much less pixel-level annotations. The improvement is more evident on fine-grained datasets like CUB and Food that are far from the domain of COCO-Stuff, which serves as a training set for many OVS methods. 

%% file: plots/retrieval.tex
\begin{ShotMiouAxis}[
  width=0.8\linewidth, height=3.5cm,
  scale only axis,
  xtick={1,2,3,5,10,20},
  xmin=0.8, xmax=20.3,
  ymin=21.5, ymax=51,
  xlabel={Support images per class},
  ylabel={mIoU (\%)},
  legend style={
      draw=black!60,
      fill=white,
      rounded corners=5pt,
      font=\tiny,
    at={(rel axis cs:0.95,0.02)},
    anchor=south east,
  },
]

  \addlegendimage{empty legend}
  \addlegendentry{\ours}

  \AddShotMiouCurve{$\mathcal{V}_r$}{appleteal}
    { (1,41.59) (2,45.16) (3,46.78) (5,47.87) (10,49.02) (20,49.74) }

  \AddShotMiouCurve{$\mathcal{V}_{\mathcal{C}_r}$}{appletealD2}
    { (1,40.84) (2,44.38) (3,46.11) (5,47.37) (10,48.50) (20,49.09) }

  \AddShotMiouCurve{random from $\mathcal{V}_{\mathcal{C}_r}$}{appletealL2}
    { (1,39.02) (2,42.09) (3,43.34) (5,43.81) (10,43.59) (20,42.94) }

  \AddShotMiouCurve{random from $\mathcal{V}$}{appletealD3}
    { (1,35.12) (2,36.48) (3,36.72) (5,36.37) (10,35.69) (20,34.50) }

  \AddShotMiouCurve{furthest from $\mathcal{V}_{\mathcal{C}_r}$}{appletealL4}
    { (1,24.07) (2,24.99) (3,25.67) (5,25.46) (10,25.19) (20,24.85) }

\end{ShotMiouAxis}

%% file: tab/ablation_method.tex
\begin{tabular}{llll}
    \toprule
    \textbf{Method} & \textbf{$B=1$} & \textbf{$B=5$} & \textbf{$B=10$} \\
    \midrule
    \ours                          & 41.59 & 47.87 & 49.02 \\
    \ours\ w/o $w_c$               & 41.20 \textcolor{blue}{\,-0.39} & 47.43 \textcolor{blue}{\,-0.44} & 48.54 \textcolor{blue}{\,-0.48} \\
    \ours\ w/o $w_c\,$ $\Lambda = \{0.8\}$ & 36.40 \textcolor{blue}{\,-5.19} & 46.55 \textcolor{blue}{\,-1.32} & 48.38 \textcolor{blue}{\,-0.64} \\
    \ours\ w/o text                & 34.11 \textcolor{blue}{\,-7.48} & 45.71 \textcolor{blue}{\,-2.16} & 48.00 \textcolor{blue}{\,-1.02} \\
    \bottomrule
\end{tabular}%

%% file: plots/offline.tex
\begin{ShotMiouAxis}[
  width=0.8\linewidth, height=3.5cm,
  scale only axis,
  xtick={1,2,3,5,10,20},
  xmin=0.8, xmax=20.3,
  ymin=10, ymax=45.5,
  legend style={
      draw=black!60,
      fill=white,
      rounded corners=5pt,
      font=\tiny
  },
  xlabel={Support images per class},
]

  \AddShotMiouCurve{\ours}{appleteal}
  { (1,30.82) (2,32.88) (3,34.14) (5,35.27) (10,36.81) (20,38.54) }

  \AddShotMiouCurve{\ours w/o text }{appleteal}
    { (1,24.28) (2,28.98) (3,32.28) (5,34.04) (10,35.65) (20,36.93) }[mark=x, mark size=2pt,    mark options={
     line cap=round,
     draw=appleteal,
     line width=0.7pt,
     double=black,
     double distance=0.55pt
   }]

  \AddShotMiouCurve{Offline linear clf on visual class features}{appleorange}
    { (1,24.20) (2,29.66) (3,31.73) (5,33.34) (10,34.79) (20,35.25) }

  \AddShotMiouCurve{Offline linear clf on pixel-level annotations}{appleorangeL2}
    { (1,14.45) (2,20.56) (3,24.33) (5,29.47) (10,34.80) (20,38.05) }

  \AddShotMiouCurve{Offline finetuning on pixel-level annotations}{appleorangeD2}
    { (1,21.64) (2,28.77) (3,32.26) (5,35.55) (10,39.55) (20,43.22) }

  \AddShotMiouCurve{Offline finetuning on pixel-level annotations + RNS}{appleteal}
    { (1,29.51) (2,34.91) (3,37.87) (5,39.36) (10,41.83) (20,44.22) }[mark=star,
     mark size=2.2pt,
     mark options={
        line cap=round,
        draw=appleteal,
        line width=0.7pt,
        fill=appleteal
     }
    ]

  \AddZeroShotLine{Zero-shot}{applebrown}{21.26}

\end{ShotMiouAxis}

%% file: tab/soup_tmp.tex
\begin{tabular}{lcccccccccc}
\toprule
\textbf{Method} & \textbf{Training set} & \textbf{Annot.} & \textbf{VOC} & \textbf{City} & \textbf{ADE} & \textbf{C-59} & \textbf{Food} & \textbf{CUB} & \textbf{Avg.} \\
\midrule
SAN$^{\dag}$~\cite{xu2023san} & COCO & 118k & -- & -- & 32.1 & 57.7 & 24.5 & 19.3 & -- \\
CAT-Seg~\cite{cho2024catseg} & COCO & 118k & \underline{82.5} & 47.0$^{*}$ & 37.9 & \underline{63.3} & 33.3 & 22.9 & 47.8 \\
LPOSS+~\cite{stojnic2025lposs} & \xmark & \xmark & 62.4 & 37.9 & 22.3 & 38.6 & 26.1 & 12.0 & 33.2 \\
CorrCLIP~\cite{zhang2025corrclip} & \xmark & \xmark & 76.4 & 49.9 & 28.8 & 47.9 & -- & -- & -- \\
\midrule
DINOv3.txt~\cite{simeoni2025dinov3} + SAM & \xmark & \xmark & 31.3 & 39.3 & 27.7 & 36.3 & 27.2 & 5.8 & 27.9 \\
\rowcolor{appletealL3}
+ \ours~$B=1$ & Domain & 66 & 73.2 & 59.1 & 37.3 & 52.7 & 42.8 & 34.0 & 49.9 \\
\rowcolor{appletealL3}
+ \ours~$B=20$ & Domain & 964 & 82.1 & 61.7 & 47.8 & 62.5 & \textbf{52.2} & \underline{65.2} & \underline{61.9} \\
\midrule
\midrule
\textbf{Fully Supervised} & Domain & 20k & \textbf{90.4} & \textbf{87.0} & \textbf{63.0} & \textbf{70.3} & \underline{45.1} & \textbf{84.6} & \textbf{73.4} \\
\bottomrule
\end{tabular}%

%% file: sec/conclusion.tex
\section{Conclusion}
\label{sec:conclusion}

We introduce \ours, a retrieval–augmented test-time adapter for open-vocabulary segmentation that learns a lightweight per-image linear classifier on frozen VLM features by fusing textual class features with retrieved visual support features. The method operates on patches or region proposals, handles full and partial support with a single objective, and avoids any type of hand-crafted solutions. Across six benchmarks and two backbones, \ours consistently outperforms all baselines and competitors. Finally, the dynamic support mechanism makes \ours naturally suited for open-world fine-grained tasks, \ie personalized segmentation.

%% file: sec/acknowledgements.tex
\section*{Acknowledgments}
\label{sec:acknowledgements}
This work was supported by the Czech Technical University in Prague grant No. SGS23/173/OHK3/3T/13, the EU Horizon Europe programme MSCA PF RAVIOLI (No. 101205297), and the Junior Star GACR GM 21-28830M. We acknowledge VSB–Technical University of Ostrava, IT4Innovations National Supercomputing Center, Czech Republic, for awarding this project (OPEN-33-67) access to the LUMI supercomputer, owned by the EuroHPC Joint Undertaking, hosted by CSC (Finland) and the LUMI consortium, through the Ministry of Education, Youth and Sports of the Czech Republic via the e-INFRA CZ project (ID: 90254). The access to the computational infrastructure of the OP VVV funded project CZ.02.1.01/0.0/0.0/16\_019/0000765 ``Research Center for Informatics'' is also gratefully acknowledged.

%% file: sec/supp.tex
\clearpage
\setcounter{page}{1}
\maketitlesupplementary

\section*{Supplementary Overview}
Supplementary material contains:
\begin{itemize}
  \item \textbf{Experimental setup details (Section~\ref{sec:experimental_details})} — evaluation protocol (Section~\ref{sec:protocol}), implementation specifics (Section~\ref{sec:implementation}), competitor configurations (Section~\ref{sec:competitors}), and details on offline baseline comparisons (Section~\ref{sec:offline_baselines}).
  \item \textbf{Additional experimental results (Section~\ref{sec:add_results})} — out-of-domain support (Section~\ref{sec:ood}), partial visual support results on finegrained datasets (Section~\ref{sec:partial_finegrained}), backbone comparison (Section~\ref{sec:backbones}), seen/unseen class analysis (Section~\ref{sec:unseen}), a complete version of Table 2 in the main paper (Section~\ref{sec:ovss_vs_supervised}), runtime comparisons (Section~\ref{runtimes}), and an additional ablation study (Section~\ref{sec:ablation_k}).
  \item \textbf{Qualitative results (Section~\ref{sec:qual})} — SAM Mask vs.\ patch-level prediction comparisons (Section~\ref{sec:sam_vs_patch}), qualitative comparisons on considered baselines (Section~\ref{sec:qual_modalities} and Section~\ref{sec:qual_competitors}), and support sets for personalized segmentation results reported in the main paper (Section~\ref{personalization}).
  \item \textbf{Per dataset results} of Figures 3-4 of the main paper are reported in \autoref{fig:full-clip-per-dataset}, \autoref{fig:full-dino-per-dataset}, \autoref{fig:partial-visual-per-dataset} and \autoref{fig:partial-text-per-dataset}.
\end{itemize}

\section{Experimental setup details}
\label{sec:experimental_details}

\subsection{Evaluation protocol}
\label{sec:protocol}

To construct our few-shot benchmark, we sample support images from the \emph{training split} of each dataset. We create a random permutation of the classes. Then we iterate over classes and randomly sample \(B\) images per class. If an encountered class already appears in previously sampled images \(B\) times, we skip it.  This procedure intentionally \emph{over-represents frequent classes}, \eg \texttt{road} in driving scenes, preserving a realistic long tail distribution. We create the support set with \emph{four random seeds}, except for VOC and City that we use \emph{eight random seeds}, and report average performance. For partial support experiments, we randomly drop visual or textual support for a fraction of classes. For partial textual support this corresponds to a fraction of test class names, while for partial visual support this corresponds to the annotated visual features of visually supported classes.

\begin{figure}[t]
  \centering
  \resizebox{0.98\linewidth}{!}{\includegraphics{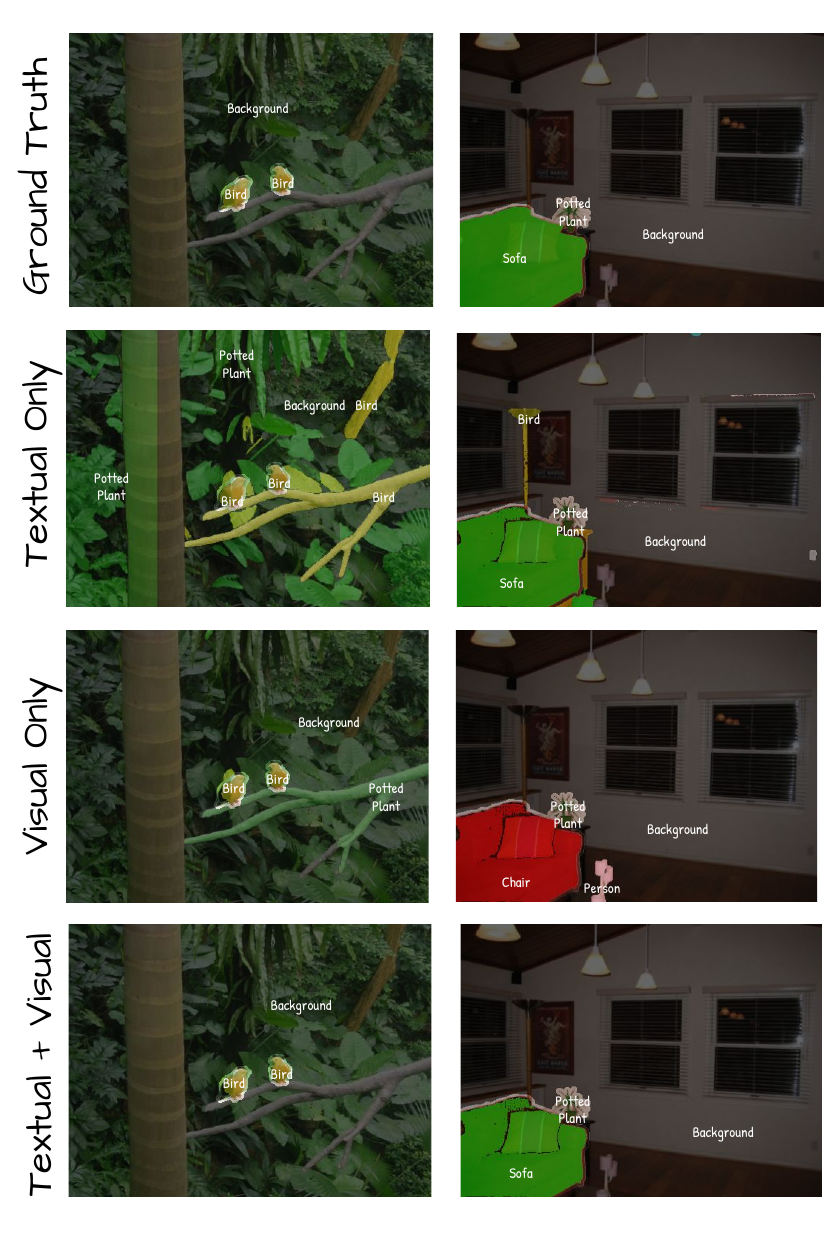}}
  \vspace{-16pt}
  \caption{\textbf{Open-vocabulary segmentation (OVS) results.} We compare three settings: (i) \emph{textual-only} support (zero-shot OVS), (ii) a simplified version of \texttt{\ours} using \emph{visual-only} support, and (iii) the full \texttt{\ours} combining \emph{textual+visual} support. Text-only support leads to ambiguous predictions (tree branch as bird, and various background hallucinations). Visual-only support helps disambiguate some classes but still confuses contextually similar objects (sofa--chair, tree branch--potted plant). \ours effectively combines both modalities to achieve accurate segmentation.}
  \vspace{-12pt}
  \label{fig:ovs_suppl}
\end{figure}

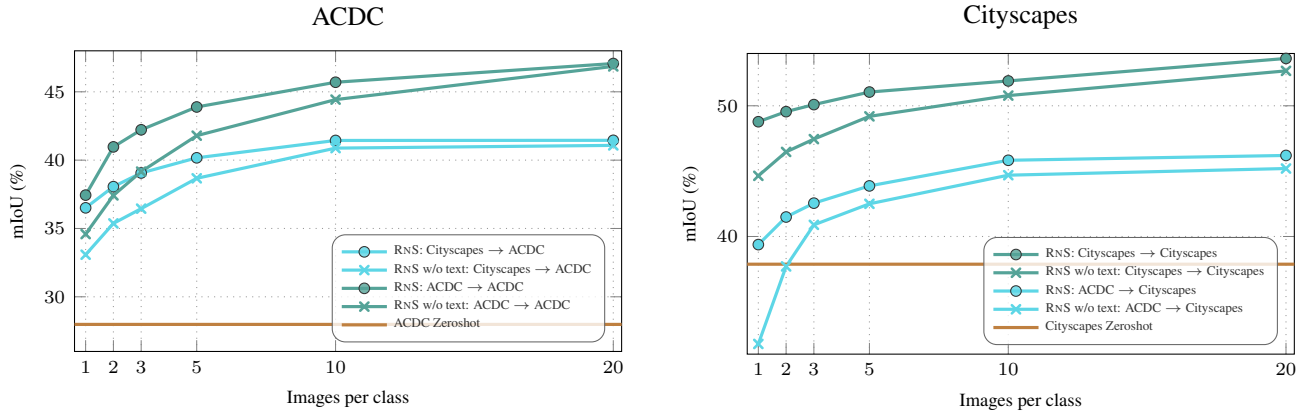
\begin{figure*}
  \vspace{-8pt}
  \centering

  \begin{minipage}{0.49\linewidth}
    \centering
    \input{plots/ood2}
  \end{minipage}
  \hfill
  \begin{minipage}{0.49\linewidth}
    \centering
    \input{plots/ood}
  \end{minipage}

  \vspace{-5pt}
  \caption{\textbf{Out-of-domain vs.\ in-domain visual support between Cityscapes and ACDC.}
  The \emph{left} plot reports performance on \textbf{ACDC} as we vary the number of visual support images per class from either Cityscapes (out-of-domain, Cityscapes $\rightarrow$ ACDC) or ACDC (in-domain, ACDC $\rightarrow$ ACDC).  
  The \emph{right} plot reports the analogous results when evaluating on \textbf{Cityscapes}, with support drawn from either ACDC (out-of-domain, ACDC$\rightarrow$ Cityscapes) or Cityscapes (in-domain, Cityscapes $\rightarrow$ Cityscapes).  
  In both cases we compare \ours with and without text, and include the corresponding zero-shot baseline.}
  \vspace{-10pt}
  \label{fig:ood}
\end{figure*}

\subsection{Implementation details} 
\label{sec:implementation}

We use OpenCLIP ViT-B/16~\cite{cherti2023reproducible} trained on LAION~\cite{schuhmann2022laion} and apply the MaskCLIP trick~\cite{maskclip}. 
For DINOv3~\cite{simeoni2025dinov3}, we use the public ViT-L/16 checkpoint adapted with dino.txt~\cite{jose2024dinov2meetstextunified} to derive text-aligned patch features, denoted by DINOv3.txt. For SigLIP2~\cite{tschannen2025siglip2multilingualvisionlanguage}, we use the ViT-L/16 variant trained on images of resolution 512×512. For optional region proposals, we use SAM 2.1~\cite{ravi2024sam2segmentimages} Hiera-L with a $32{\times}32$ grid of points (one mask per point) and non-maximum suppression to ensure non-overlap. Pixels not belonging to any mask form an additional separate mask.
For support images we extract dense patch features $X^{i}\!\in\!\mathbb{R}^{n\times d}$ using a sliding window of fixed crop size and stride, down-sample its mask $Y^{i}$ to patch labels $P^{i}\!\in\![0,1]^{n\times C}$ via label aggregation within each patch region. 
At inference, we resize images to a fixed shorter side size, preserving aspect ratio, and extract dense features using a sliding window with fixed crop size and stride. \emph{Textual class features} $\mathbf{t}_c$ are obtained with standard CLIP ImageNet-1k templates, \eg, \texttt{a photo of a \{class\}}~\cite{radford2021learning}. We average the text features across templates for each class. We do not expand class names beyond this.
Regarding \ours, we fix hyperparameters across all experiments to \( k{=}4,\; \tau{=}0.1,\; \beta_{f}{=}1.5,\; \beta_{p}{=}0.2\), and $\Lambda = \{0.9\,, 0.8\,, 0.6\,, 0.4\,, 0.2\,, 0.0\}$. Our model is a linear classifier $g_{\theta}$ trained per test image. We train for $700$ steps with a learning rate $0.02$, optimizing using Adam~\cite{kingma2014adam} in full-batch mode, \ie no mini-batch stochasticity is involved.

\subsection{Details on competitors}
\label{sec:competitors}

For \textbf{\knnclip}~\cite{gui2024knnclip}, we follow the official implementation. To ensure a fair comparison we fix the hyperparameters across datasets and few-shot settings. The values are chosen based on the average test set performance on the considered benchmarks. The w/o text variant of the method arises by applying $T{=} 1.0$ to the confidence threshold they introduce on zero-shot predictions. Please refer to the original paper for additional information.

\textbf{\freeda}~\cite{freeda} generates images and pseudo-labels using a diffusion model, to construct a set of weakly labeled synthetic prototypes similar to our per-image visual class features. In their setup, these synthetic features are indexed via descriptive text features. At test time, the text feature of each test class is used to retrieve the most similar text indexes and thus match that class with similar synthetic prototypes. In our case, each per-image visual class feature already has a ground-truth label, so the correspondence to test classes is known. We therefore remove the retrieval step and assume oracle access to this matching when adapting their method to real visual support.

We fix the hyperparameters of the method in the same way that we do for \knnclip. The w/o text variant is obtained by applying $\beta{=}1.0$ which is the weight used to linearly combine local visual similarities with global textual similarities. Please refer to the original paper for additional information.

In both methods, in the partial text support setup, we replace missing textual class features with the average textual class feature across classes with an available class name. 

\subsection{Details on comparison to offline baselines}
\label{sec:offline_baselines}

In Figure 6 of the main paper we compare \ours with baselines trained in an offline, closed-vocabulary way. To ensure a fair comparison we use OpenCLIP ViT-B/16 features and no mask proposer across all methods. For each method we tune learning rate, batch size, and number of training iterations on an $85{-}15\%$ train–validation split of each support set. We use the Optuna library to get a per-support-set “optimal’’ hyperparameter set, based on validation performance, for all methods.

To motivate this tuning, \autoref{fig:offline_fixed_params} compares \ours and the linear offline classifier on per-image visual class features under two fixed hyperparameter settings. The linear classifier is highly sensitive to this choice: performance can differ by more than 30 mIoU, and even collapses under mismatched settings. In contrast, \ours changes only modestly across the two configurations and stays close to its tuned “optimal’’ curve. This shows that offline baselines need careful hyperparameter tuning to stay competitive, largely because the effective training set size changes drastically between low-shot settings. In contrast, \ours is much more robust to fixed training settings: by retrieving only support features relevant to each test image, it keeps the effective training set size more stable.

\section{Additional experimental results}
\label{sec:add_results}

\subsection{Out-of-domain visual support}
\label{sec:ood}

In \autoref{fig:ood}, we analyze how \ours behaves when the visual support comes from out-of-domain images of the test classes. We use ACDC~\cite{Sakaridis2021ACDC}, which provides annotations for the Cityscapes~\cite{cordts2016cityscapes} class set but is captured under adverse conditions (fog, snow, rain, night). Out-of-domain support is consistently weaker than in-domain support, yet it still yields substantial gains over zero-shot segmentation and continues to improve as we add more visual examples. Remarkably, when evaluating on Cityscapes, \ours improves over the zero-shot baseline even when using ACDC as visual support, despite the fact that many ACDC scenes have ambiguous semantics (\eg, sidewalks completely covered by snow).

\begin{figure}
  \vspace{-5pt}
  \centering
  \input{plots/backbones}
  \vspace{-22pt}
  \caption{\textbf{Backbone comparison.} We compare the performance of \ours using three different vision--language backbones (OpenCLIP ViT-B/16, DINOV3.txt ViT-L/16, and SigLIP2 ViT-L/16), along with the corresponding zero-shot baselines for each one of them.}
  \vspace{-5pt}
  \label{fig:avg_backbones}
\end{figure}
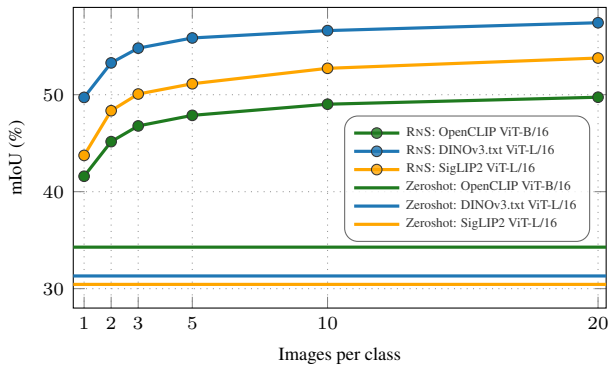

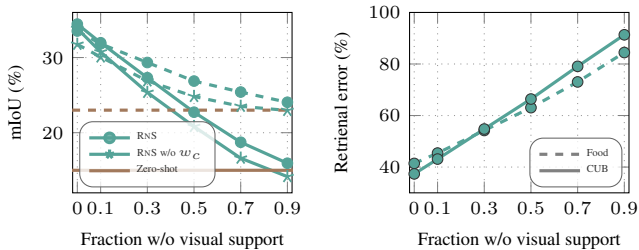
\begin{figure}
  \vspace{-2pt}
  \centering
  \begin{tabular}{@{\hspace{-7pt}}c@{\hspace{1pt}}c@{\hspace{-7pt}}}
  \input{plots/rebuttal/partial_visual_support_finegrained} & \input{plots/rebuttal/partial_visual_support_retrieval_quality} \\
  \end{tabular}
  \vspace{-12pt}
  \caption{\textbf{Partial visual (left) and Retrieval quality (right).}  OpenCLIP (ViT- B/16) + SAM 2.1 for region proposals are used.
  \label{fig:partial_finegrained}}
  \vspace{-12pt}
\end{figure} 

\subsection{Partial visual support on fine-grained datasets.}
\label{sec:partial_finegrained}

In~\autoref{fig:partial_finegrained} left we present partial visual support performance in Food and CUB. On the right plot we present  the retrieval error (percentage of retrieved instances from classes that are not present in each test image). 
Even for high retrieval errors, \ours effectively takes advantage of support examples and outperforms zero-shot. Moreover, the class relevance weights defined in Eq. 8 of the main manuscript suppress the loss of retrieved instances irrelevant to the test image, resulting in a significant improvement in such scenarios (\autoref{fig:partial_finegrained} left - curves w/o $w_c$).

\begin{figure}[t]
  \vspace{-5pt}
  \centering
  \input{plots/offline_fixed_params.tex}
  \vspace{-10pt}
  \caption{\textbf{Comparison to offline baseline with and w/o hyperparameter tuning}. We compare \ours against the offline linear classifier trained on per image visual class features on VOC. We include the curves presented in Figure 6 that use hyperparameter tuning per $B$ and support seed (noted as “optimal’’). We report the performance of both methods on two hyperparameter configurations: \emph{hyperparameter} $\mathbf{set~1}$ which corresponds to an optimal set of hyperparameters of the linear classifier for $B = 1$, and \emph{hyperparameter} $\mathbf{set~2}$ which corresponds to an optimal set of hyperparameters of the linear classifier for $B = 20$.}
  \label{fig:offline_fixed_params}
\end{figure}
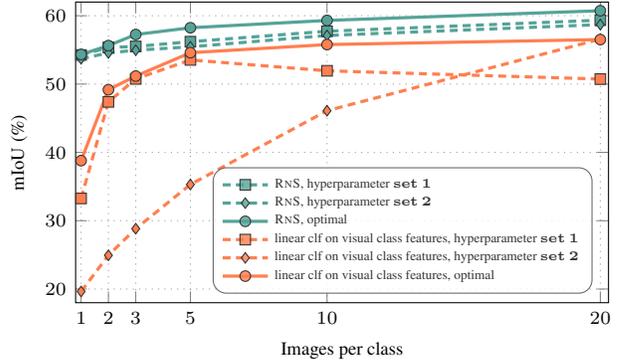

\begin{table}
    \centering
    \begin{subtable}{\linewidth}
        \centering
        \resizebox{\linewidth}{!}{%
        \begin{tabular}{lccccccccc}
            \toprule
            \textbf{Method} 
                & \multicolumn{3}{c}{\textbf{10\% unseen}} 
                & \multicolumn{3}{c}{\textbf{50\% unseen}} 
                & \multicolumn{3}{c}{\textbf{90\% unseen}} \\
            \cmidrule(lr){2-4}\cmidrule(lr){5-7}\cmidrule(lr){8-10}
                & \textbf{S} & \textbf{U} & \textbf{A}
                & \textbf{S} & \textbf{U} & \textbf{A}
                & \textbf{S} & \textbf{U} & \textbf{A} \\
            \midrule
            Zero-shot         & 52.13 & 59.61 & 52.85 & 56.24 & 49.12 & 52.85 & 69.08 & 50.14 & 52.85 \\
            \ours\ \emph{(Partial)} & 72.55 & 62.80 & 71.63 & 76.39 & 50.01 & 63.83 & 73.73 & 49.54 & 52.04 \\
            \ours\ \emph{(Full)}    & 71.83 & 77.89 & 72.41 & 77.42 & 66.89 & 72.41 & 88.98 & 69.65 & 72.41 \\
            \bottomrule
        \end{tabular}%
        }
        \subcaption{VOC}
    \end{subtable}
    
    \vspace{0.75em}
    
    \begin{subtable}{\linewidth}
        \centering
        \resizebox{\linewidth}{!}{%
        \begin{tabular}{lccccccccc}
            \toprule
            \textbf{Method} 
                & \multicolumn{3}{c}{\textbf{10\% unseen}} 
                & \multicolumn{3}{c}{\textbf{50\% unseen}} 
                & \multicolumn{3}{c}{\textbf{90\% unseen}} \\
            \cmidrule(lr){2-4}\cmidrule(lr){5-7}\cmidrule(lr){8-10}
                & \textbf{S} & \textbf{U} & \textbf{A}
                & \textbf{S} & \textbf{U} & \textbf{A}
                & \textbf{S} & \textbf{U} & \textbf{A} \\
            \midrule
            Zero-shot         & 21.87 & 31.49 & 22.83 & 21.82 & 23.85 & 22.83 & 22.06 & 22.92 & 22.83 \\
            \ours\ \emph{(Partial)} & 34.30 & 30.19 & 33.89 & 32.09 & 22.49 & 27.29 & 27.10 & 22.56 & 23.01 \\
            \ours\ \emph{(Full)}    & 34.95 & 36.05 & 35.06 & 35.47 & 34.65 & 35.06 & 36.18 & 34.94 & 35.06 \\
            \bottomrule
        \end{tabular}%
        }
        \subcaption{ADE20K}
    \end{subtable}

    \vspace{-5pt}
    \caption{\textbf{Partial seen/unseen classes.} We report the performance of \ours under varying fractions of unseen classes (classes w/o visual support) on VOC (a) and ADE20K (b). Columns show mean IoU on seen (S), \ie classes with visual support, unseen (U), \ie class that lack visual support, and all (A), \ie the entire class set. We report the zero-shot (no class visually supported) and \ours with full visual support (all classes visually supported) on the same class sets as a reference.}
    \vspace{-10pt}
    \label{fig:unseen}
\end{table}

\begin{table*}[t]
\vspace{-10pt}
\centering
\renewcommand{\arraystretch}{1.15}
\setlength{\tabcolsep}{4pt}
\resizebox{\linewidth}{!}{%
\input{tab/soup_suppl}

}
\vspace{-5pt}
\caption{\textbf{OVS \vs fully supervised segmentation.} \emph{Fully Supervised}: best method picked per dataset. $^{*}$: self-evaluated, $^{\dag}$: from CAT-Seg. Mask2Former numbers from DINOv3~\cite{simeoni2025dinov3} paper. \emph{Domain}: using annotations from each training set. \emph{Annot:} the number of pixel-level annotated images that each method uses. For \emph{Domain} we report the ADE  annotations. 
\label{tab:ovseg_comparison_full}
\vspace{-5pt}}
\end{table*}

\subsection{Backbone comparison}
\label{sec:backbones}

\autoref{fig:avg_backbones} compares \ours across three vision-language backbones. DINOv3.txt ViT-L/16 achieves the highest mIoU and benefits the most from additional visual support, while SigLIP2 ViT-L/16 follows closely. OpenCLIP ViT-B/16 performs consistently lower but still improves steadily with more support images. In all cases, \ours significantly surpasses the respective zero-shot baselines, showing that the method is effective across backbones of different capacity and training regimes.

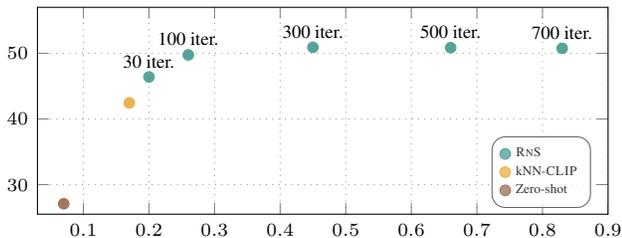
\begin{figure}[h!]
  \centering
  \input{./plots/rebuttal/performance_time.tex}
  \vspace{-12pt}
  \caption{\textbf{Average performance (mIoU) \vs inference time (s).} DINOv3.txt, patch-level; $B{=}5$. Avg. on VOC, ADE, Stuff. A single NVIDIA $A100$ GPU is used.}
  \label{fig:efficiency_vs_performance}
  \vspace{-9pt}
\end{figure}

\subsection{Performance on seen/unseen classes}
\label{sec:unseen}

In \autoref{fig:unseen} we report the performance of our partial visual support setup by separately measuring accuracy on seen classes (those with visual support) and unseen classes (those without support). The tables show that, when only a small fraction of classes is unseen, our partially supported variant performs on seen classes almost identically to \ours with full support. As the proportion of unseen classes increases, a widening gap appears. This is expected: as the number of unseen classes increases so does the number of false-positives for unseen classes, which impacts the performance on seen ones too.

At the same time, performance on unseen classes remains close to the zero-shot baseline across all settings. This confirms that \ours does not degrade zero-shot behavior for classes without visual support—when support is absent, the model naturally falls back to its zero-shot capabilities.

\subsection{Full \textbf{\textit{OVS vs fully supervised}} table}
\label{sec:ovss_vs_supervised}

\autoref{tab:ovseg_comparison_full} is a complete version of Table 2, presented in the main paper. We augment the table with a column that mentions the backbones used within each method. We include additional OVS methods from the three categories mentioned in the main manuscript. We also include the individual \emph{fully supervised} methods reported on each dataset.

\begin{table}[t]
  \vspace{0pt}
  \centering
  \resizebox{\linewidth}{!}{%
    \input{tab/ablation_k}
  }
  \vspace{-4pt}
  \caption{\textbf{Ablations of the k-NN retrieval hyperparameter $K$ in \ours.}
  We report average mIoU across the considered datasets for three different numbers of available support images per class ($B=1$, $B=5$, $B=10$).
  The row with $K=4$ (highlighted) corresponds to the configuration used in \ours, and blue numbers denote the difference to this row in the same column.}
  \label{tab:ablation_k}
  \vspace{-4pt}
\end{table}

\subsection{Efficiency vs Runtime comparison.}
\label{runtimes}

In \autoref{fig:efficiency_vs_performance} we compare the performance and inference time of \ours, using different number of training iterations, with feedforward methods like \knnclip and zero-shot baseline. We observe that the performance of \ours is robust under less iterations, while efficiency becomes comparable to feedforward \knnclip. Thus, the overhead of \ours can be reduced while still achieving quite a performance boost.

\subsection{Ablation on K in kNN retrieval}
\label{sec:ablation_k}

In \autoref{tab:ablation_k}, we study the effect of the kNN retrieval hyperparameter $K$.
Using a single neighbor ($K=1$) per patch/region leads to clearly lower performance across all budgets $B$, confirming that aggregating information from multiple retrieved exemplars is beneficial. For $K \geq 4$, performance varies only marginally, with $K=4$--$16$ achieving very similar mIoU, indicating that our method is robust to the exact choice of $K$ once it is larger than $1$. We therefore use $K=4$ as our default setting.

\section{Qualitative results}
\label{sec:qual}

\subsection{SAM mask vs patch-level predictions}
\label{sec:sam_vs_patch}

\begin{figure*}[t]
\centering
\setlength{\tabcolsep}{2pt} 
\renewcommand{\arraystretch}{2.5} 
\begin{tabular}{c c c c c c}
\textbf{\emph{Image}} &
\textbf{\emph{GT}} &
\textbf{\emph{Patch feats}} &
\textbf{\ours \emph{+ Patch}} &
\textbf{\emph{SAM mask feats}} &
\textbf{\ours \emph{+ SAM}} \\
\includegraphics[width=0.16\textwidth]{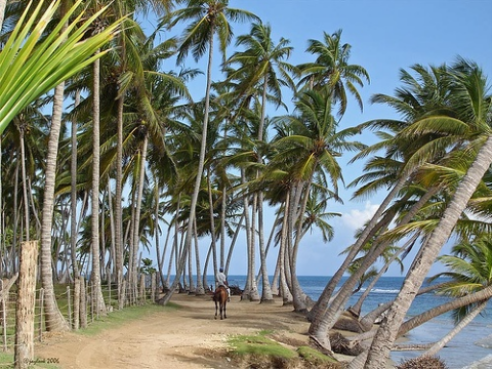} &
\includegraphics[width=0.16\textwidth]{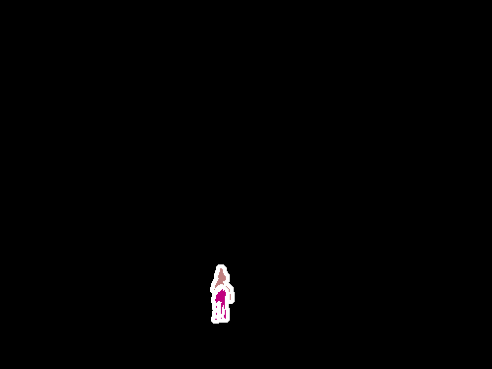} &
\includegraphics[width=0.16\textwidth]{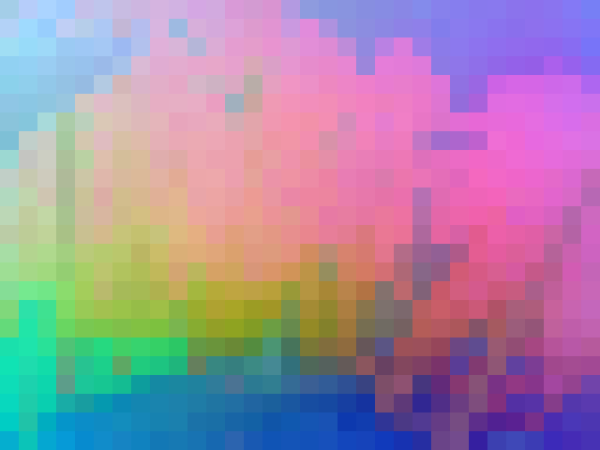} &
\includegraphics[width=0.16\textwidth]{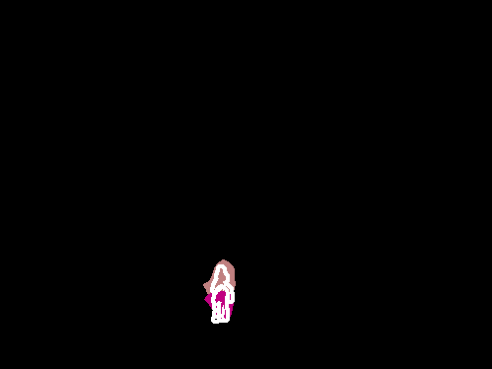} &
\includegraphics[width=0.16\textwidth]{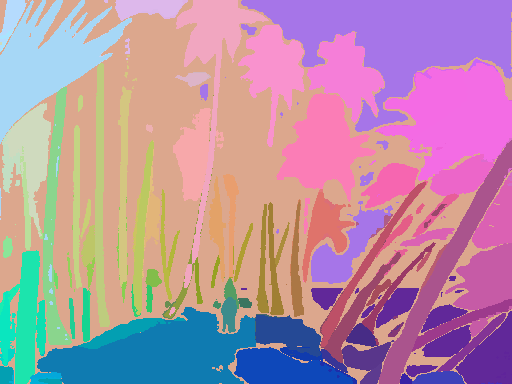} &
\includegraphics[width=0.16\textwidth]{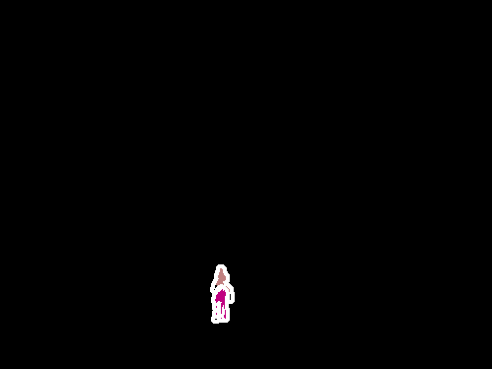} \\
\includegraphics[width=0.16\textwidth]{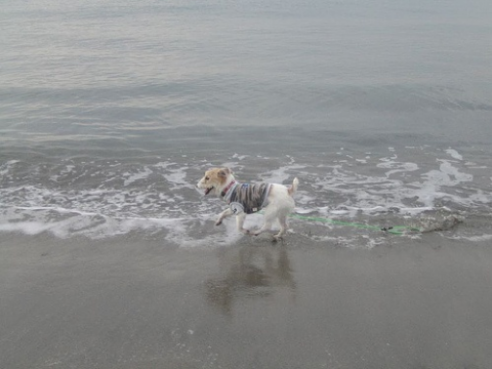} &
\includegraphics[width=0.16\textwidth]{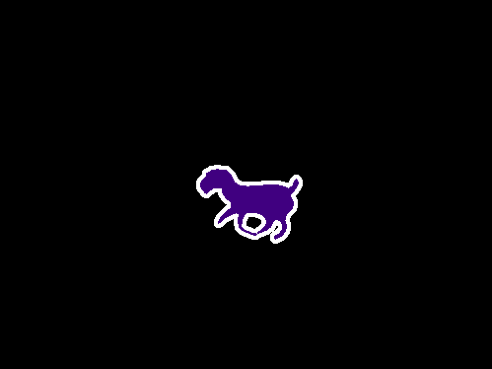} &
\includegraphics[width=0.16\textwidth]{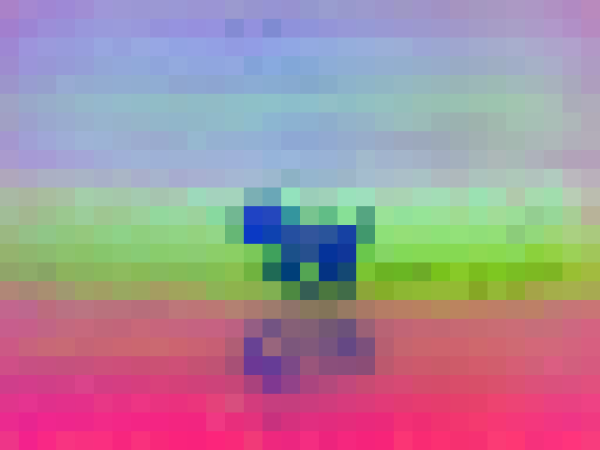} &
\includegraphics[width=0.16\textwidth]{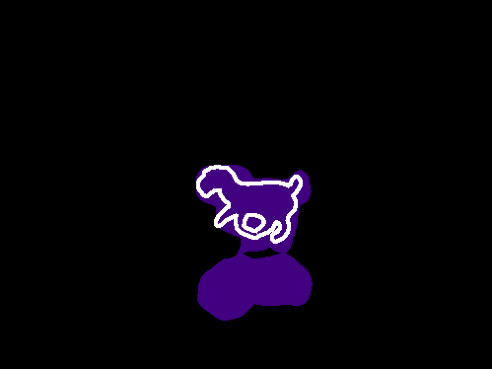} &
\includegraphics[width=0.16\textwidth]{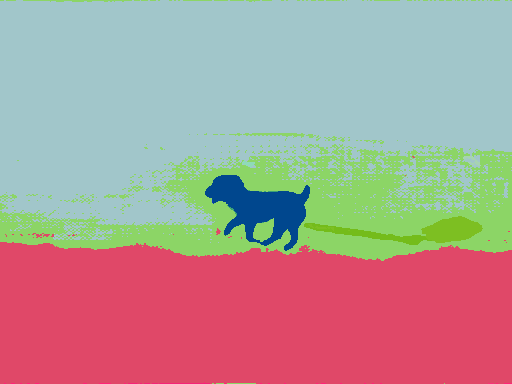} &
\includegraphics[width=0.16\textwidth]{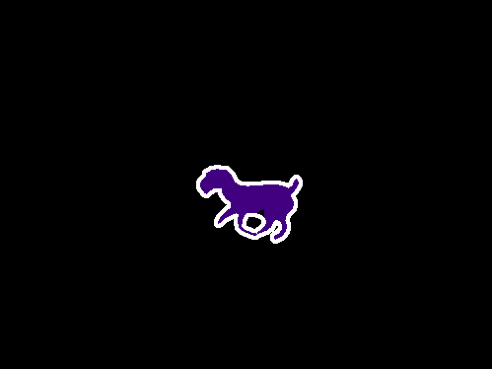} \\
\includegraphics[width=0.16\textwidth]{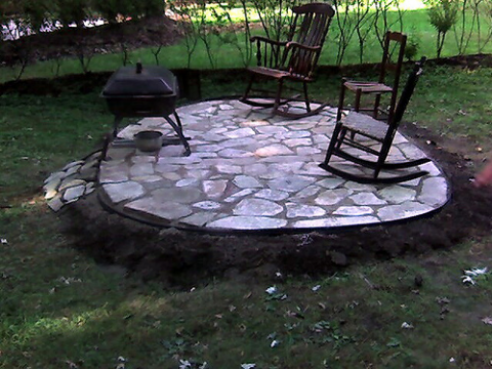} &
\includegraphics[width=0.16\textwidth]{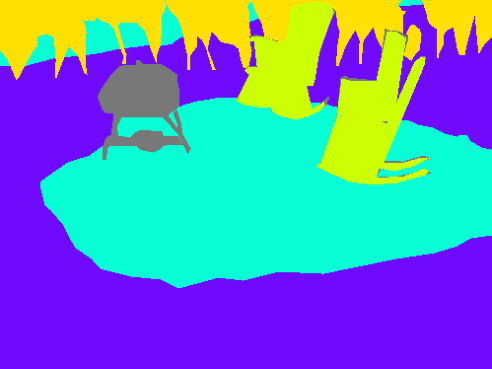} &
\includegraphics[width=0.16\textwidth]{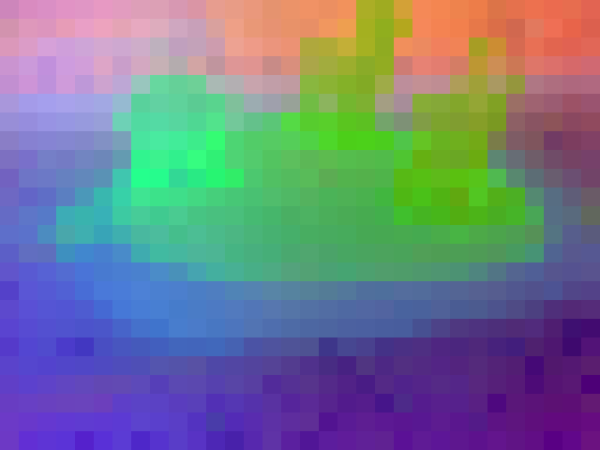} &
\includegraphics[width=0.16\textwidth]{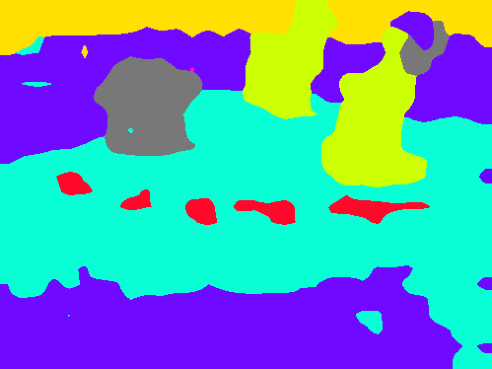} &
\includegraphics[width=0.16\textwidth]{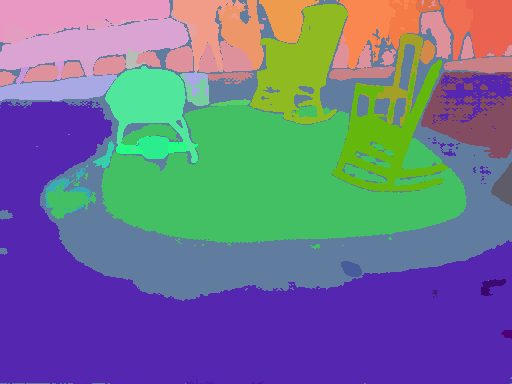} &
\includegraphics[width=0.16\textwidth]{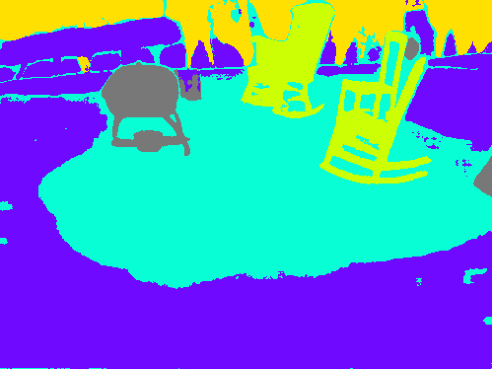} \\
\includegraphics[width=0.16\textwidth]{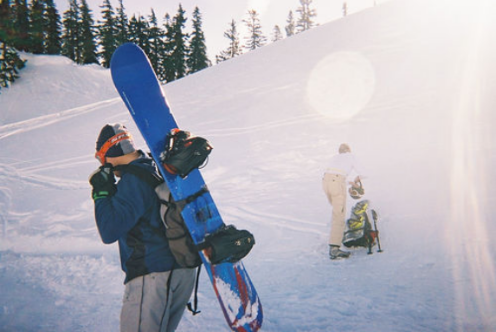} &
\includegraphics[width=0.16\textwidth]{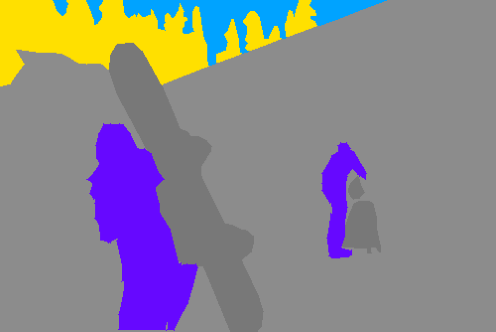} &
\includegraphics[width=0.16\textwidth]{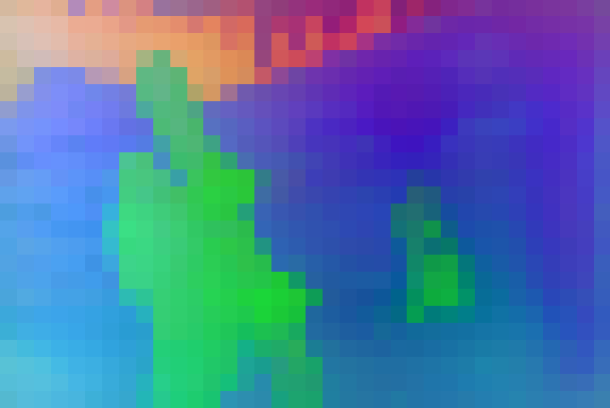} &
\includegraphics[width=0.16\textwidth]{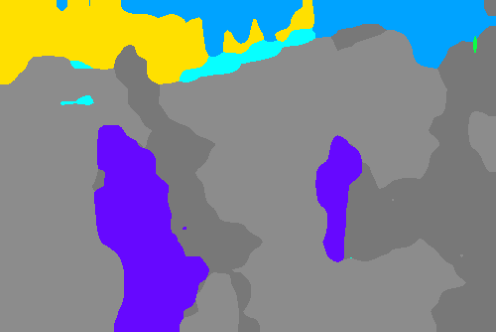} &
\includegraphics[width=0.16\textwidth]{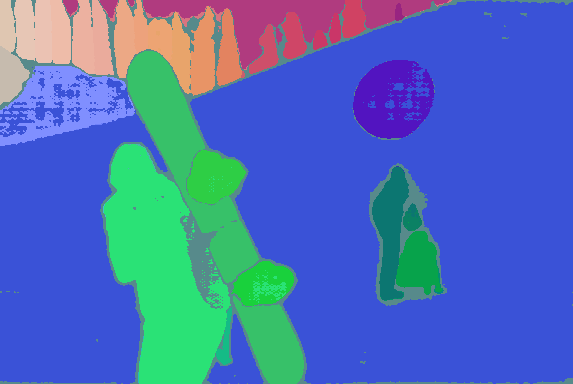} &
\includegraphics[width=0.16\textwidth]{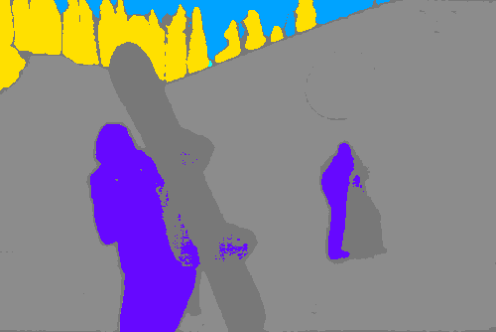} \\
\includegraphics[width=0.16\textwidth]{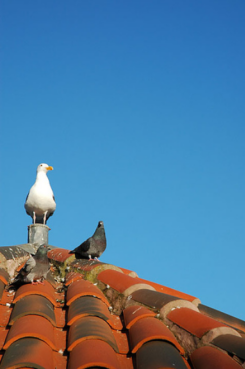} &
\includegraphics[width=0.16\textwidth]{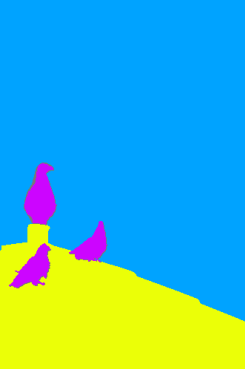} &
\includegraphics[width=0.16\textwidth]{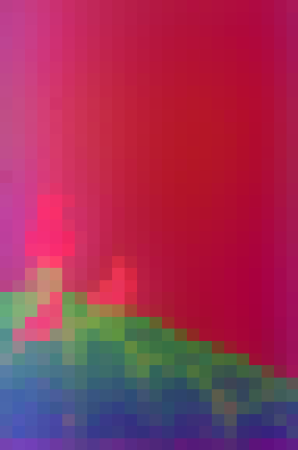} &
\includegraphics[width=0.16\textwidth]{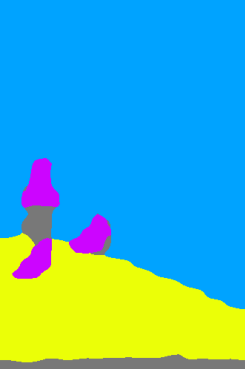} &
\includegraphics[width=0.16\textwidth]{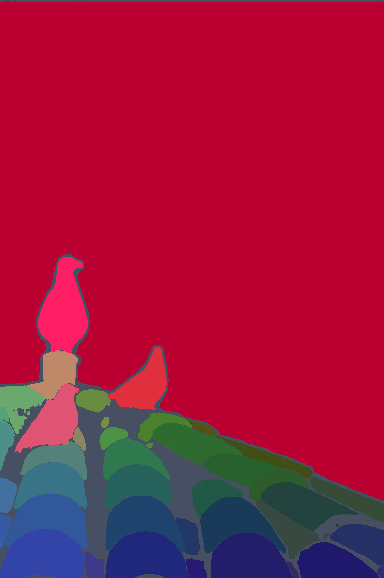} &
\includegraphics[width=0.16\textwidth]{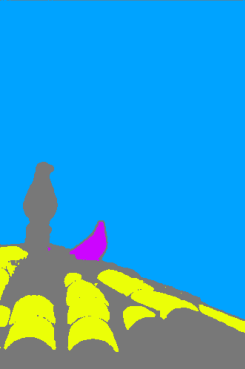} \\
\includegraphics[width=0.16\textwidth]{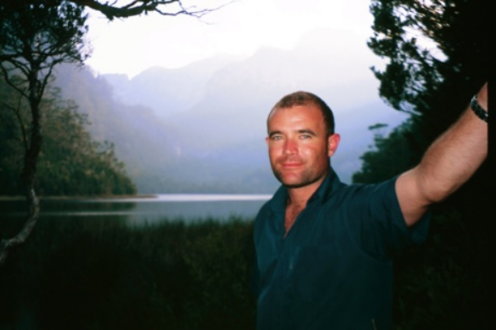} &
\includegraphics[width=0.16\textwidth]{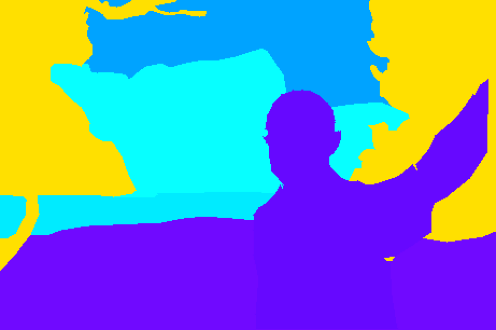} &
\includegraphics[width=0.16\textwidth]{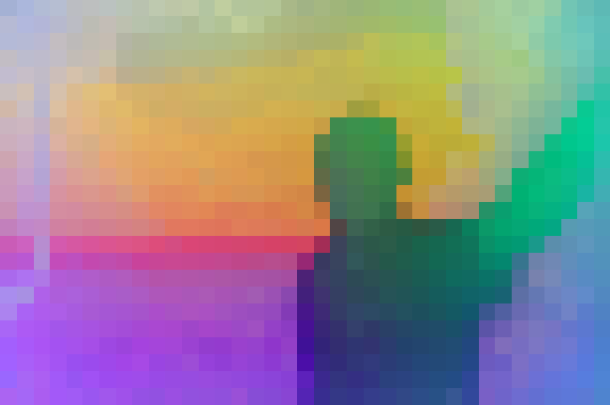} &
\includegraphics[width=0.16\textwidth]{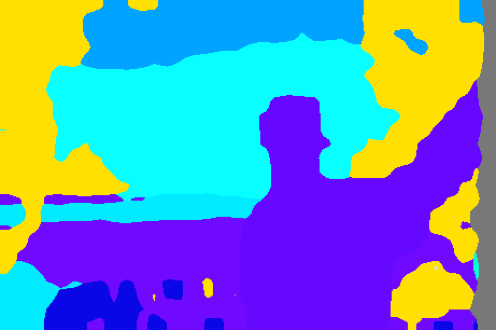} &
\includegraphics[width=0.16\textwidth]{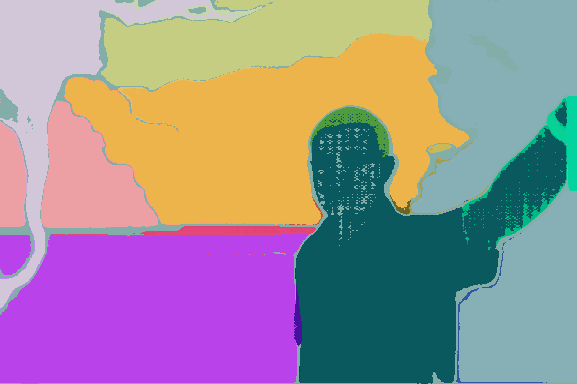} &
\includegraphics[width=0.16\textwidth]{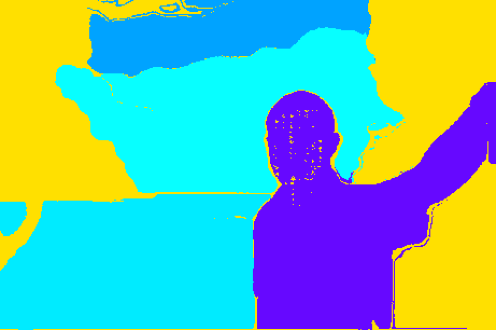} \\
\includegraphics[width=0.16\textwidth]{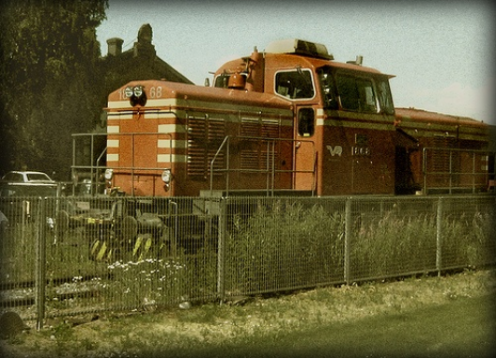} &
\includegraphics[width=0.16\textwidth]{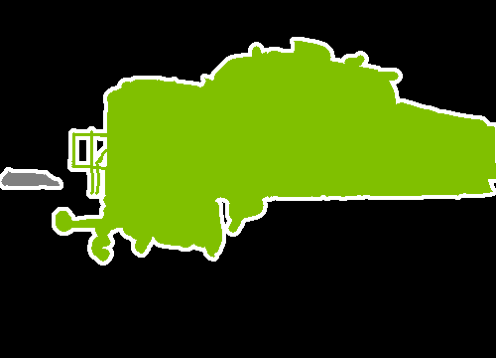} &
\includegraphics[width=0.16\textwidth]{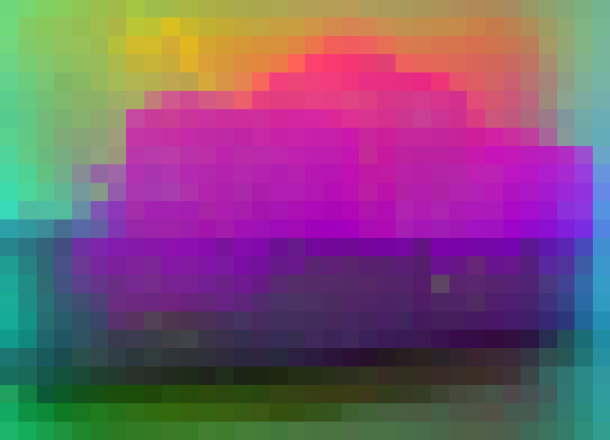} &
\includegraphics[width=0.16\textwidth]{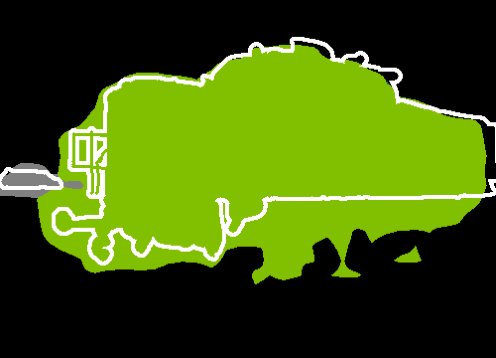} &
\includegraphics[width=0.16\textwidth]{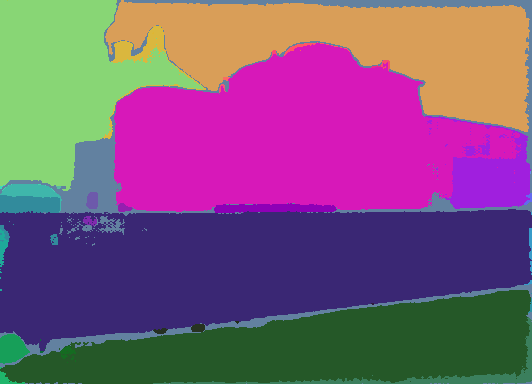} &
\includegraphics[width=0.16\textwidth]{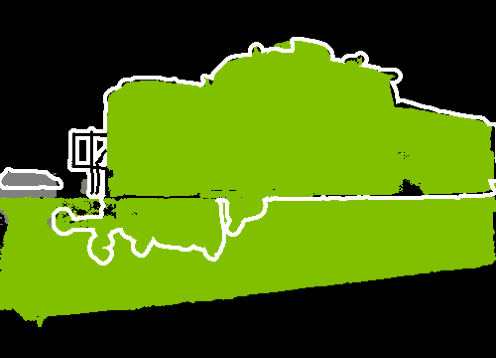} \\
\end{tabular}
\caption{\textbf{Qualitative comparison of patch-level vs.\ mask-level segmentation.} Each row shows the input image, ground-truth mask, the PCA projection of either patch features or SAM mask features (average VLM patch features on top of SAM mask), and the corresponding predictions of \ours when applied on patches or on region proposals. Both variants use DINOv3.txt features~\cite{simeoni2025dinov3}, and SAM 2.1 is used as the mask proposal generator~\cite{ravi2024sam2segmentimages}.
}
\label{fig:sam_vs_patch}
\end{figure*}

In~\autoref{fig:sam_vs_patch} we qualitatively compare two variants of \ours: one where the test-time linear layer is applied directly to patch-level features (\ours + Patch), and one where it operates on mask embeddings obtained by pooling features within SAM2.1 mask proposals (\ours + SAM). As expected, SAM2.1 proposals follow object boundaries much more closely than fixed patches, which yields noticeably sharper segmentation masks. Aggregating features within each mask also provides a form of spatial denoising, suppressing spurious patch-level predictions. SAM further exhibits a stronger notion of objectness, grouping together parts of the same object even under challenging appearance changes (\eg, shadows in the second row).

At the same time, SAM does not always respect the semantic granularity required by the task and can over- or under-segment regions. In the bottom three rows, this leads to masks that merge distinct semantic regions or split single objects into multiple segments, introducing ambiguity despite the improved alignment with image structure.

\subsection{Unimodal vs.\ multimodal support} 
\label{sec:qual_modalities}
\begin{figure*}[t]
\centering
\setlength{\tabcolsep}{2pt}      
\renewcommand{\arraystretch}{2.5} 
\begin{tabular}{c c c c c c c}
\textbf{\emph{Image}} &
\textbf{\emph{GT}} &
\textbf{\emph{Zero-shot}} &
\textbf{\freeda} &
\textbf{\knnclip} &
\textbf{\ours \emph{w/o text}} &
\textbf{\ours} \\
\includegraphics[width=0.14\textwidth]{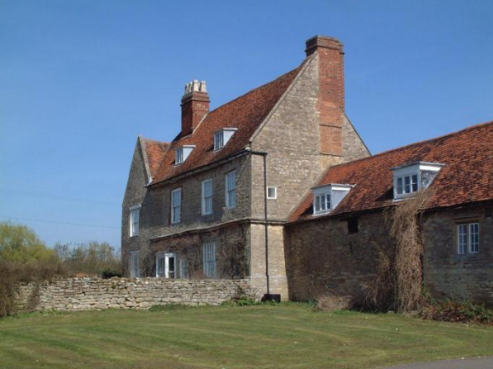} &
\includegraphics[width=0.14\textwidth]{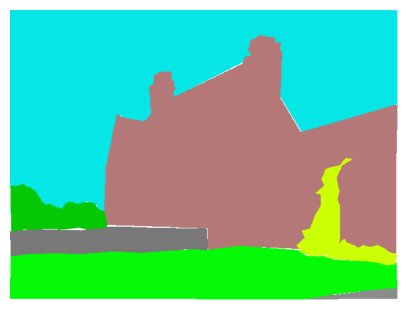} &
\includegraphics[width=0.14\textwidth]{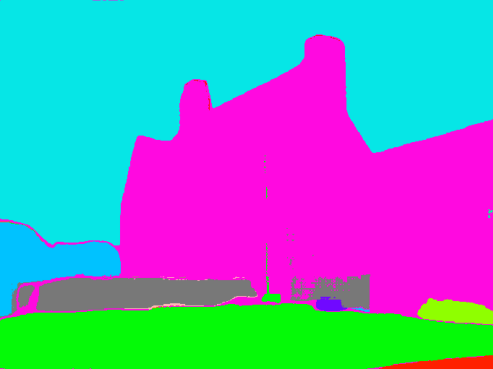} &
\includegraphics[width=0.14\textwidth]{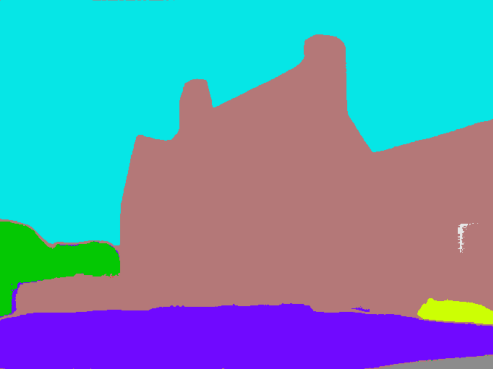} &
\includegraphics[width=0.14\textwidth]{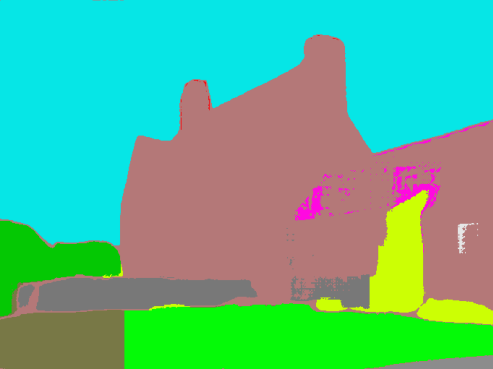} &
\includegraphics[width=0.14\textwidth]{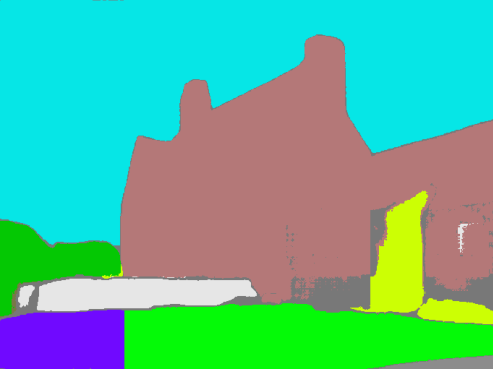} &
\includegraphics[width=0.14\textwidth]{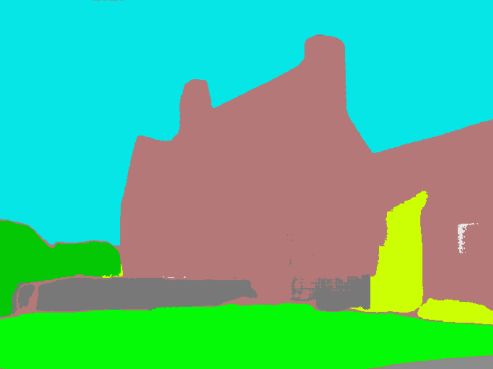} \\
\includegraphics[width=0.14\textwidth]{images/img5.png} &
\includegraphics[width=0.14\textwidth]{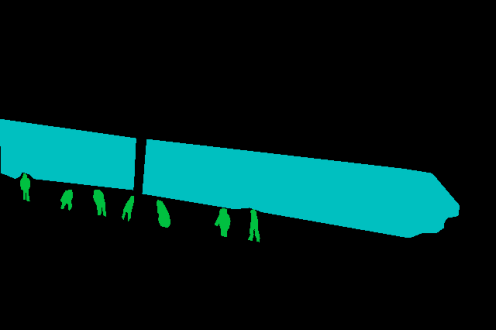} &
\includegraphics[width=0.14\textwidth]{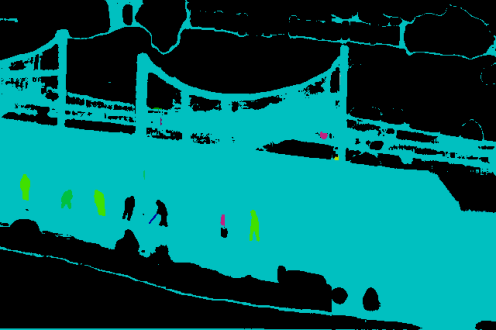} &
\includegraphics[width=0.14\textwidth]{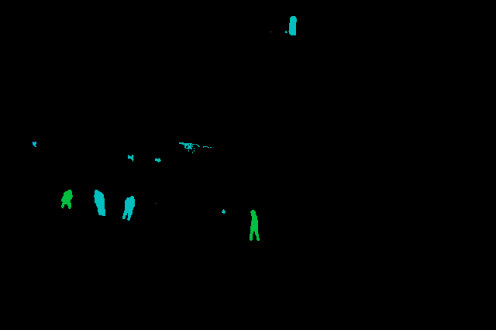} &
\includegraphics[width=0.14\textwidth]{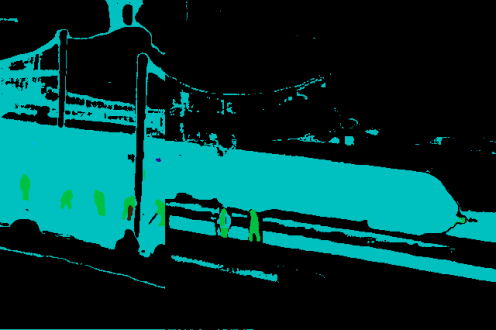} &
\includegraphics[width=0.14\textwidth]{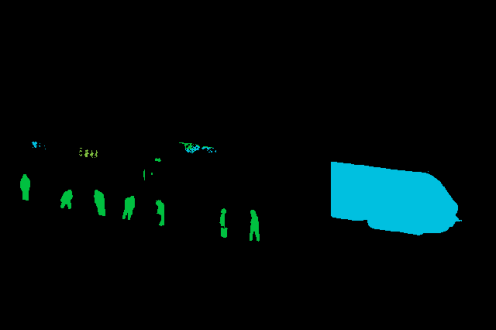} &
\includegraphics[width=0.14\textwidth]{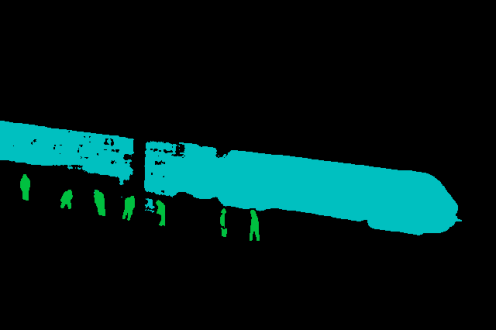} \\
\includegraphics[width=0.14\textwidth]{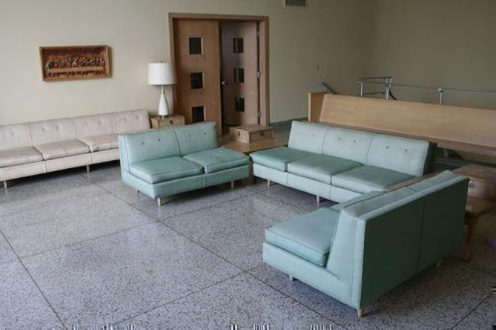} &
\includegraphics[width=0.14\textwidth]{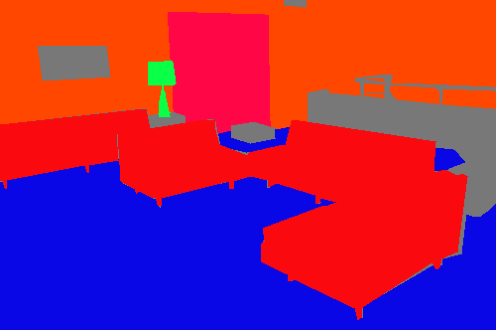} &
\includegraphics[width=0.14\textwidth]{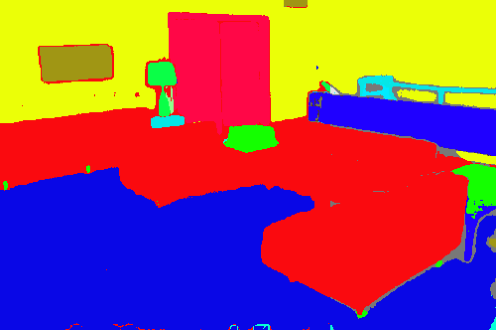} &
\includegraphics[width=0.14\textwidth]{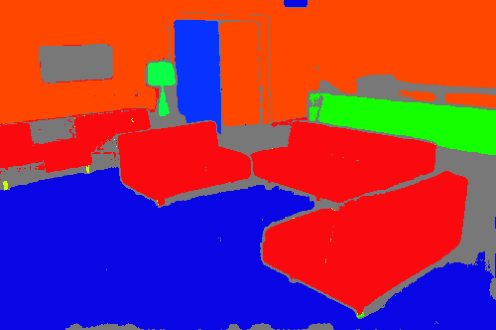} &
\includegraphics[width=0.14\textwidth]{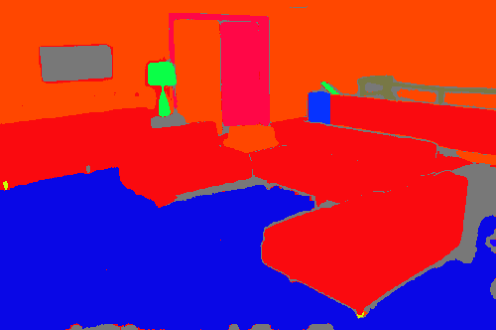} &
\includegraphics[width=0.14\textwidth]{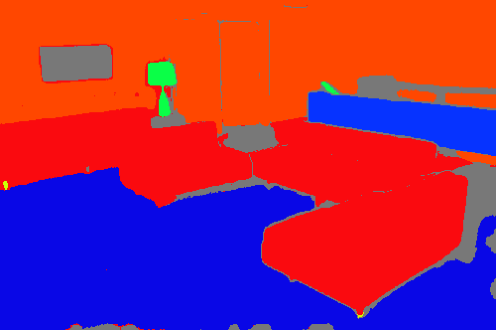} &
\includegraphics[width=0.14\textwidth]{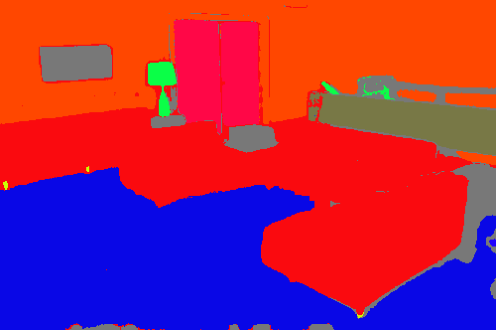} \\
\includegraphics[width=0.14\textwidth]{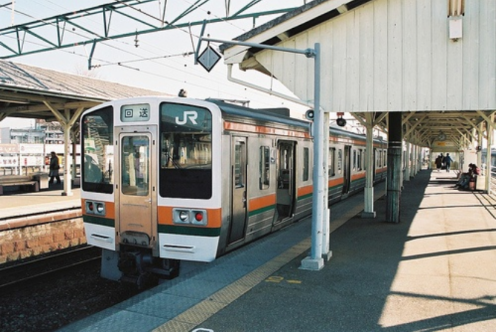} &
\includegraphics[width=0.14\textwidth]{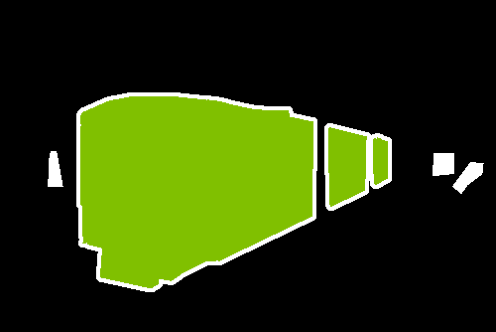} &
\includegraphics[width=0.14\textwidth]{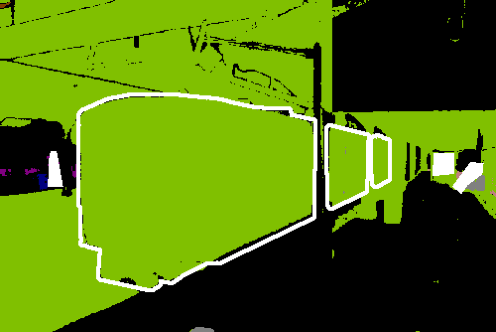} &
\includegraphics[width=0.14\textwidth]{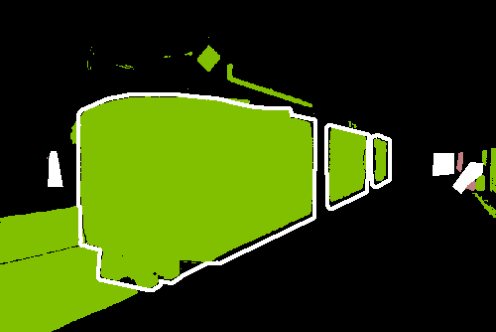} &
\includegraphics[width=0.14\textwidth]{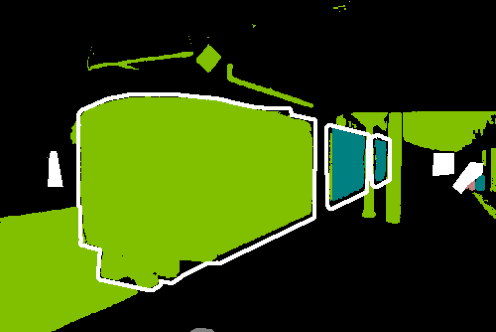} &
\includegraphics[width=0.14\textwidth]{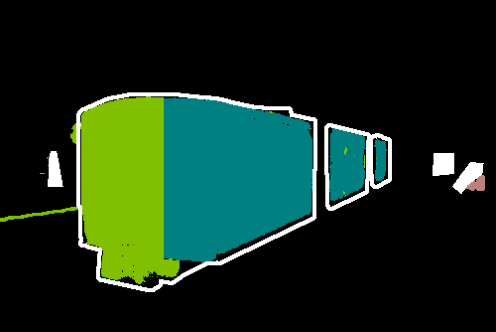} &
\includegraphics[width=0.14\textwidth]{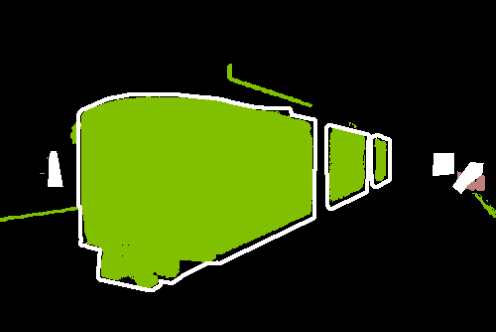} \\
\includegraphics[width=0.14\textwidth]{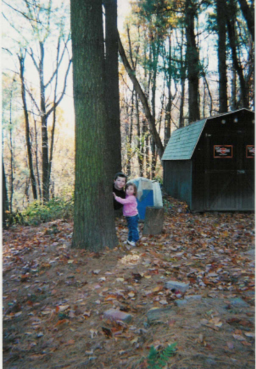} &
\includegraphics[width=0.14\textwidth]{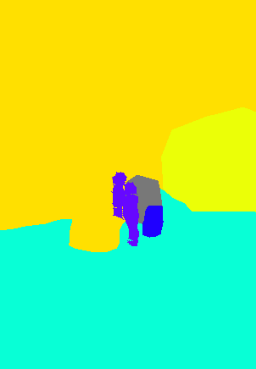} &
\includegraphics[width=0.14\textwidth]{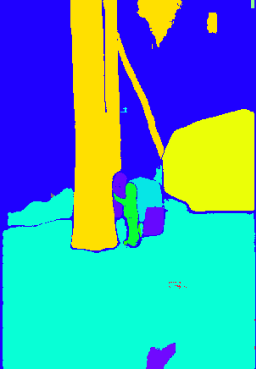} &
\includegraphics[width=0.14\textwidth]{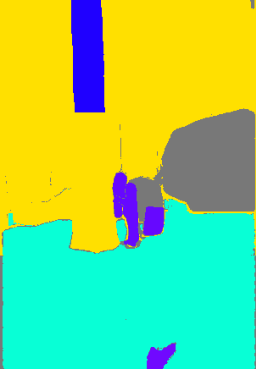} &
\includegraphics[width=0.14\textwidth]{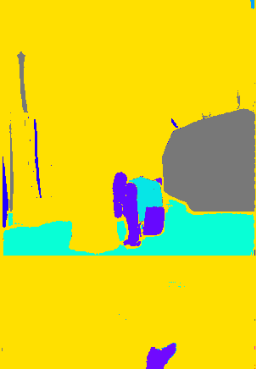} &
\includegraphics[width=0.14\textwidth]{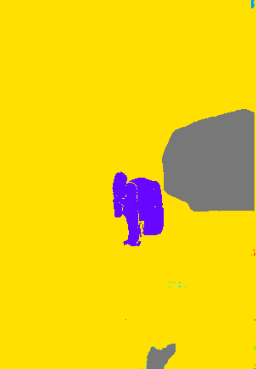}
&
\includegraphics[width=0.14\textwidth]{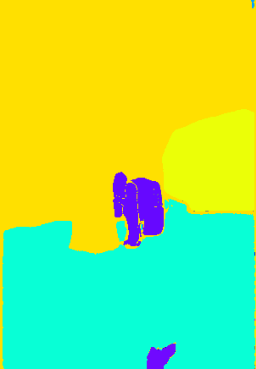} \\
\includegraphics[width=0.14\textwidth]{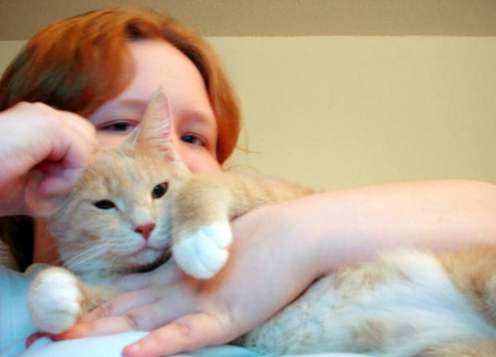} &
\includegraphics[width=0.14\textwidth]{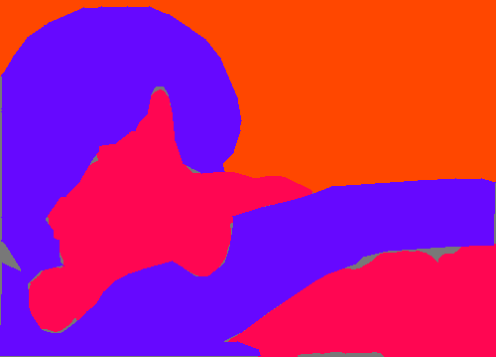} &
\includegraphics[width=0.14\textwidth]{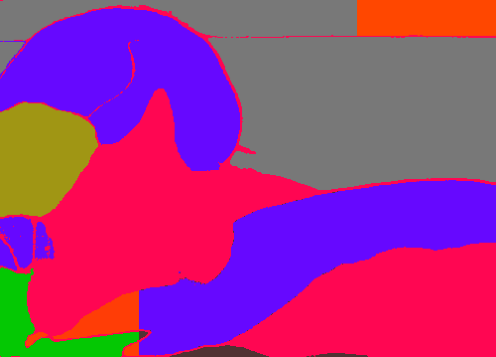} &
\includegraphics[width=0.14\textwidth]{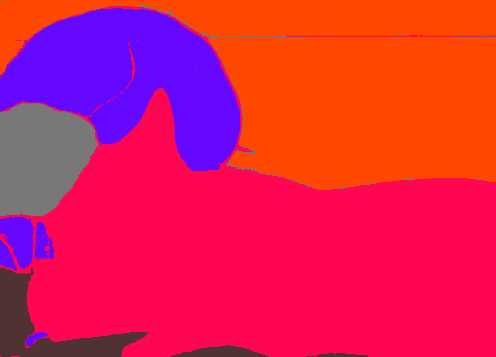} &
\includegraphics[width=0.14\textwidth]{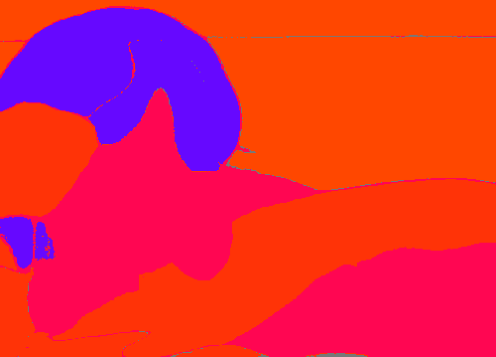} &
\includegraphics[width=0.14\textwidth]{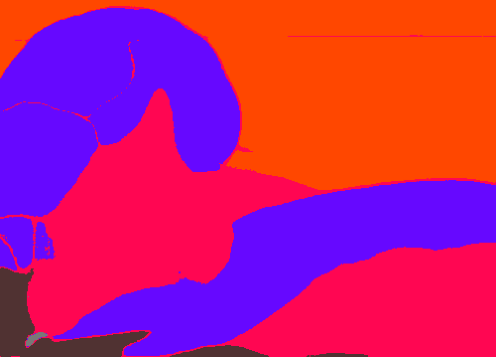} &
\includegraphics[width=0.14\textwidth]{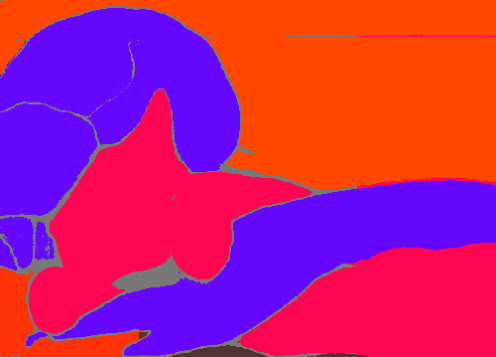} \\
\includegraphics[width=0.14\textwidth]{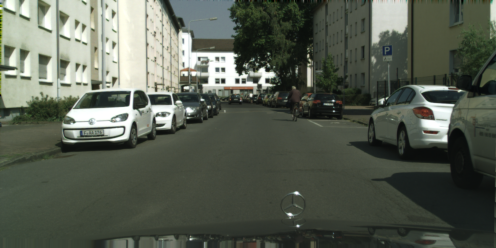} &
\includegraphics[width=0.14\textwidth]{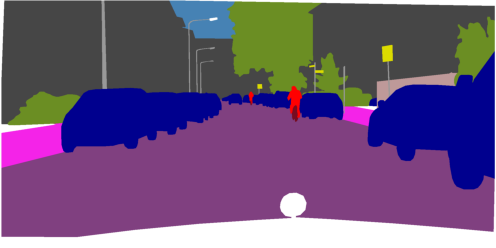} &
\includegraphics[width=0.14\textwidth]{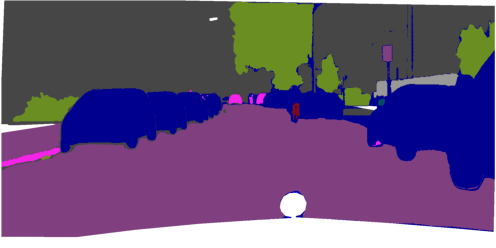} &
\includegraphics[width=0.14\textwidth]{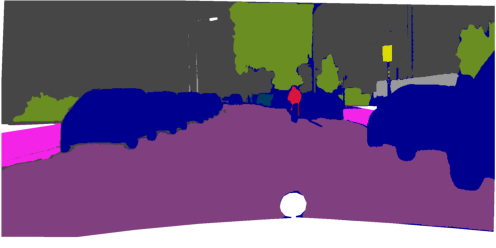} &
\includegraphics[width=0.14\textwidth]{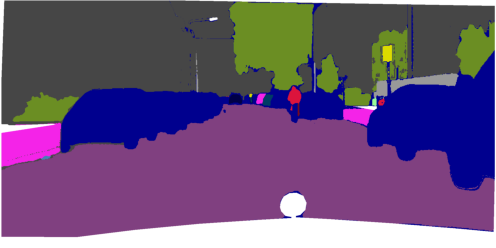} &
\includegraphics[width=0.14\textwidth]{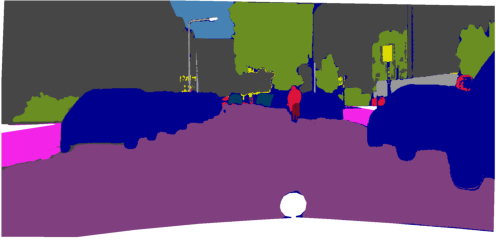} &
\includegraphics[width=0.14\textwidth]{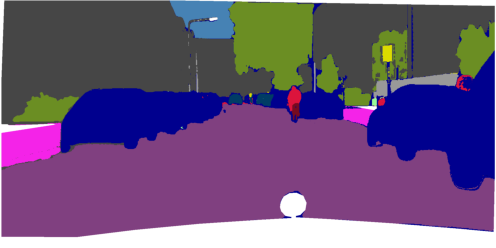} \\
\end{tabular}
\caption{\textbf{Qualitative comparisons} between zero-shot baseline, \freeda, \knnclip and \ours with and without class name information. All visually supported methods use one support image per class ($B{=}1$). All methods use OpenCLIP ViT-B/16 as VLM features~\cite{simeoni2025dinov3} and SAM 2.1 as region proposal generator~\cite{ravi2024sam2segmentimages}.}
\label{tab:qualitative}
\end{figure*}

In ~\autoref{fig:ovs_suppl} and \autoref{tab:qualitative} we qualitatively compare methods that rely only on text (Zero-shot), only on visual support (\ours\emph{ w/o text}), or on both modalities (\ours). Using only class names often produces semantically ambiguous labels when several categories share similar appearance or context (\eg, house vs.\ building in the first row, wall vs.\ building in the third row of Fig.~\ref{tab:qualitative}). Conversely, relying only on visual exemplars can confuse contextually similar objects (\eg, train vs.\ bus in the fourth row). \ours leverages both textual and visual support: text provides a semantic prior that separates related categories, while visual examples anchor this prior to the image appearance, leading to more accurate segmentations across diverse scenes.

\subsection{Comparisons with visually supported methods} 
\label{sec:qual_competitors}
In ~\autoref{tab:qualitative} we additionally compare \ours to \freeda and \knnclip, which also use visual exemplars but rely on fixed, handcrafted fusion of text and visual cues. Such rigid fusion often struggles when the modalities conflict or when one is unreliable. In contrast, \ours learns how to fuse text and visual support, adapting to each image and yielding cleaner boundaries and fewer semantic confusions.

\subsection{Personalized segmentation}
\label{personalization}

\begin{figure*}[t]
    \centering
    \input{fig/personalization_memory}
    \caption{\textbf{Visual support sets for personalized segmentation.} \ours used for personalized segmentation using OpenCLIP ViT-B/16 features and SAM 2.1 as region proposer. Initially, visual support includeds images in \emph{support before} and is later expanded to \emph{support after}, with \emph{support before} $\subset$ \emph{support after}. \textbf{{\color{personalgreen}Green:}} personalized instance, \textbf{{\color{personalred}red:}} generic class, and \textbf{{\color{black}black:}} background. 
  Images from PODS~\cite{sundaram2025personalized}, i-CIR~\cite{psomas2025instancelevel}, and self-collected.}
    \label{fig:personalization_memory}
\end{figure*}
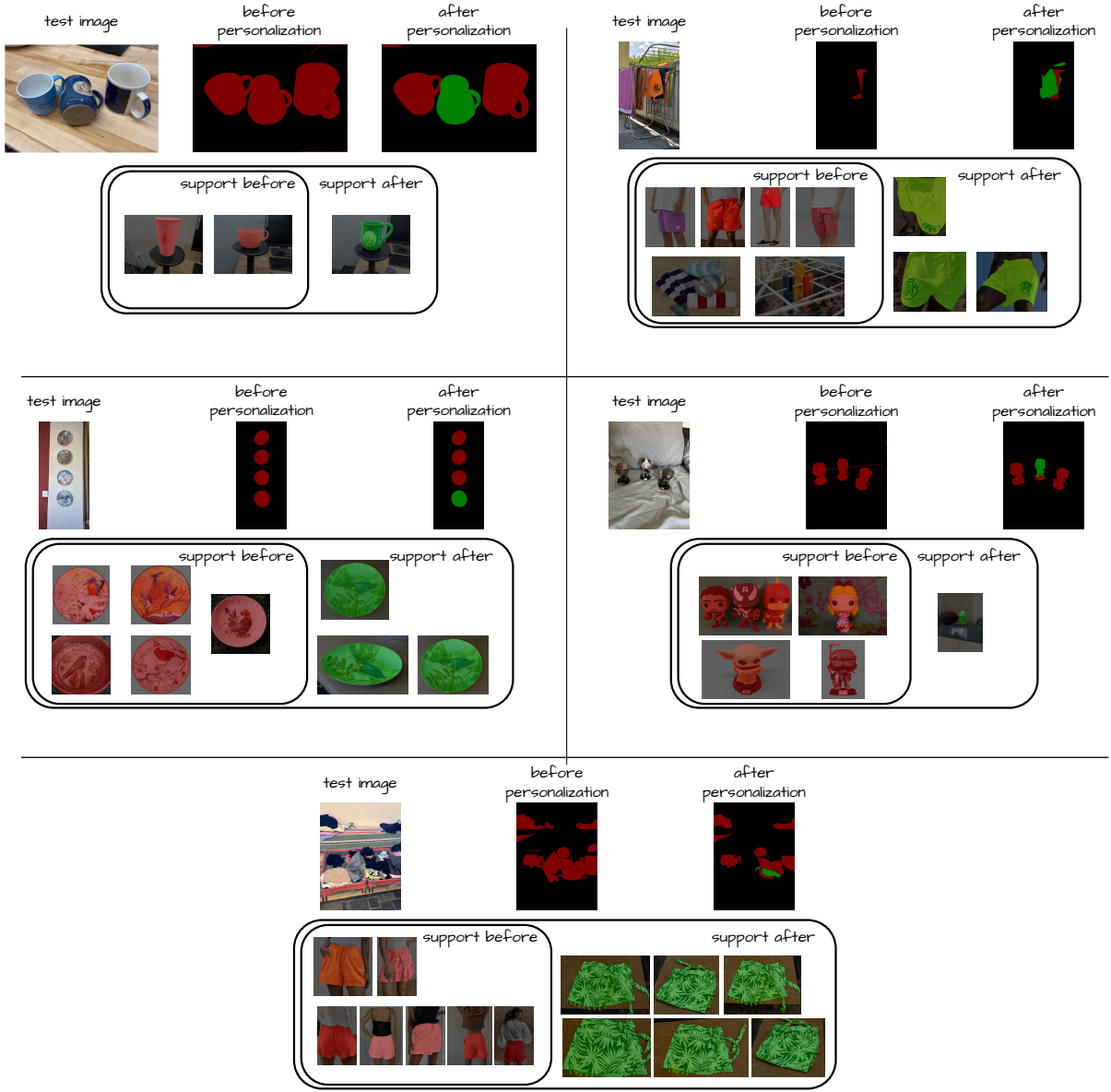

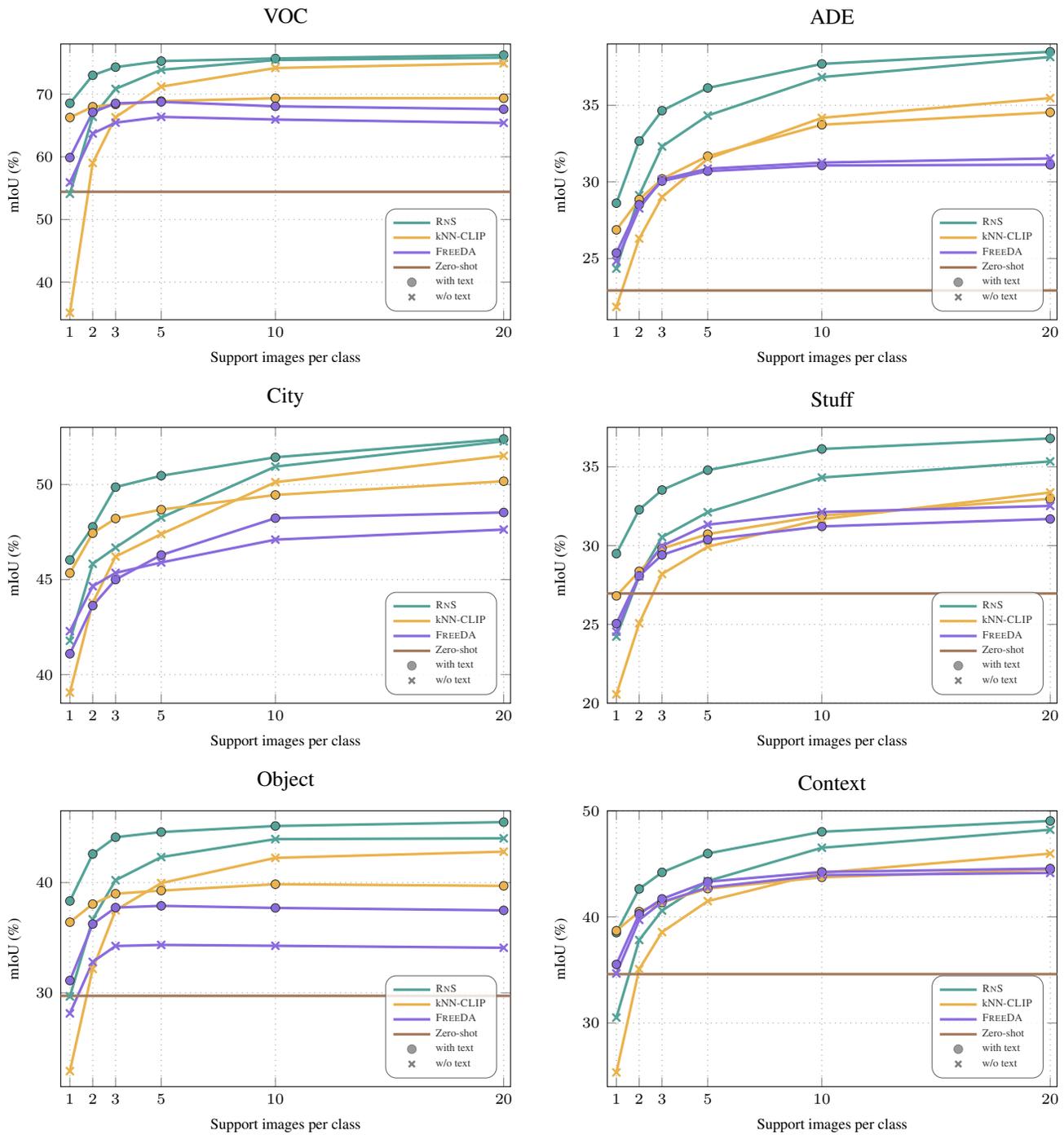
\begin{figure*}[t]
  \vspace{-10pt}
  \centering
  \setlength{\tabcolsep}{4pt} 
  \begin{tabular}{@{}c c@{}}    \input{plots/supp/full_visual_support_varying_mem_clip/voc} & \input{plots/supp/full_visual_support_varying_mem_clip/ade} \\    \input{plots/supp/full_visual_support_varying_mem_clip/city} &    \input{plots/supp/full_visual_support_varying_mem_clip/stuff} \\    \input{plots/supp/full_visual_support_varying_mem_clip/object} &  \input{plots/supp/full_visual_support_varying_mem_clip/context} \\
  \end{tabular}
  \vspace{-8pt}
  \caption{\textbf{Full textual and visual support per dataset (OpenCLIP, ViT-B/16).} We compare zero-shot, \ours, \knnclip and \freeda and their variants without class name information  (w/o text) for increasing number of support images per class. SAM~2.1 is used for region proposals.}
  \label{fig:full-clip-per-dataset}
\end{figure*}

In~\autoref{fig:personalization_memory} we show the visual support sets used to perform personalized segmentation in Figure 7 of the main paper.

\begin{figure*}[t]
  \vspace{-5pt}
  \centering
  \setlength{\tabcolsep}{4pt}
  \begin{tabular}{@{}c c@{}}
  \input{plots/supp/full_visual_support_varying_mem_dino/voc}   &    \input{plots/supp/full_visual_support_varying_mem_dino/ade}  \\    \input{plots/supp/full_visual_support_varying_mem_dino/city}   &    \input{plots/supp/full_visual_support_varying_mem_dino/stuff} \\   \input{plots/supp/full_visual_support_varying_mem_dino/object} &  \input{plots/supp/full_visual_support_varying_mem_dino/context} \\
  \end{tabular}
  \vspace{-8pt}
  \caption{\textbf{Full textual and visual support per dataset (DINOv3.txt, ViT-L/16).} We compare zero-shot, \ours, \knnclip and \ours without class name information  (w/o text) for increasing number of support images per class. Prediction at the patch level or SAM~2.1 is used for region proposals.}
  \label{fig:full-dino-per-dataset}
  \vspace{-8pt}
\end{figure*}
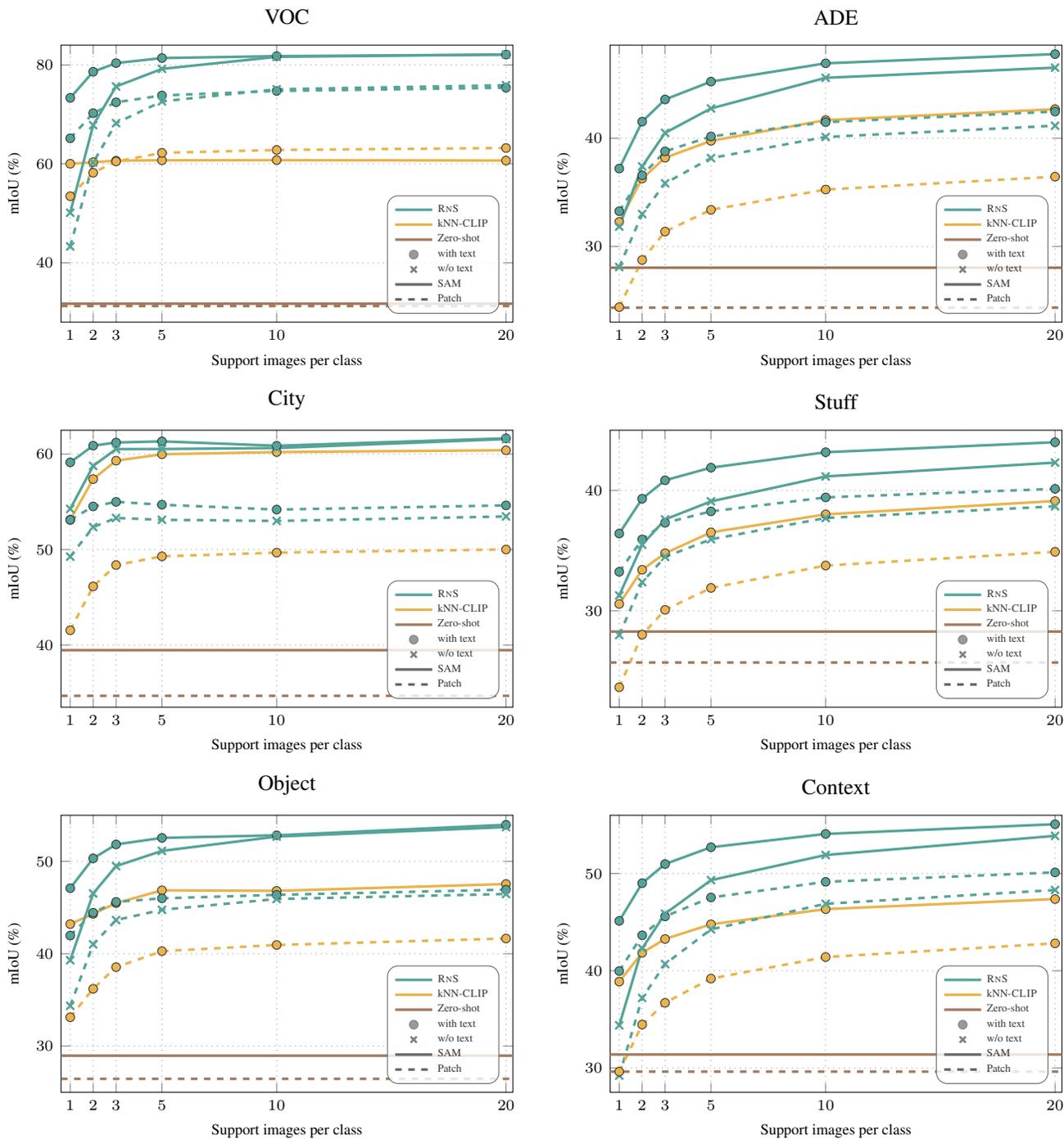

\begin{figure*}[t]
  \vspace{-5pt}
  \centering
  \setlength{\tabcolsep}{4pt}
  \begin{tabular}{@{}c c@{}}
    \input{plots/supp/partial_visual_support_mem/voc}    &
    \input{plots/supp/partial_visual_support_mem/ade}   \\
    \input{plots/supp/partial_visual_support_mem/city}    &
    \input{plots/supp/partial_visual_support_mem/stuff}  \\
    \input{plots/supp/partial_visual_support_mem/object} &
    \input{plots/supp/partial_visual_support_mem/context}\\
  \end{tabular}
  \vspace{-8pt}
  \caption{\textbf{Partial visual support setting per dataset.}
  Results of zero-shot, \ours, \knnclip, and \freeda, along with ablations of \ours without text and without the pseudo-label loss. OpenCLIP ViT-B/16 and SAM~2.1 are used. A fraction of classes lack visual examples, while $B=3$ for the remaining classes.}
  \label{fig:partial-visual-per-dataset}
  \vspace{-8pt}
\end{figure*}
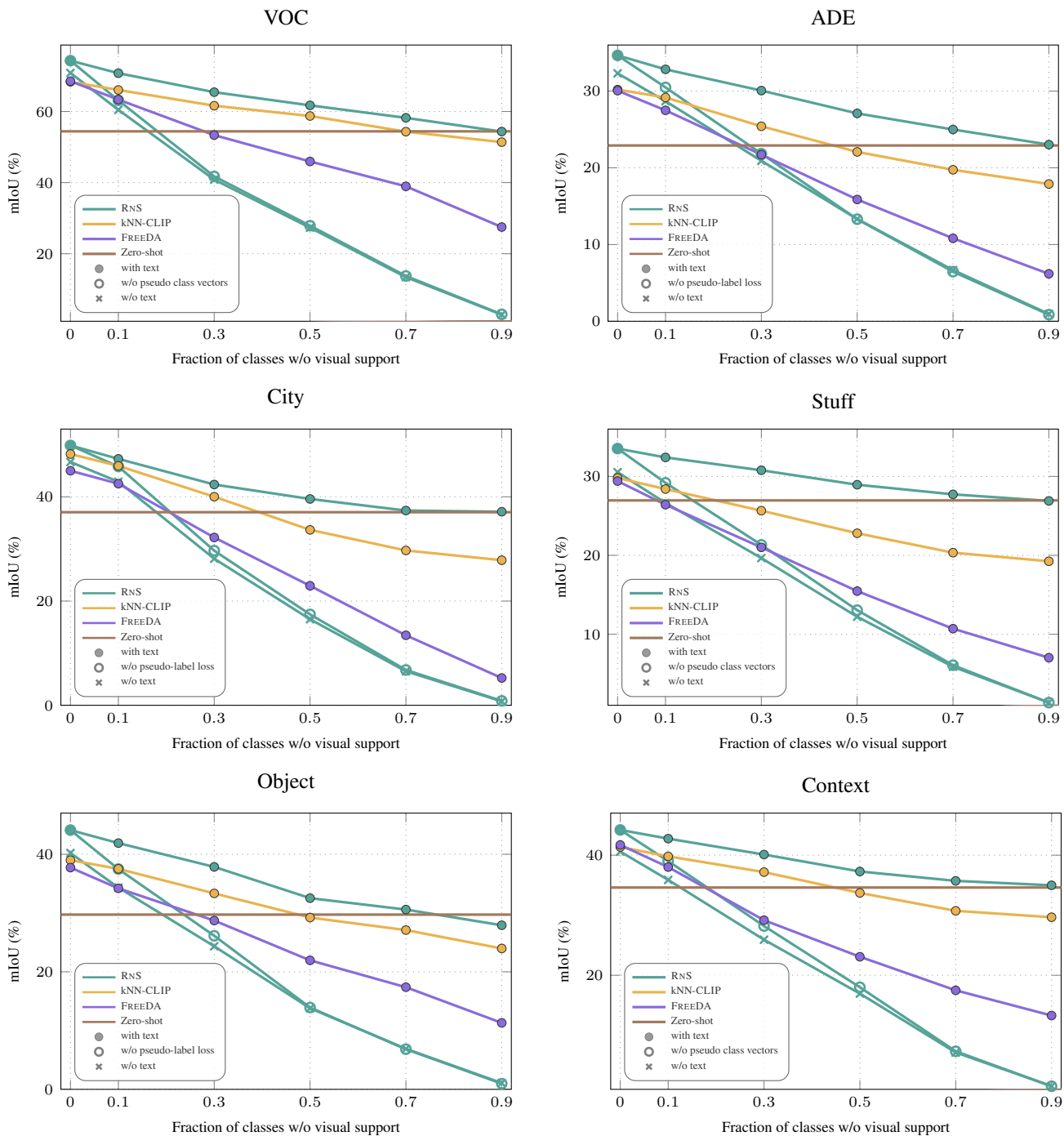

\begin{figure*}[t]
  \vspace{-5pt}
  \centering
  \setlength{\tabcolsep}{4pt}
  \begin{tabular}{@{}c c@{}}
    \input{plots/supp/partial_textual_support_mem/voc}    &
    \input{plots/supp/partial_textual_support_mem/ade}   \\
    \input{plots/supp/partial_textual_support_mem/city}    &
    \input{plots/supp/partial_textual_support_mem/stuff}  \\
    \input{plots/supp/partial_textual_support_mem/object} &
    \input{plots/supp/partial_textual_support_mem/context}\\
  \end{tabular}
  \vspace{-8pt}
  \caption{\textbf{Partial textual support setting per dataset.}
  Results of Zero-shot, \ours, \knnclip, and \freeda, together with their variants without class name information (w/o text). OpenCLIP ViT-B/16 and SAM~2.1 are used. A fraction of classes lack textual class names, and $B=1$ for all classes.}
  \label{fig:partial-text-per-dataset}
  \vspace{-8pt}
\end{figure*}
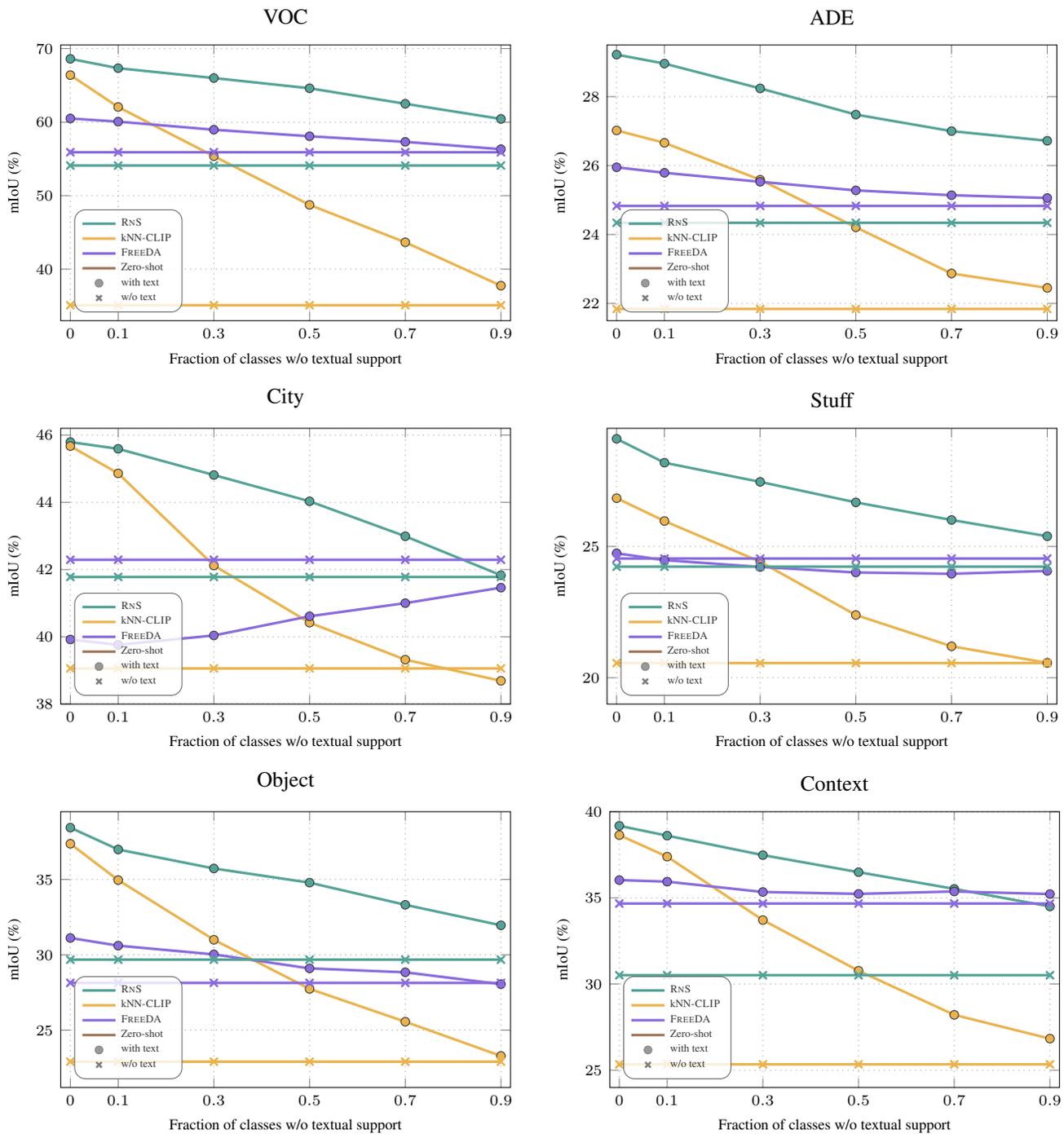

%% file: plots/ood2.tex
\begin{ShotMiouAxis}[
  width=0.85\linewidth, height=4.0cm,
  scale only axis,
  xtick={1,2,3,5,10,20},
  xmin=0.6, xmax=20.3,
  ymin=26, ymax=48,
  title={ACDC},
  xlabel={Images per class},
  ylabel={mIoU (\%)},
  legend style={
      draw=black!60,
      fill=white,
      fill opacity=0.8,
      rounded corners=5pt,
      font=\tiny,
      anchor=south east,
  },
]

  \AddShotMiouCurve{\ours: Cityscapes $\rightarrow$ ACDC}{ourscyan}{
    (1,36.51)
    (2,38.06)
    (3,39.06)
    (5,40.17)
    (10,41.44)
    (20,41.45)
  }

  \AddShotMiouCurve{\ours w/o text: Cityscapes $\rightarrow$ ACDC}{ourscyan}{
    (1,33.08)
    (2,35.37)
    (3,36.45)
    (5,38.67)
    (10,40.88)
    (20,41.08)
  }[mark=x, mark size=2.2pt,
    mark options={
      line cap=round,
      draw=ourscyan,
      line width=1.0pt,
      double=black,
      double distance=0.55pt
    }
  ]

  \AddShotMiouCurve{\ours: ACDC $\rightarrow$ ACDC}{appleteal}{
    (1,37.44)
    (2,40.97)
    (3,42.22)
    (5,43.89)
    (10,45.70)
    (20,47.07)
  }

  \AddShotMiouCurve{\ours w/o text: ACDC $\rightarrow$ ACDC}{appleteal}{
    (1,34.59)
    (2,37.42)
    (3,39.16)
    (5,41.79)
    (10,44.43)
    (20,46.86)
  }[mark=x, mark size=2.2pt,
    mark options={
      line cap=round,
      draw=appleteal,
      line width=1.0pt,
      double=black,
      double distance=0.55pt
    }
  ]

  \AddZeroShotLine{ACDC Zeroshot}{brown}{27.98}

\end{ShotMiouAxis}

%% file: plots/ood.tex
\begin{ShotMiouAxis}[
  width=0.85\linewidth, height=4.0cm,
  scale only axis,
  xtick={1,2,3,5,10,20},
  xmin=0.6, xmax=20.3,
  ymin=31, ymax=54,
  title={Cityscapes},
  xlabel={Images per class},
  ylabel={mIoU (\%)},
  legend style={
      draw=black!60,
      fill=white,
      fill opacity=0.8,
      rounded corners=5pt,
      font=\tiny,
      anchor=south east,
  },
]

  \AddShotMiouCurve{\ours: Cityscapes $\rightarrow$ Cityscapes}{appleteal}{
    (1,48.78)
    (2,49.55)
    (3,50.09)
    (5,51.05)
    (10,51.90)
    (20,53.62)
  }

  \AddShotMiouCurve{\ours w/o text: Cityscapes $\rightarrow$ Cityscapes}{appleteal}{
    (1,44.64)
    (2,46.47)
    (3,47.45)
    (5,49.19)
    (10,50.78)
    (20,52.67)
  }[mark=x, mark size=2.2pt,
    mark options={
      line cap=round,
      draw=appleteal,
      line width=1.0pt,
      double=black,
      double distance=0.55pt
    }
  ]

  \AddShotMiouCurve{\ours: ACDC $\rightarrow$ Cityscapes}{ourscyan}{
    (1,39.38)
    (2,41.49)
    (3,42.55)
    (5,43.87)
    (10,45.83)
    (20,46.20)
  }

  \AddShotMiouCurve{\ours w/o text: ACDC $\rightarrow$ Cityscapes}{ourscyan}{
    (1,31.78)
    (2,37.71)
    (3,40.89)
    (5,42.50)
    (10,44.69)
    (20,45.20)
  }[mark=x, mark size=2.2pt,
    mark options={
      line cap=round,
      draw=ourscyan,
      line width=1.0pt,
      double=black,
      double distance=0.55pt
    }
  ]

  \AddZeroShotLine{Cityscapes Zeroshot}{brown}{37.88}

\end{ShotMiouAxis}

%% file: plots/backbones.tex
\begin{ShotMiouAxis}[
  width=0.85\linewidth, height=4.0cm,
  scale only axis,
  title={Backbone Comparison},
  xtick={1,2,3,5,10,20},
  xmin=0.6, xmax=20.3,
  ymin=28, ymax=59,
xlabel={Images per class},
  ylabel={mIoU (\%)},
  legend style={
      draw=black!60,
      fill=white,
      fill opacity=0.8,
      rounded corners=5pt,
      font=\tiny,
      at={(rel axis cs:0.97,0.22)},
      anchor=south east,
  },
]

  \AddShotMiouCurve{\ours: OpenCLIP ViT-B/16}{oursgreen}{
    (1,41.59) (2,45.16) (3,46.78) (5,47.87) (10,49.02) (20,49.74)
  }

  \AddShotMiouCurve{\ours: DINOv3.txt ViT-L/16}{kclipblue}{
    (1,49.72) (2,53.28) (3,54.80) (5,55.85) (10,56.61) (20,57.42)
  }

  \AddShotMiouCurve{\ours: SigLIP2 ViT-L/16}{freedaorange}{
    (1,43.74)
    (2,48.35)
    (3,50.07)
    (5,51.13)
    (10,52.72)
    (20,53.78)
  }

  \AddZeroShotLine{Zeroshot: OpenCLIP ViT-B/16}{oursgreen}{34.28}
  \AddZeroShotLine{Zeroshot: DINOv3.txt ViT-L/16}{kclipblue}{31.31}
  \AddZeroShotLine{Zeroshot: SigLIP2 ViT-L/16}{freedaorange}{30.44}

\end{ShotMiouAxis}

%% file: plots/rebuttal/partial_visual_support_finegrained.tex
\begin{DropFracAxis}[
  xlabel={Fraction w/o visual support},
  ylabel={mIoU (\%)},
  width=.35\linewidth, height=2.4cm,
  scale only axis,
  xtick={0.0,0.1,0.3,0.5,0.7,0.9},
  xmin=-0.02, xmax=0.92,
  ymin=12, ymax=36,
  legend style={font=\fontsize{4pt}{4pt}\selectfont, row sep=-2pt},
]

\addplot[color=appleteal, line width=1.2pt, mark=*,] coordinates {(0,0)(1,1)};
\addlegendentry{\ours};

\addplot[color=appleteal, line width=1.2pt, mark=star,] coordinates {(0,0)(1,1)};
\addlegendentry{\ours w/o $w_{c}$};

\addplot[color=applebrown, line width=1.2pt] coordinates {(0,0)(1,1)};
\addlegendentry{Zero-shot};



\addplot+[
  color=appleteal,
  line width=1.2pt,
  mark=*,
  mark size=2.0pt,
  mark options={solid, fill=appleteal, line width=0.4pt},
  dashed,
  forget plot,
] coordinates {
  (0.00,33.53) (0.10,31.83) (0.30,29.34) (0.50,26.87) (0.70,25.41) (0.90,24.05)
};

\addplot+[
  color=appleteal,
  line width=1.2pt,
  mark=star,
  mark size=2.5pt,
  mark options={solid, fill=appleteal, line width=0.7pt},
  dashed,
  forget plot,
] coordinates {
  (0.00,31.76) (0.10,30.07) (0.30,26.89) (0.50,24.82) (0.70,23.51) (0.90,22.96)
};

\addplot+[
  color=appleteal,
  line width=1.2pt,
  mark=*,
  mark size=2.0pt,
  mark options={solid, fill=appleteal, line width=0.4pt},
  solid,
  forget plot,
] coordinates {
  (0.00,34.45) (0.10,31.97) (0.30,27.32) (0.50,22.73) (0.70,18.73) (0.90,15.94)
};

\addplot+[
  color=appleteal,
  line width=1.2pt,
  mark=star,
  mark size=2.5pt,
  mark options={solid, fill=appleteal, line width=0.7pt},
  solid,
  forget plot,
] coordinates {
  (0.00,33.86) (0.10,30.74) (0.30,25.36) (0.50,20.82) (0.70,16.61) (0.90,14.16)
};

\addplot+[
  color=applebrown,
  line width=1.2pt,
  dashed,
  forget plot,
] coordinates {(-0.09,23.0) (0.99,23.0)};

\addplot+[
  color=applebrown,
  line width=1.2pt,
  solid,
  forget plot,
] coordinates {(-0.09,15.0) (0.99,15.0)};

\end{DropFracAxis}

%% file: plots/rebuttal/partial_visual_support_retrieval_quality.tex
\begin{DropFracAxis}[
  xlabel={Fraction w/o visual support},
  ylabel={Retrienal error (\%)},
  width=.35\linewidth, height=2.4cm,
  scale only axis,
  xtick={0.0,0.1,0.3,0.5,0.7,0.9},
  xmin=-0.02, xmax=0.92,
  ymin=30, ymax=100,
  legend pos=south east,
  legend style={font=\fontsize{4pt}{4pt}\selectfont, row sep=-2pt},
]

\addplot[color=gray, line width=1.2pt, dashed] coordinates {(0,0)(1,1)};
\addlegendentry{Food};

\addplot[color=gray, line width=1.2pt] coordinates {(0,0)(1,1)};
\addlegendentry{CUB};

\addplot+[
  color=appleteal,
  line width=1.2pt,
  mark=*,
  mark size=2.0pt,
  mark options={solid, fill=appleteal, draw=black!80, line width=0.4pt},
  dashed,
  forget plot,
] coordinates {
  (0.00,41.41) (0.10,45.43) (0.30,54.2) (0.50,63.06) (0.70,73.04) (0.90,84.45)
};

\addplot+[
  color=appleteal,
  line width=1.2pt,
  mark=*,
  mark size=2.0pt,
  mark options={solid, fill=appleteal, draw=black!80, line width=0.4pt},
  solid,
  forget plot,
] coordinates {
  (0.00,37.36) (0.10,43.11) (0.30,54.76) (0.50,66.4) (0.70,79.08) (0.90,91.33)
};

\end{DropFracAxis}

%% file: plots/offline_fixed_params.tex
\begin{ShotMiouAxis}[
  width=0.85\linewidth, height=4.0cm,
  scale only axis,
  xtick={1,2,3,5,10,20},
  xmin=0.8, xmax=20.3,
  ymin=18, ymax=62,
  xlabel={Images per class},
  ylabel={mIoU (\%)},
  title={Offline Baselines},
  legend style={
      draw=black!60,
      fill=white,
      rounded corners=5pt,
      font=\tiny
  },
]

  \AddShotMiouCurve{\ours, hyperparameter $\mathbf{set~1}$}{appleteal}
    { (1,54.25) (2,55.36) (3,55.50) (5,56.20) (10,57.73) (20,59.35) }[mark=square*,mark size=2pt,densely dashed,]

  \AddShotMiouCurve{\ours, hyperparameter $\mathbf{set~2}$}{appleteal}
    { (1,53.70) (2,54.56) (3,54.95) (5,55.45) (10,57.11) (20,58.70) }[mark=diamond*,mark size=2pt,densely dashed,]

  \AddShotMiouCurve{\ours, optimal}{appleteal}
    { (1,54.33) (2,55.63) (3,57.25) (5,58.24) (10,59.31) (20,60.73) }

  \AddShotMiouCurve{linear clf on visual class features, hyperparameter $\mathbf{set~1}$}{appleorange}
    { (1,33.26) (2,47.39) (3,50.75) (5,53.53) (10,51.94) (20,50.73) }[mark=square*,mark size=2pt,densely dashed,]

  \AddShotMiouCurve{linear clf on visual class features, hyperparameter $\mathbf{set~2}$}{appleorange}
    { (1,19.64) (2,24.94) (3,28.83) (5,35.30) (10,46.10) (20,56.58) }[mark=diamond*,mark size=2pt,densely dashed,]

  \AddShotMiouCurve{linear clf on visual class features, optimal}{appleorange}
    { (1,38.79) (2,49.17) (3,51.17) (5,54.59) (10,55.78) (20,56.51) }

\end{ShotMiouAxis}

%% file: tab/soup_suppl.tex
\begin{tabular}{lccccccccccc}
\toprule
\textbf{Method} & \textbf{Training set} & \textbf{Annot.} & \textbf{Backbones} & \textbf{VOC} & \textbf{City} & \textbf{ADE} & \textbf{C-59} & \textbf{Food} & \textbf{CUB} & \textbf{Avg.} \\
\midrule
ODISE~\cite{xu2023odise} & COCO & 118k & Stable Diffusion v1.3, CLIP (ViT-L/14) & \underline{84.6} & -- & 29.9 & 57.3 & -- & -- & -- \\
OVSeg$^{\dag}$~\cite{liang2023open} & COCO & 118k & CLIP (ViT-L/14), Swin-B &  -- & -- & 29.6 & 55.7 & 16.4 & 14.0 & -- \\
SAN$^{\dag}$~\cite{xu2023san} & COCO & 118k & CLIP (ViT-L/14) & -- & -- & 32.1 & 57.7 & 24.5 & 19.3 & -- \\
CAT-Seg~\cite{cho2024catseg} & COCO & 118k & CLIP (ViT-L/14) & 82.5 & 47.0$^{*}$ & 37.9 & \underline{63.3} & 33.3 & 22.9 & 47.8 \\
LPOSS+~\cite{stojnic2025lposs} & \xmark & \xmark & OpenCLIP (ViT-B/16), DINO (ViT-B/16) & 62.4 & 37.9 & 22.3 & 38.6 & 26.1 & 12.0 & 33.2 \\
CLIP-DINOiser~\cite{clipdinoiser} & \xmark & \xmark & OpenCLIP (ViT-B/16), DINO (ViT-B/16) & 62.1 & 31.7 & 20.0 & 35.9 & -- & -- & -- \\
CorrCLIP~\cite{zhang2025corrclip} & \xmark & \xmark & OpenCLIP (ViT-H/14), DINO (ViT-B/8), SAM2.1 (Hiera-L) & 76.4 & 49.9 & 28.8 & 47.9 & -- & -- & -- \\
\midrule
MaskCLIP~\cite{maskclip} + SAM & \xmark & \xmark & OpenCLIP (ViT-B/16), SAM2.1 (Hiera-L) & 54.4 & 37.1 & 22.9 & 38.0 & 23.0 & 15.0 & 31.7 \\
\rowcolor{appletealL3} 
+ \ours~$B=1$ & Domain & 66 & OpenCLIP (ViT-B/16), SAM2.1 (Hiera-L) & 68.6 & 46.1 & 28.6 & 45.9 & 28.1 & 24.9 & 40.4 \\
\rowcolor{appletealL3}
+ \ours~$B=20$ & Domain & 964 & OpenCLIP (ViT-B/16), SAM2.1 (Hiera-L) & 76.2 & 52.5 & 38.5 & 54.4 & 37.2 & 50.6 & 51.6 \\
DINOv3.txt~\cite{simeoni2025dinov3} + SAM & \xmark & \xmark & DINOv3 (ViT-L/16), SAM2.1 (Hiera-L) & 31.3 & 39.3 & 27.7 & 36.3 & 27.2 & 5.8 & 27.9 \\
\rowcolor{appletealL3}
+ \ours~$B=1$ & Domain & 66 & DINOv3 (ViT-L/16), SAM2.1 (Hiera-L) & 73.2 & 59.1 & 37.3 & 52.7 & 42.8 & 34.0 & 49.9 \\
\rowcolor{appletealL3}
+ \ours~$B=20$ & Domain & 964 & DINOv3 (ViT-L/16), SAM2.1 (Hiera-L) & 82.1 & 61.7 & 47.8 & 62.5 & \textbf{52.2} & \underline{65.2} & \underline{61.9} \\
\midrule
InternImage~\cite{wang2023internimage} & Domain & -- & InternImage-H & -- & \textbf{87.0} & \underline{62.9} & \textbf{70.3} & -- &  -- & --\\
SETR~\cite{zheng2021segformer} & Domain & -- & SeTR-MLA & -- & 76.7 & 48.6 & -- & \underline{45.1} & -- & --\\
GFN~\cite{zhang2022guided} & Domain & -- & ResNet-101 & -- & -- & -- & -- & -- & \textbf{84.6} & --\\
Mask2Former~\cite{cheng2022mask2former} & Domain & -- & DINOv3 (7B) & \textbf{90.4} & \underline{86.7} & \textbf{63.0} & -- & -- & -- & --\\
\hdashline[2pt/4pt]
\textbf{Fully Supervised} & Domain & 20k & Best of Above & \textbf{90.4} & \textbf{87.0} & \textbf{63.0} & \textbf{70.3} & \underline{45.1} & \textbf{84.6} & \textbf{73.4} \\
\bottomrule
\end{tabular}%

%% file: plots/rebuttal/performance_time.tex
\begin{tikzpicture}
\begin{axis}[
    width=1.1\linewidth,
    height=0.52\linewidth,
    xmin=0.03,
    ymin=25.5,
    ymax=57.0,
    grid=both,
    xmax=0.9,
    grid style={dotted,gray!80},
    legend pos=south east, 
    legend style={
      draw=black!60,
      fill=white,
      fill opacity=0.8,
      rounded corners=5pt,
      font=\tiny,
      row sep=-2pt,
    },
    legend cell align=left,
    xlabel style={font=\scriptsize},
    ylabel style={font=\scriptsize},
    x tick label style={font=\scriptsize},
    y tick label style={font=\scriptsize},
]

\addplot[
    only marks,
    mark=*,
    appleteal,
] coordinates {
    (0.83,50.75)
    (0.66,50.85)
    (0.45,50.91)
    (0.26,49.76)
    (0.20,46.40)
};
\addlegendentry{\ours}

\node[black] at (axis cs:0.83,50.75) [anchor=south] {\scriptsize 700 iter.};
\node[black] at (axis cs:0.66,50.85) [anchor=south] {\scriptsize 500 iter.};
\node[black] at (axis cs:0.45,50.91) [anchor=south] {\scriptsize 300 iter.};
\node[black] at (axis cs:0.26,49.76) [anchor=south] {\scriptsize 100 iter.};
\node[black] at (axis cs:0.20,46.40) [anchor=south] {\scriptsize 30 iter.};

\addplot[
    only marks,
    mark=*,
    applemustard,
] coordinates {
    (0.17,42.45)
};
\addlegendentry{\knnclip}

\addplot[
    only marks,
    mark=*,
    applebrown,
] coordinates {
    (0.07,27.09)
};
\addlegendentry{Zero-shot}


\end{axis}
\end{tikzpicture}

%% file: tab/ablation_k.tex
\begin{tabular}{llll}
    \toprule
    \textbf{Method} & \textbf{$B=1$} & \textbf{$B=5$} & \textbf{$B=10$} \\
    \midrule
    \ours\ ($K=1$) 
        & 40.02 \textcolor{blue}{\,-1.57} 
        & 46.83 \textcolor{blue}{\,-1.04} 
        & 48.06 \textcolor{blue}{\,-0.96} \\
    \rowcolor{appletealL3}
    \ours\ ($K=4$) 
        & 41.59 
        & 47.87 
        & 49.02 \\
    \ours\ ($K=16$) 
        & 41.97 \textcolor{blue}{\,+0.38} 
        & 47.78 \textcolor{blue}{\,-0.09} 
        & 48.96 \textcolor{blue}{\,-0.06} \\
    \ours\ ($K=32$) 
        & 42.02 \textcolor{blue}{\,+0.43} 
        & 47.59 \textcolor{blue}{\,-0.28} 
        & 48.84 \textcolor{blue}{\,-0.18} \\
    \bottomrule
\end{tabular}

%% file: fig/personalization_memory.tex
\begin{tabular}{@{\hspace{-10pt}}c@{\hspace{10pt}}|c@{\hspace{-10pt}}}
      \includegraphics[width=0.45\linewidth]{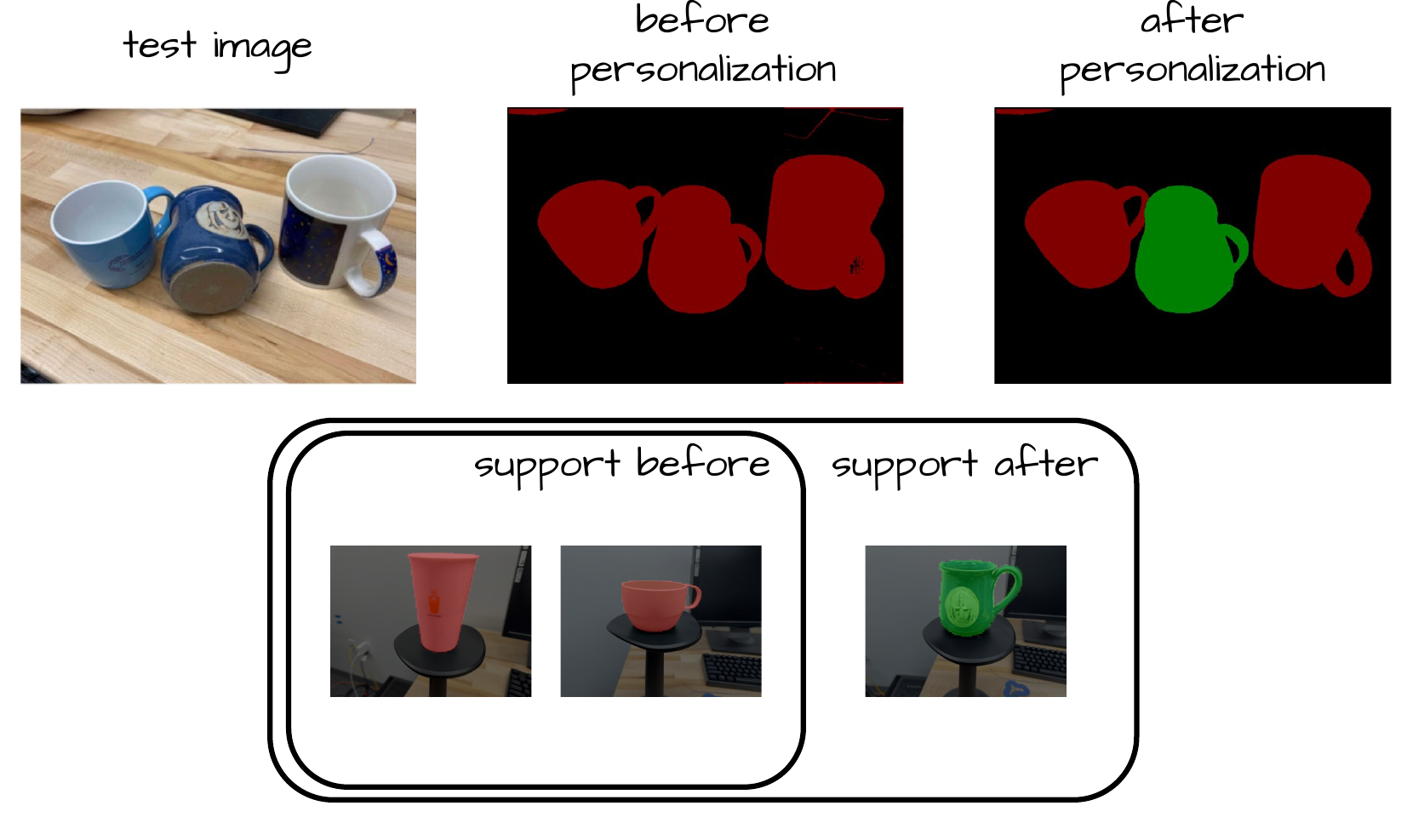} &
      \includegraphics[width=0.45\linewidth]{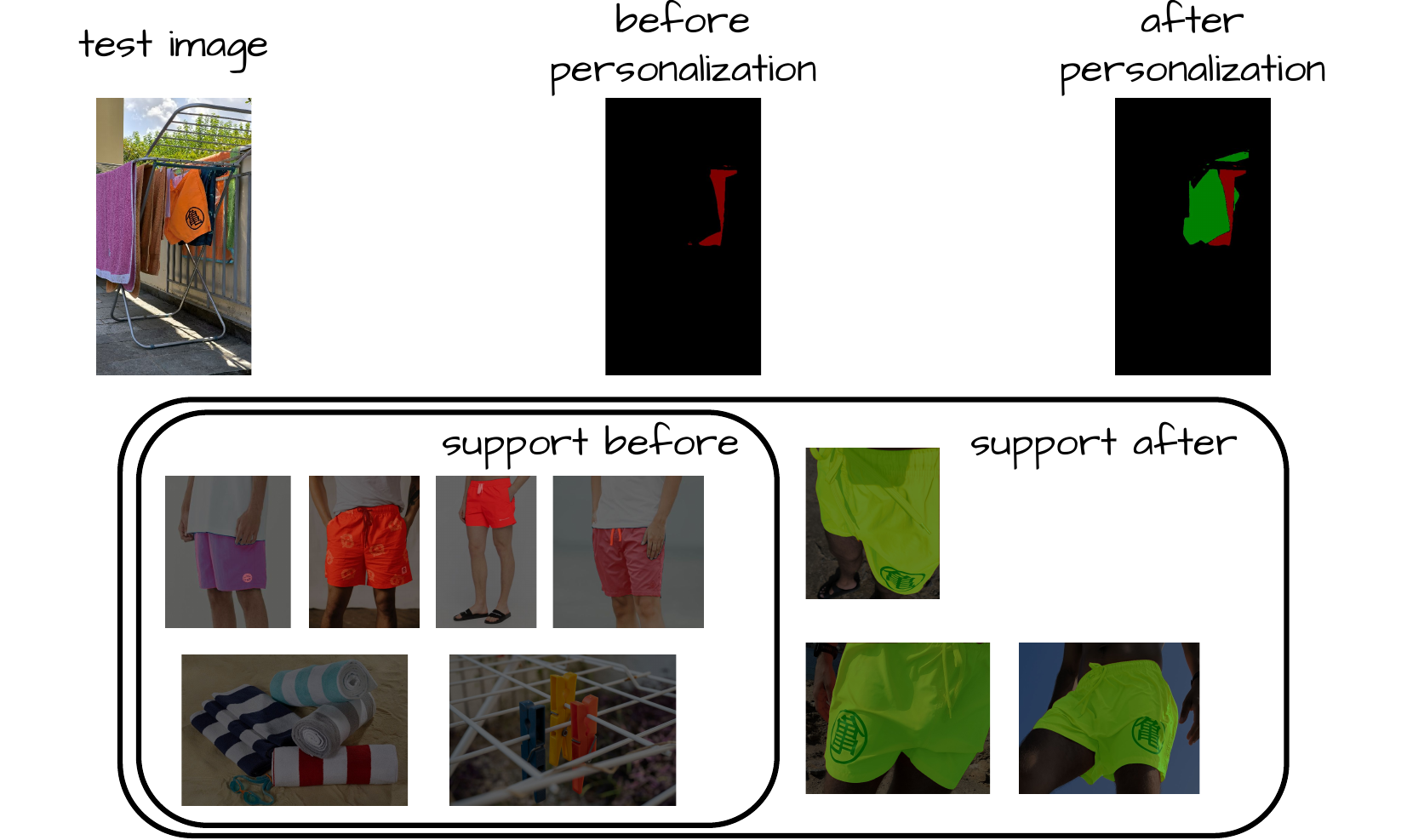} \\ [0.5cm]
      \midrule
      \includegraphics[width=0.45\linewidth]{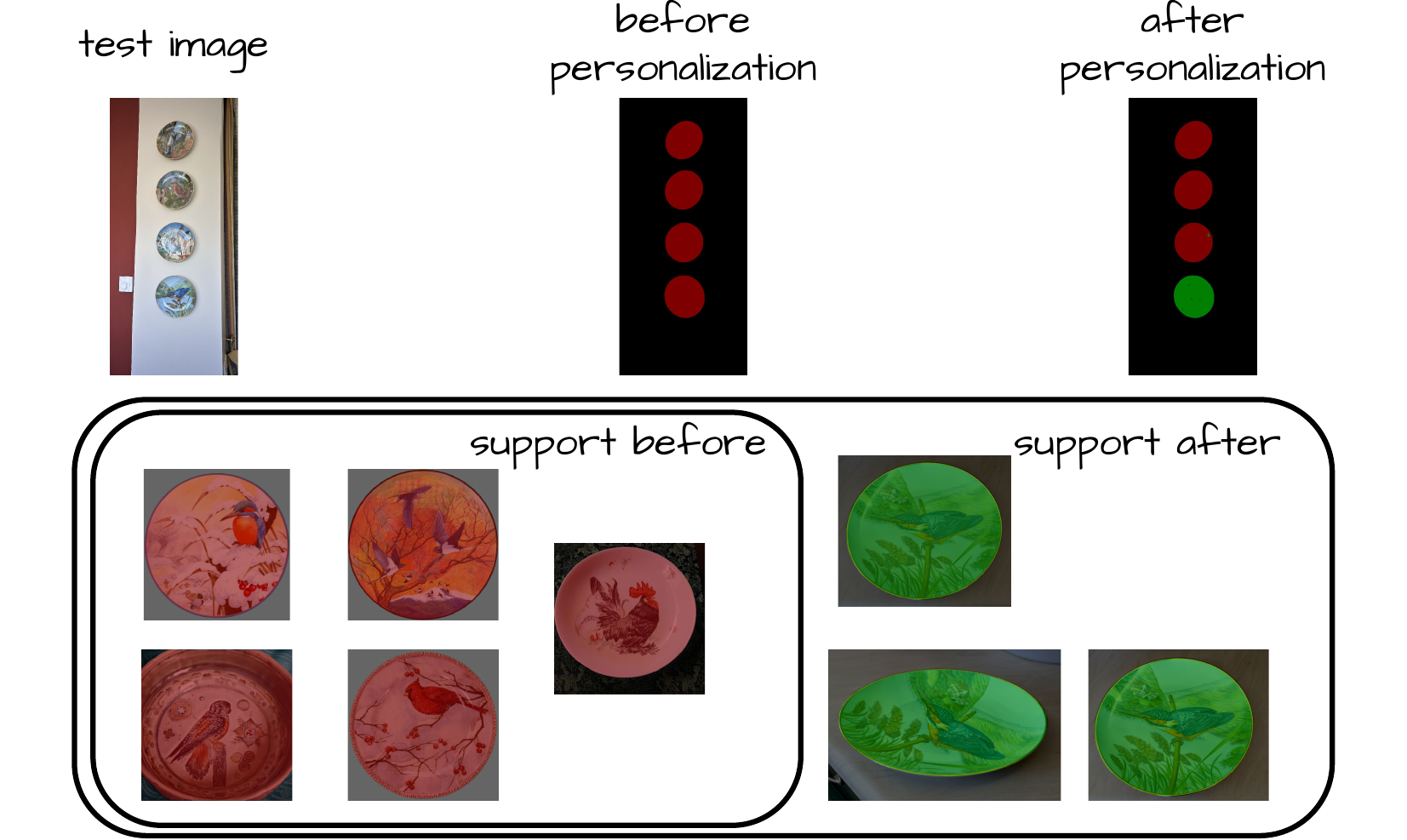} &
      \includegraphics[width=0.45\linewidth]{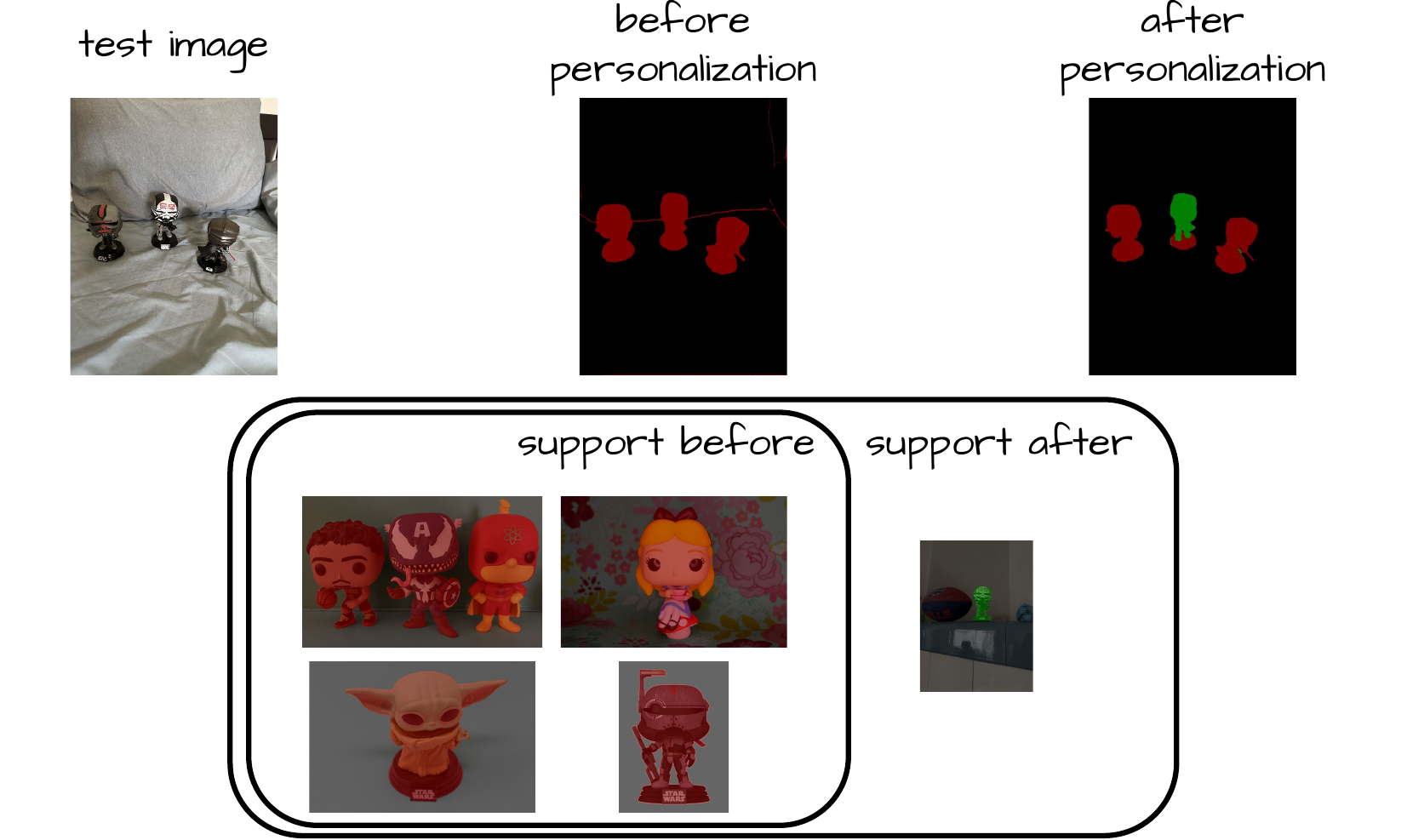} \\ [0.5cm]
      \midrule
      \multicolumn{2}{c}{\includegraphics[width=0.45\linewidth]{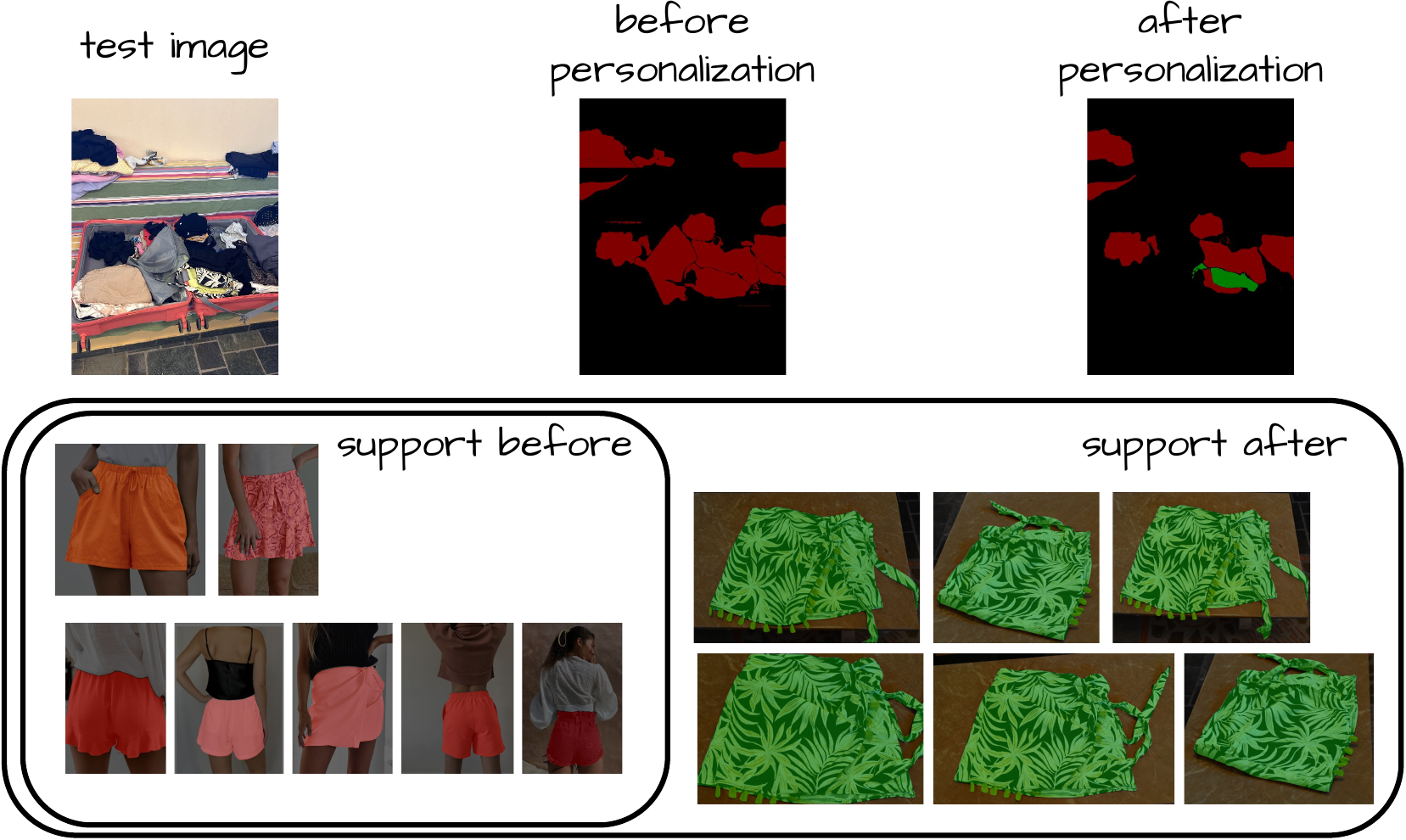}}
\end{tabular}

%% file: plots/supp/full_visual_support_varying_mem_clip/voc.tex
\begin{ShotMiouAxis}[
  width=0.42\linewidth, height=4.5cm,
  title={VOC},
  scale only axis,
  xtick={1,2,3,5,10,20},
  xmin=0.6, xmax=20.3,
  ymin=34, ymax=78,
  xlabel={Support images per class},
]
\addplot[color=appleteal, line width=1.2pt] coordinates {(0,0)(1,1)}; \addlegendentry{\ours};
\addplot[color=applemustard, line width=1.2pt] coordinates {(0,0)(1,1)}; \addlegendentry{\knnclip};
\addplot[color=applepurple, line width=1.2pt] coordinates {(0,0)(1,1)}; \addlegendentry{\freeda};
\addplot[color=applebrown, line width=1.2pt] coordinates {(0,0)(1,1)}; \addlegendentry{Zero-shot};
\addplot[only marks, mark=*, mark options={fill=gray, draw=gray}] coordinates {(0,0)}; \addlegendentry{with text};
\addplot[only marks, mark=x, mark options={draw=gray, line width=1pt}] coordinates {(0,0)}; \addlegendentry{w/o text};

\AddShotMiouCurve{}{appleteal}{(1,68.54) (2,73.01) (3,74.31) (5,75.27) (10,75.69) (20,76.24)}
\AddShotMiouCurve{}{appleteal}{(1,54.10) (2,66.38) (3,70.84) (5,73.88) (10,75.44) (20,75.82)}[mark=x, mark size=2.2pt, mark options={line cap=round, draw=appleteal, line width=1pt, double=black, double distance=0.55pt}
]

\AddShotMiouCurve{}{applemustard}{(1,66.27) (2,67.99) (3,68.36) (5,68.89) (10,69.35) (20,69.35)}
\AddShotMiouCurve{}{applemustard}{(1,35.10) (2,59.02) (3,66.31) (5,71.20) (10,74.17) (20,74.91)}[mark=x, mark size=2.2pt, mark options={line cap=round, draw=applemustard, line width=1pt, double=black, double distance=0.55pt}
]

\AddShotMiouCurve{}{applepurple}{(1,59.88) (2,67.11) (3,68.53) (5,68.77) (10,68.06) (20,67.60)}
\AddShotMiouCurve{}{applepurple}{(1,55.91) (2,63.72) (3,65.45) (5,66.37) (10,65.95) (20,65.41)}[mark=x, mark size=2.2pt, mark options={line cap=round, draw=applepurple, line width=1pt, double=black, double distance=0.55pt}]

\AddZeroShotLine{}{applebrown}{54.42}

\end{ShotMiouAxis}

%% file: plots/supp/full_visual_support_varying_mem_clip/ade.tex
\begin{ShotMiouAxis}[
  width=0.42\linewidth, height=4.5cm,
  title={ADE},
  scale only axis,
  xtick={1,2,3,5,10,20},
  xmin=0.6, xmax=20.3,
  ymin=21, ymax=39,
  xlabel={Support images per class},
]
\addplot[color=appleteal, line width=1.2pt] coordinates {(0,0)(1,1)}; \addlegendentry{\ours};
\addplot[color=applemustard, line width=1.2pt] coordinates {(0,0)(1,1)}; \addlegendentry{\knnclip};
\addplot[color=applepurple, line width=1.2pt] coordinates {(0,0)(1,1)}; \addlegendentry{\freeda};
\addplot[color=applebrown, line width=1.2pt] coordinates {(0,0)(1,1)}; \addlegendentry{Zero-shot};
\addplot[only marks, mark=*, mark options={fill=gray, draw=gray}] coordinates {(0,0)}; \addlegendentry{with text};
\addplot[only marks, mark=x, mark options={draw=gray, line width=1pt}] coordinates {(0,0)}; \addlegendentry{w/o text};

\AddShotMiouCurve{}{appleteal}{(1,28.61) (2,32.67) (3,34.65) (5,36.13) (10,37.70) (20,38.49)}
\AddShotMiouCurve{}{appleteal}{(1,24.34) (2,29.13) (3,32.31) (5,34.33) (10,36.84) (20,38.15)}[mark=x, mark size=2.2pt, mark options={line cap=round, draw=appleteal, line width=1pt, double=black, double distance=0.55pt}
]

\AddShotMiouCurve{}{applemustard}{(1,26.87) (2,28.85) (3,30.19) (5,31.68) (10,33.73) (20,34.54)}
\AddShotMiouCurve{}{applemustard}{(1,21.84) (2,26.30) (3,29.01) (5,31.51) (10,34.18) (20,35.47)}[mark=x, mark size=2.2pt, mark options={line cap=round, draw=applemustard, line width=1pt, double=black, double distance=0.55pt}
]

\AddShotMiouCurve{}{applepurple}{(1,25.36) (2,28.51) (3,30.05) (5,30.70) (10,31.07) (20,31.12)}
\AddShotMiouCurve{}{applepurple}{(1,24.83) (2,28.27) (3,30.15) (5,30.86) (10,31.26) (20,31.53)}[mark=x, mark size=2.2pt, mark options={line cap=round, draw=applepurple, line width=1pt, double=black, double distance=0.55pt}]

\AddZeroShotLine{}{applebrown}{22.91}

\end{ShotMiouAxis}

%% file: plots/supp/full_visual_support_varying_mem_clip/city.tex
\begin{ShotMiouAxis}[
  width=0.42\linewidth, height=4.5cm,
  title={City},
  scale only axis,
  xtick={1,2,3,5,10,20},
  xmin=0.6, xmax=20.3,
  ymin=38.5, ymax=53,
  xlabel={Support images per class},
]
\addplot[color=appleteal, line width=1.2pt] coordinates {(0,0)(1,1)}; \addlegendentry{\ours};
\addplot[color=applemustard, line width=1.2pt] coordinates {(0,0)(1,1)}; \addlegendentry{\knnclip};
\addplot[color=applepurple, line width=1.2pt] coordinates {(0,0)(1,1)}; \addlegendentry{\freeda};
\addplot[color=applebrown, line width=1.2pt] coordinates {(0,0)(1,1)}; \addlegendentry{Zero-shot};
\addplot[only marks, mark=*, mark options={fill=gray, draw=gray}] coordinates {(0,0)}; \addlegendentry{with text};
\addplot[only marks, mark=x, mark options={draw=gray, line width=1pt}] coordinates {(0,0)}; \addlegendentry{w/o text};

\AddShotMiouCurve{}{appleteal}{(1,46.03) (2,47.77) (3,49.86) (5,50.46) (10,51.43) (20,52.39)}
\AddShotMiouCurve{}{appleteal}{(1,41.78) (2,45.83) (3,46.68) (5,48.26) (10,50.94) (20,52.27)}[mark=x, mark size=2.2pt, mark options={line cap=round, draw=appleteal, line width=1pt, double=black, double distance=0.55pt}
]

\AddShotMiouCurve{}{applemustard}{(1,45.34) (2,47.44) (3,48.21) (5,48.68) (10,49.45) (20,50.17)}
\AddShotMiouCurve{}{applemustard}{(1,39.06) (2,43.78) (3,46.21) (5,47.39) (10,50.12) (20,51.51)}[mark=x, mark size=2.2pt, mark options={line cap=round, draw=applemustard, line width=1pt, double=black, double distance=0.55pt}
]

\AddShotMiouCurve{}{applepurple}{(1,41.10) (2,43.63) (3,45.00) (5,46.29) (10,48.23) (20,48.53)}
\AddShotMiouCurve{}{applepurple}{(1,42.29) (2,44.65) (3,45.35) (5,45.90) (10,47.10) (20,47.63)}[mark=x, mark size=2.2pt, mark options={line cap=round, draw=applepurple, line width=1pt, double=black, double distance=0.55pt}]

\AddZeroShotLine{}{applebrown}{37.05}

\end{ShotMiouAxis}

%% file: plots/supp/full_visual_support_varying_mem_clip/stuff.tex
\begin{ShotMiouAxis}[
  width=0.42\linewidth, height=4.5cm,
  title={Stuff},
  scale only axis,
  xtick={1,2,3,5,10,20},
  xmin=0.6, xmax=20.3,
  ymin=20, ymax=37.5,
  xlabel={Support images per class},
]
\addplot[color=appleteal, line width=1.2pt] coordinates {(0,0)(1,1)}; \addlegendentry{\ours};
\addplot[color=applemustard, line width=1.2pt] coordinates {(0,0)(1,1)}; \addlegendentry{\knnclip};
\addplot[color=applepurple, line width=1.2pt] coordinates {(0,0)(1,1)}; \addlegendentry{\freeda};
\addplot[color=applebrown, line width=1.2pt] coordinates {(0,0)(1,1)}; \addlegendentry{Zero-shot};
\addplot[only marks, mark=*, mark options={fill=gray, draw=gray}] coordinates {(0,0)}; \addlegendentry{with text};
\addplot[only marks, mark=x, mark options={draw=gray, line width=1pt}] coordinates {(0,0)}; \addlegendentry{w/o text};

\AddShotMiouCurve{}{appleteal}{(1,29.49) (2,32.28) (3,33.53) (5,34.79) (10,36.13) (20,36.80)}
\AddShotMiouCurve{}{appleteal}{(1,24.23) (2,28.05) (3,30.54) (5,32.12) (10,34.32) (20,35.34)}[mark=x, mark size=2.2pt, mark options={line cap=round, draw=appleteal, line width=1pt, double=black, double distance=0.55pt}
]

\AddShotMiouCurve{}{applemustard}{(1,26.81) (2,28.37) (3,29.81) (5,30.73) (10,31.91) (20,32.97)}
\AddShotMiouCurve{}{applemustard}{(1,20.56) (2,25.07) (3,28.20) (5,29.93) (10,31.68) (20,33.37)}[mark=x, mark size=2.2pt, mark options={line cap=round, draw=applemustard, line width=1pt, double=black, double distance=0.55pt}
]

\AddShotMiouCurve{}{applepurple}{(1,25.05) (2,28.08) (3,29.41) (5,30.38) (10,31.22) (20,31.69)}
\AddShotMiouCurve{}{applepurple}{(1,24.54) (2,28.09) (3,29.98) (5,31.33) (10,32.13) (20,32.52)}[mark=x, mark size=2.2pt, mark options={line cap=round, draw=applepurple, line width=1pt, double=black, double distance=0.55pt}]

\AddZeroShotLine{}{applebrown}{26.96}

\end{ShotMiouAxis}

%% file: plots/supp/full_visual_support_varying_mem_clip/object.tex
\begin{ShotMiouAxis}[
  width=0.42\linewidth, height=4.5cm,
  title={Object},
  scale only axis,
  xtick={1,2,3,5,10,20},
  xmin=0.6, xmax=20.3,
  ymin=21.5, ymax=46.5,
  xlabel={Support images per class},
]
\addplot[color=appleteal, line width=1.2pt] coordinates {(0,0)(1,1)}; \addlegendentry{\ours};
\addplot[color=applemustard, line width=1.2pt] coordinates {(0,0)(1,1)}; \addlegendentry{\knnclip};
\addplot[color=applepurple, line width=1.2pt] coordinates {(0,0)(1,1)}; \addlegendentry{\freeda};
\addplot[color=applebrown, line width=1.2pt] coordinates {(0,0)(1,1)}; \addlegendentry{Zero-shot};
\addplot[only marks, mark=*, mark options={fill=gray, draw=gray}] coordinates {(0,0)}; \addlegendentry{with text};
\addplot[only marks, mark=x, mark options={draw=gray, line width=1pt}] coordinates {(0,0)}; \addlegendentry{w/o text};

\AddShotMiouCurve{}{appleteal}{(1,38.33) (2,42.59) (3,44.11) (5,44.58) (10,45.13) (20,45.48)}
\AddShotMiouCurve{}{appleteal}{(1,29.68) (2,36.67) (3,40.21) (5,42.31) (10,43.94) (20,44.01)}[mark=x, mark size=2.2pt, mark options={line cap=round, draw=appleteal, line width=1pt, double=black, double distance=0.55pt}
]

\AddShotMiouCurve{}{applemustard}{(1,36.41) (2,38.05) (3,38.99) (5,39.27) (10,39.85) (20,39.70)}
\AddShotMiouCurve{}{applemustard}{(1,22.91) (2,32.17) (3,37.49) (5,39.95) (10,42.24) (20,42.80)}[mark=x, mark size=2.2pt, mark options={line cap=round, draw=applemustard, line width=1pt, double=black, double distance=0.55pt}
]

\AddShotMiouCurve{}{applepurple}{(1,31.12) (2,36.24) (3,37.73) (5,37.89) (10,37.70) (20,37.48)}
\AddShotMiouCurve{}{applepurple}{(1,28.14) (2,32.81) (3,34.25) (5,34.35) (10,34.27) (20,34.09)}[mark=x, mark size=2.2pt, mark options={line cap=round, draw=applepurple, line width=1pt, double=black, double distance=0.55pt}]

\AddZeroShotLine{}{applebrown}{29.73}

\end{ShotMiouAxis}

%% file: plots/supp/full_visual_support_varying_mem_clip/context.tex
\begin{ShotMiouAxis}[
  width=0.42\linewidth, height=4.5cm,
  title={Context},
  scale only axis,
  xtick={1,2,3,5,10,20},
  xmin=0.6, xmax=20.3,
  ymin=24, ymax=50,
  xlabel={Support images per class},
]
\addplot[color=appleteal, line width=1.2pt] coordinates {(0,0)(1,1)}; \addlegendentry{\ours};
\addplot[color=applemustard, line width=1.2pt] coordinates {(0,0)(1,1)}; \addlegendentry{\knnclip};
\addplot[color=applepurple, line width=1.2pt] coordinates {(0,0)(1,1)}; \addlegendentry{\freeda};
\addplot[color=applebrown, line width=1.2pt] coordinates {(0,0)(1,1)}; \addlegendentry{Zero-shot};
\addplot[only marks, mark=*, mark options={fill=gray, draw=gray}] coordinates {(0,0)}; \addlegendentry{with text};
\addplot[only marks, mark=x, mark options={draw=gray, line width=1pt}] coordinates {(0,0)}; \addlegendentry{w/o text};

\AddShotMiouCurve{}{appleteal}{(1,38.52) (2,42.64) (3,44.20) (5,45.98) (10,48.03) (20,49.05)}
\AddShotMiouCurve{}{appleteal}{(1,30.51) (2,37.83) (3,40.60) (5,43.38) (10,46.52) (20,48.22)}[mark=x, mark size=2.2pt, mark options={line cap=round, draw=appleteal, line width=1pt, double=black, double distance=0.55pt}
]

\AddShotMiouCurve{}{applemustard}{(1,38.70) (2,40.50) (3,41.36) (5,42.67) (10,43.73) (20,44.46)}
\AddShotMiouCurve{}{applemustard}{(1,25.34) (2,35.08) (3,38.57) (5,41.49) (10,44.17) (20,45.97)}[mark=x, mark size=2.2pt, mark options={line cap=round, draw=applemustard, line width=1pt, double=black, double distance=0.55pt}
]

\AddShotMiouCurve{}{applepurple}{(1,35.52) (2,40.28) (3,41.70) (5,43.32) (10,44.24) (20,44.56)}
\AddShotMiouCurve{}{applepurple}{(1,34.67) (2,39.72) (3,41.41) (5,42.78) (10,43.94) (20,44.15)}[mark=x, mark size=2.2pt, mark options={line cap=round, draw=applepurple, line width=1pt, double=black, double distance=0.55pt}]

\AddZeroShotLine{}{applebrown}{34.61}

\end{ShotMiouAxis}

%% file: plots/supp/full_visual_support_varying_mem_dino/voc.tex
\begin{ShotMiouAxis}[
  title={VOC},
  width=0.42\linewidth, height=4.5cm,
  scale only axis,
  xtick={1,2,3,5,10,20},
  xmin=0.6, xmax=20.3,
  ymin=28, ymax=84,
  legend style={
      draw=black!60,
      fill=white,
      rounded corners=5pt,
      font=\tiny,
      anchor=south east,
  },
]
\AddShotMiouCurve{}{applemustard}{
  (1,60.01) (2,60.33) (3,60.63) (5,60.72) (10,60.75) (20,60.66)
}

\AddShotMiouDashedCurve{}{appleteal}{
  (1,65.16) (2,70.25) (3,72.45) (5,73.83) (10,74.74) (20,75.39)
}
\AddShotMiouDashedCurve{}{appleteal}{
  (1,43.30) (2,60.25) (3,68.23) (5,72.63) (10,75.08) (20,75.90)
}[mark=x, mark size=2.2pt, mark options={solid, line cap=round, draw=appleteal, line width=1pt, double=black, double distance=0.55pt}]
\AddShotMiouDashedCurve{}{applemustard}{
  (1,53.45) (2,58.19) (3,60.48) (5,62.22) (10,62.80) (20,63.20)
}

\AddShotMiouCurve{}{appleteal}{
  (1,73.34) (2,78.62) (3,80.37) (5,81.39) (10,81.78) (20,82.08)
}
\AddShotMiouCurve{}{appleteal}{
  (1,50.12) (2,67.82) (3,75.65) (5,79.20) (10,81.63) (20,82.11)
}[mark=x, mark size=2.2pt, mark options={line cap=round, draw=appleteal, line width=1pt, double=black, double distance=0.55pt}]

\AddZeroShotLine{}{applebrown}{31.72}
\AddZeroShotDashedLine{}{applebrown}{31.24}

\addplot[color=appleteal, line width=1.2pt] coordinates {(0,0)(1,1)}; \addlegendentry{\ours};
\addplot[color=applemustard, line width=1.2pt] coordinates {(0,0)(1,1)}; \addlegendentry{\knnclip};
\addplot[color=applebrown, line width=1.2pt] coordinates {(0,0)(1,1)}; \addlegendentry{Zero-shot};
\addplot[only marks, mark=*, mark options={fill=gray, draw=gray}] coordinates {(0,0)}; \addlegendentry{with text};
\addplot[only marks, mark=x, mark options={draw=gray, line width=1pt}] coordinates {(0,0)}; \addlegendentry{w/o text};
\addlegendimage{no markers, solid,  line width=1.2pt}\addlegendentry{SAM}
\addlegendimage{no markers, dashed, line width=1.2pt}\addlegendentry{Patch}
\end{ShotMiouAxis}

%% file: plots/supp/full_visual_support_varying_mem_dino/ade.tex
\begin{ShotMiouAxis}[
  title={ADE},
  width=0.42\linewidth, height=4.5cm,
  scale only axis,
  xtick={1,2,3,5,10,20},
  xmin=0.6, xmax=20.3,
  ymin=23, ymax=48.6,
  legend style={draw=black!60, fill=white, rounded corners=5pt, font=\tiny,  anchor=south east},
]
\AddShotMiouCurve{}{applemustard}{
  (1,32.26) (2,36.27) (3,38.19) (5,39.75) (10,41.67) (20,42.68)
}

\AddShotMiouDashedCurve{}{appleteal}{
  (1,33.25) (2,36.56) (3,38.78) (5,40.16) (10,41.48) (20,42.46)
}
\AddShotMiouDashedCurve{}{appleteal}{
  (1,28.09) (2,32.99) (3,35.83) (5,38.16) (10,40.11) (20,41.15)
}[mark=x, mark size=2.2pt, mark options={solid, line cap=round, draw=appleteal, line width=1pt, double=black, double distance=0.55pt}]
\AddShotMiouDashedCurve{}{applemustard}{
  (1,24.38) (2,28.74) (3,31.37) (5,33.38) (10,35.25) (20,36.44)
}

\AddShotMiouCurve{}{appleteal}{
  (1,37.19) (2,41.53) (3,43.58) (5,45.23) (10,46.92) (20,47.77)
}
\AddShotMiouCurve{}{appleteal}{
  (1,31.83) (2,37.39) (3,40.49) (5,42.75) (10,45.58) (20,46.52)
}[mark=x, mark size=2.2pt, mark options={line cap=round, draw=appleteal, line width=1pt, double=black, double distance=0.55pt}]

\AddZeroShotLine{}{applebrown}{28.03}
\AddZeroShotDashedLine{}{applebrown}{24.33}

\addplot[color=appleteal, line width=1.2pt] coordinates {(0,0)(1,1)}; \addlegendentry{\ours};
\addplot[color=applemustard, line width=1.2pt] coordinates {(0,0)(1,1)}; \addlegendentry{\knnclip};
\addplot[color=applebrown, line width=1.2pt] coordinates {(0,0)(1,1)}; \addlegendentry{Zero-shot};
\addplot[only marks, mark=*, mark options={fill=gray, draw=gray}] coordinates {(0,0)}; \addlegendentry{with text};
\addplot[only marks, mark=x, mark options={draw=gray, line width=1pt}] coordinates {(0,0)}; \addlegendentry{w/o text};
\addlegendimage{no markers, solid,  line width=1.2pt}\addlegendentry{SAM}
\addlegendimage{no markers, dashed, line width=1.2pt}\addlegendentry{Patch}
\end{ShotMiouAxis}

%% file: plots/supp/full_visual_support_varying_mem_dino/city.tex
\begin{ShotMiouAxis}[
  title={City},
  width=0.42\linewidth, height=4.5cm,
  scale only axis,
  xtick={1,2,3,5,10,20},
  xmin=0.6, xmax=20.3,
  ymin=33.5, ymax=62.5,
  legend style={draw=black!60, fill=white, rounded corners=5pt, font=\tiny,  anchor=south east},
]
\AddShotMiouCurve{}{applemustard}{
  (1,53.12) (2,57.38) (3,59.31) (5,59.98) (10,60.21) (20,60.40)
}

\AddShotMiouDashedCurve{}{appleteal}{
  (1,53.11) (2,54.52) (3,55.00) (5,54.70) (10,54.19) (20,54.63)
}
\AddShotMiouDashedCurve{}{appleteal}{
  (1,49.26) (2,52.36) (3,53.31) (5,53.12) (10,53.00) (20,53.47)
}[mark=x, mark size=2.2pt, mark options={solid, line cap=round, draw=appleteal, line width=1pt, double=black, double distance=0.55pt}]
\AddShotMiouDashedCurve{}{applemustard}{
  (1,41.55) (2,46.14) (3,48.38) (5,49.29) (10,49.68) (20,50.01)
}

\AddShotMiouCurve{}{appleteal}{
  (1,59.13) (2,60.88) (3,61.21) (5,61.33) (10,60.87) (20,61.65)
}
\AddShotMiouCurve{}{appleteal}{
  (1,54.26) (2,58.76) (3,60.52) (5,60.53) (10,60.63) (20,61.57)
}[mark=x, mark size=2.2pt, mark options={line cap=round, draw=appleteal, line width=1pt, double=black, double distance=0.55pt}]

\AddZeroShotLine{}{applebrown}{39.47}
\AddZeroShotDashedLine{}{applebrown}{34.69}

\addplot[color=appleteal, line width=1.2pt] coordinates {(0,0)(1,1)}; \addlegendentry{\ours};
\addplot[color=applemustard, line width=1.2pt] coordinates {(0,0)(1,1)}; \addlegendentry{\knnclip};
\addplot[color=applebrown, line width=1.2pt] coordinates {(0,0)(1,1)}; \addlegendentry{Zero-shot};
\addplot[only marks, mark=*, mark options={fill=gray, draw=gray}] coordinates {(0,0)}; \addlegendentry{with text};
\addplot[only marks, mark=x, mark options={draw=gray, line width=1pt}] coordinates {(0,0)}; \addlegendentry{w/o text};
\addlegendimage{no markers, solid,  line width=1.2pt}\addlegendentry{SAM}
\addlegendimage{no markers, dashed, line width=1.2pt}\addlegendentry{Patch}
\end{ShotMiouAxis}

%% file: plots/supp/full_visual_support_varying_mem_dino/stuff.tex
\begin{ShotMiouAxis}[
  title={Stuff},
  width=0.42\linewidth, height=4.5cm,
  scale only axis,
  xtick={1,2,3,5,10,20},
  xmin=0.6, xmax=20.3,
  ymin=22, ymax=45,
  legend style={draw=black!60, fill=white, rounded corners=5pt, font=\tiny, anchor=south east},
]
\AddShotMiouCurve{}{applemustard}{
  (1,30.57) (2,33.41) (3,34.78) (5,36.52) (10,38.01) (20,39.12)
}

\AddShotMiouDashedCurve{}{appleteal}{
  (1,33.24) (2,35.93) (3,37.31) (5,38.26) (10,39.42) (20,40.13)
}
\AddShotMiouDashedCurve{}{appleteal}{
  (1,27.99) (2,32.38) (3,34.48) (5,35.94) (10,37.71) (20,38.68)
}[mark=x, mark size=2.2pt, mark options={solid, line cap=round, draw=appleteal, line width=1pt, double=black, double distance=0.55pt}]
\AddShotMiouDashedCurve{}{applemustard}{
  (1,23.65) (2,28.02) (3,30.08) (5,31.90) (10,33.76) (20,34.90)
}

\AddShotMiouCurve{}{appleteal}{
  (1,36.42) (2,39.31) (3,40.84) (5,41.89) (10,43.17) (20,44.00)
}
\AddShotMiouCurve{}{appleteal}{
  (1,31.28) (2,35.49) (3,37.59) (5,39.09) (10,41.16) (20,42.31)
}[mark=x, mark size=2.2pt, mark options={line cap=round, draw=appleteal, line width=1pt, double=black, double distance=0.55pt}]

\AddZeroShotLine{}{applebrown}{28.28}
\AddZeroShotDashedLine{}{applebrown}{25.71}

\addplot[color=appleteal, line width=1.2pt] coordinates {(0,0)(1,1)}; \addlegendentry{\ours};
\addplot[color=applemustard, line width=1.2pt] coordinates {(0,0)(1,1)}; \addlegendentry{\knnclip};
\addplot[color=applebrown, line width=1.2pt] coordinates {(0,0)(1,1)}; \addlegendentry{Zero-shot};
\addplot[only marks, mark=*, mark options={fill=gray, draw=gray}] coordinates {(0,0)}; \addlegendentry{with text};
\addplot[only marks, mark=x, mark options={draw=gray, line width=1pt}] coordinates {(0,0)}; \addlegendentry{w/o text};
\addlegendimage{no markers, solid,  line width=1.2pt}\addlegendentry{SAM}
\addlegendimage{no markers, dashed, line width=1.2pt}\addlegendentry{Patch}
\end{ShotMiouAxis}

%% file: plots/supp/full_visual_support_varying_mem_dino/object.tex
\begin{ShotMiouAxis}[
  title={Object},
  width=0.42\linewidth, height=4.5cm,
  scale only axis,
  xtick={1,2,3,5,10,20},
  xmin=0.6, xmax=20.3,
  ymin=25, ymax=55,
  legend style={draw=black!60, fill=white, rounded corners=5pt, font=\tiny,  anchor=south east},
]
\AddShotMiouCurve{}{applemustard}{
  (1,43.21) (2,44.32) (3,45.48) (5,46.87) (10,46.81) (20,47.55)
}

\AddShotMiouDashedCurve{}{appleteal}{
  (1,41.97) (2,44.45) (3,45.62) (5,46.00) (10,46.38) (20,46.94)
}
\AddShotMiouDashedCurve{}{appleteal}{
  (1,34.36) (2,41.02) (3,43.65) (5,44.76) (10,45.93) (20,46.47)
}[mark=x, mark size=2.2pt, mark options={solid, line cap=round, draw=appleteal, line width=1pt, double=black, double distance=0.55pt}]
\AddShotMiouDashedCurve{}{applemustard}{
  (1,33.12) (2,36.20) (3,38.55) (5,40.28) (10,40.95) (20,41.65)
}

\AddShotMiouCurve{}{appleteal}{
  (1,47.10) (2,50.33) (3,51.84) (5,52.55) (10,52.83) (20,53.97)
}
\AddShotMiouCurve{}{appleteal}{
  (1,39.30) (2,46.53) (3,49.49) (5,51.14) (10,52.69) (20,53.72)
}[mark=x, mark size=2.2pt, mark options={line cap=round, draw=appleteal, line width=1pt, double=black, double distance=0.55pt}]

\AddZeroShotLine{}{applebrown}{28.96}
\AddZeroShotDashedLine{}{applebrown}{26.45}

\addplot[color=appleteal, line width=1.2pt] coordinates {(0,0)(1,1)}; \addlegendentry{\ours};
\addplot[color=applemustard, line width=1.2pt] coordinates {(0,0)(1,1)}; \addlegendentry{\knnclip};
\addplot[color=applebrown, line width=1.2pt] coordinates {(0,0)(1,1)}; \addlegendentry{Zero-shot};
\addplot[only marks, mark=*, mark options={fill=gray, draw=gray}] coordinates {(0,0)}; \addlegendentry{with text};
\addplot[only marks, mark=x, mark options={draw=gray, line width=1pt}] coordinates {(0,0)}; \addlegendentry{w/o text};
\addlegendimage{no markers, solid,  line width=1.2pt}\addlegendentry{SAM}
\addlegendimage{no markers, dashed, line width=1.2pt}\addlegendentry{Patch}
\end{ShotMiouAxis}

%% file: plots/supp/full_visual_support_varying_mem_dino/context.tex
\begin{ShotMiouAxis}[
  title={Context},
  width=0.42\linewidth, height=4.5cm,
  scale only axis,
  xtick={1,2,3,5,10,20},
  xmin=0.6, xmax=20.3,
  ymin=27.5, ymax=56,
  legend style={draw=black!60, fill=white, rounded corners=5pt, font=\tiny,  anchor=south east},
]
\AddShotMiouCurve{}{applemustard}{
  (1,38.88) (2,41.85) (3,43.27) (5,44.78) (10,46.34) (20,47.38)
}

\AddShotMiouDashedCurve{}{appleteal}{
  (1,39.97) (2,43.65) (3,45.59) (5,47.54) (10,49.15) (20,50.12)
}
\AddShotMiouDashedCurve{}{appleteal}{
  (1,29.21) (2,37.20) (3,40.69) (5,44.26) (10,46.89) (20,48.31)
}[mark=x, mark size=2.2pt, mark options={solid, line cap=round, draw=appleteal, line width=1pt, double=black, double distance=0.55pt}]
\AddShotMiouDashedCurve{}{applemustard}{
  (1,29.62) (2,34.46) (3,36.70) (5,39.19) (10,41.41) (20,42.82)
}

\AddShotMiouCurve{}{appleteal}{
  (1,45.13) (2,49.01) (3,50.98) (5,52.71) (10,54.08) (20,55.08)
}
\AddShotMiouCurve{}{appleteal}{
  (1,34.38) (2,42.30) (3,45.87) (5,49.32) (10,51.91) (20,53.88)
}[mark=x, mark size=2.2pt, mark options={line cap=round, draw=appleteal, line width=1pt, double=black, double distance=0.55pt}]

\AddZeroShotLine{}{applebrown}{31.38}
\AddZeroShotDashedLine{}{applebrown}{29.62}

\addplot[color=appleteal, line width=1.2pt] coordinates {(0,0)(1,1)}; \addlegendentry{\ours};
\addplot[color=applemustard, line width=1.2pt] coordinates {(0,0)(1,1)}; \addlegendentry{\knnclip};
\addplot[color=applebrown, line width=1.2pt] coordinates {(0,0)(1,1)}; \addlegendentry{Zero-shot};
\addplot[only marks, mark=*, mark options={fill=gray, draw=gray}] coordinates {(0,0)}; \addlegendentry{with text};
\addplot[only marks, mark=x, mark options={draw=gray, line width=1pt}] coordinates {(0,0)}; \addlegendentry{w/o text};
\addlegendimage{no markers, solid,  line width=1.2pt}\addlegendentry{SAM}
\addlegendimage{no markers, dashed, line width=1.2pt}\addlegendentry{Patch}
\end{ShotMiouAxis}

%% file: plots/supp/partial_visual_support_mem/voc.tex
\begin{DropFracAxis}[
  title={VOC},
  width=0.42\linewidth, height=4.5cm,
  scale only axis,
  xtick={0.0,0.1,0.3,0.5,0.7,0.9},
  xmin=-0.02, xmax=0.92,
  ymin=1.0, ymax=78.7
]
\addplot[color=appleteal, line width=1.2pt] coordinates {(0,0)(1,1)}; \addlegendentry{\ours};
\addplot[color=applemustard, line width=1.2pt] coordinates {(0,0)(1,1)}; \addlegendentry{\knnclip};
\addplot[color=applepurple, line width=1.2pt] coordinates {(0,0)(1,1)}; \addlegendentry{\freeda};
\addplot[color=applebrown, line width=1.2pt] coordinates {(0,0)(1,1)}; \addlegendentry{Zero-shot};
\addplot[only marks, mark=*, mark options={fill=gray, draw=gray}] coordinates {(0,0)}; \addlegendentry{with text};
\addplot[only marks, mark=o, mark options={draw=gray, line width=1pt}] coordinates {(0,0)}; \addlegendentry{w/o pseudo class vectors};
\addplot[only marks, mark=x, mark options={draw=gray, line width=1pt}] coordinates {(0,0)}; \addlegendentry{w/o text};

\AddMiouDropCurve
  {}{appleteal}
  { (0.00,74.31) (0.10,70.78) (0.30,65.44) (0.50,61.73) (0.70,58.22) (0.90,54.36) };

\AddMiouDropCurve
  {}{appleteal}
  { (0.00,74.31) (0.10,63.14) (0.30,41.74) (0.50,27.88) (0.70,13.74) (0.90,2.95) }[mark=o, mark size=2.2pt, mark options={
  solid, 
  line cap=round, 
  draw=appleteal, 
  line width=1pt, 
  double=black, 
  double distance=0.55pt
}];

\AddMiouDropCurve
  {}{appleteal}
  { (0.00,70.84) (0.10,60.48) (0.30,40.84) (0.50,27.32) (0.70,13.41) (0.90,2.85) }[mark=x, mark size=2.2pt, mark options={
  solid, 
  line cap=round, 
  draw=appleteal, 
  line width=1pt, 
  double=black, 
  double distance=0.55pt
}];

\AddMiouDropCurve
  {}{applemustard}
  { (0.00,68.36) (0.10,66.06) (0.30,61.63) (0.50,58.74) (0.70,54.33) (0.90,51.39) };

\AddMiouDropCurve
  {}{applepurple}
  { (0.00,68.53) (0.10,63.35) (0.30,53.36) (0.50,45.95) (0.70,38.95) (0.90,27.50) };

\AddMiouBaseline{}{applebrown}{54.42}

\end{DropFracAxis}

%% file: plots/supp/partial_visual_support_mem/ade.tex
\begin{DropFracAxis}[
  title={ADE},
  width=0.42\linewidth, height=4.5cm,
  scale only axis,
  xtick={0.0,0.1,0.3,0.5,0.7,0.9},
  xmin=-0.02, xmax=0.92,
  ymin=0.0, ymax=36
]
\addplot[color=appleteal, line width=1.0pt] coordinates {(-5.0,-5.0)(-5.0,-5.0)};
\addlegendentry{\ours};

\addplot[color=applemustard, line width=1.0pt] coordinates {(-5.0,-5.0)(-3.0,-3.0)};
\addlegendentry{\knnclip};

\addplot[color=applepurple, line width=1.0pt] coordinates {(-5.0,-5.0)(-5.0,-5.0)};
\addlegendentry{\freeda};

\addplot[color=applebrown, line width=1.0pt] coordinates {(-5.0,-5.0)(-5.0,-5.0)};
\addlegendentry{Zero-shot};

\addplot[
  only marks,
  mark=*,
  mark options={fill=gray, draw=gray},
] coordinates {(-5.0,-5.0)};
\addlegendentry{with text};

\addplot[
  only marks,
  mark=o,
  mark options={draw=gray, line width=1pt},
] coordinates {(-5.0,-5.0)};
\addlegendentry{w/o pseudo-label loss};

\addplot[
  only marks,
  mark=x,
  mark options={draw=gray, line width=1pt},
] coordinates {(-5.0,-5.0)};
\addlegendentry{w/o text};

\AddMiouDropCurve
  {}{appleteal}
  { (0.00,34.65) (0.10,32.83) (0.30,30.06) (0.50,27.09) (0.70,24.99) (0.90,23.02) };

\AddMiouDropCurve
  {}{appleteal}
  { (0.00,34.65) (0.10,30.48) (0.30,21.84) (0.50,13.30) (0.70,6.46) (0.90,0.86) }[mark=o, mark size=2.2pt, mark options={
  solid, 
  line cap=round, 
  draw=appleteal, 
  line width=1pt, 
  double=black, 
  double distance=0.55pt
}];

\AddMiouDropCurve
  {}{appleteal}
  { (0.00,32.31) (0.10,28.68) (0.30,20.91) (0.50,13.32) (0.70,6.67) (0.90,0.95) }[mark=x, mark size=2.2pt, mark options={
  solid, 
  line cap=round, 
  draw=appleteal, 
  line width=1pt, 
  double=black, 
  double distance=0.55pt
}];

\AddMiouDropCurve
  {}{applemustard}
  { (0.00,30.19) (0.10,29.16) (0.30,25.41) (0.50,22.07) (0.70,19.74) (0.90,17.89) };

\AddMiouDropCurve
  {}{applepurple}
  { (0.00,30.05) (0.10,27.49) (0.30,21.71) (0.50,15.87) (0.70,10.81) (0.90,6.18) };

\AddMiouBaseline{}{applebrown}{22.91}

\end{DropFracAxis}

%% file: plots/supp/partial_visual_support_mem/city.tex
\begin{DropFracAxis}[
  title={City},
  width=0.42\linewidth, height=4.5cm,
  scale only axis,
  xtick={0.0,0.1,0.3,0.5,0.7,0.9},
  xmin=-0.02, xmax=0.92,
  ymin=0.0, ymax=53
]
\addplot[color=appleteal, line width=1.0pt] coordinates {(-5.0,-5.0)(-5.0,-5.0)};
\addlegendentry{\ours};

\addplot[color=applemustard, line width=1.0pt] coordinates {(-5.0,-5.0)(-3.0,-3.0)};
\addlegendentry{\knnclip};

\addplot[color=applepurple, line width=1.0pt] coordinates {(-5.0,-5.0)(-5.0,-5.0)};
\addlegendentry{\freeda};

\addplot[color=applebrown, line width=1.0pt] coordinates {(-5.0,-5.0)(-5.0,-5.0)};
\addlegendentry{Zero-shot};

\addplot[
  only marks,
  mark=*,
  mark options={fill=gray, draw=gray},
] coordinates {(-5.0,-5.0)};
\addlegendentry{with text};

\addplot[
  only marks,
  mark=o,
  mark options={draw=gray, line width=1pt},
] coordinates {(-5.0,-5.0)};
\addlegendentry{w/o pseudo-label loss};

\addplot[
  only marks,
  mark=x,
  mark options={draw=gray, line width=1pt},
] coordinates {(-5.0,-5.0)};
\addlegendentry{w/o text};

\AddMiouDropCurve
  {}{appleteal}
  { (0.00,49.86) (0.10,47.28) (0.30,42.36) (0.50,39.59) (0.70,37.37) (0.90,37.14) };

\AddMiouDropCurve
  {}{appleteal}
  { (0.00,49.86) (0.10,45.82) (0.30,29.67) (0.50,17.44) (0.70,6.78) (0.90,0.85) }[mark=o, mark size=2.2pt, mark options={
  solid, 
  line cap=round, 
  draw=appleteal, 
  line width=1pt, 
  double=black, 
  double distance=0.55pt
}];

\AddMiouDropCurve
  {}{appleteal}
  { (0.00,46.68) (0.10,42.90) (0.30,28.13) (0.50,16.51) (0.70,6.53) (0.90,0.76) }[mark=x, mark size=2.2pt, mark options={
  solid, 
  line cap=round, 
  draw=appleteal, 
  line width=1pt, 
  double=black, 
  double distance=0.55pt
}];

\AddMiouDropCurve
  {}{applemustard}
  { (0.00,48.21) (0.10,45.94) (0.30,40.02) (0.50,33.66) (0.70,29.71) (0.90,27.84) };

\AddMiouDropCurve
  {}{applepurple}
  { (0.00,45.00) (0.10,42.52) (0.30,32.20) (0.50,22.93) (0.70,13.43) (0.90,5.24) };

\AddMiouBaseline{}{applebrown}{37.05}

\end{DropFracAxis}

%% file: plots/supp/partial_visual_support_mem/stuff.tex
\begin{DropFracAxis}[
  title={Stuff},
  width=0.42\linewidth, height=4.5cm,
  scale only axis,
  xtick={0.0,0.1,0.3,0.5,0.7,0.9},
  xmin=-0.02, xmax=0.92,
  ymin=1.0, ymax=36
]
\addplot[color=appleteal, line width=1.2pt] coordinates {(0,0)(1,1)}; \addlegendentry{\ours};
\addplot[color=applemustard, line width=1.2pt] coordinates {(0,0)(1,1)}; \addlegendentry{\knnclip};
\addplot[color=applepurple, line width=1.2pt] coordinates {(0,0)(1,1)}; \addlegendentry{\freeda};
\addplot[color=applebrown, line width=1.2pt] coordinates {(0,0)(1,1)}; \addlegendentry{Zero-shot};
\addplot[only marks, mark=*, mark options={fill=gray, draw=gray}] coordinates {(0,0)}; \addlegendentry{with text};
\addplot[only marks, mark=o, mark options={draw=gray, line width=1pt}] coordinates {(0,0)}; \addlegendentry{w/o pseudo class vectors};
\addplot[only marks, mark=x, mark options={draw=gray, line width=1pt}] coordinates {(0,0)}; \addlegendentry{w/o text};

\AddMiouDropCurve
  {}{appleteal}
  { (0.00,33.53) (0.10,32.41) (0.30,30.78) (0.50,28.95) (0.70,27.73) (0.90,26.90) };

\AddMiouDropCurve
  {}{appleteal}
  { (0.00,33.53) (0.10,29.19) (0.30,21.33) (0.50,13.06) (0.70,6.08) (0.90,1.34) }[mark=o, mark size=2.2pt, mark options={
  solid, 
  line cap=round, 
  draw=appleteal, 
  line width=1pt, 
  double=black, 
  double distance=0.55pt
}];

\AddMiouDropCurve
  {}{appleteal}
  { (0.00,30.54) (0.10,26.64) (0.30,19.66) (0.50,12.22) (0.70,5.88) (0.90,1.35) }[mark=x, mark size=2.2pt, mark options={
  solid, 
  line cap=round, 
  draw=appleteal, 
  line width=1pt, 
  double=black, 
  double distance=0.55pt
}];

\AddMiouDropCurve
  {}{applemustard}
  { (0.00,29.81) (0.10,28.40) (0.30,25.66) (0.50,22.80) (0.70,20.34) (0.90,19.25) };

\AddMiouDropCurve
  {}{applepurple}
  { (0.00,29.41) (0.10,26.42) (0.30,21.01) (0.50,15.47) (0.70,10.71) (0.90,7.03) };

\AddMiouBaseline{}{applebrown}{26.96}
\end{DropFracAxis}

%% file: plots/supp/partial_visual_support_mem/object.tex
\begin{DropFracAxis}[
  title={Object},
  width=0.42\linewidth, height=4.5cm,
  scale only axis,
  xtick={0.0,0.1,0.3,0.5,0.7,0.9},
  xmin=-0.02, xmax=0.92,
  ymin=0.0, ymax=47
]
\addplot[color=appleteal, line width=1.0pt] coordinates {(-5.0,-5.0)(-5.0,-5.0)};
\addlegendentry{\ours};

\addplot[color=applemustard, line width=1.0pt] coordinates {(-5.0,-5.0)(-3.0,-3.0)};
\addlegendentry{\knnclip};

\addplot[color=applepurple, line width=1.0pt] coordinates {(-5.0,-5.0)(-5.0,-5.0)};
\addlegendentry{\freeda};

\addplot[color=applebrown, line width=1.0pt] coordinates {(-5.0,-5.0)(-5.0,-5.0)};
\addlegendentry{Zero-shot};

\addplot[
  only marks,
  mark=*,
  mark options={fill=gray, draw=gray},
] coordinates {(-5.0,-5.0)};
\addlegendentry{with text};

\addplot[
  only marks,
  mark=o,
  mark options={draw=gray, line width=1pt},
] coordinates {(-5.0,-5.0)};
\addlegendentry{w/o pseudo-label loss};

\addplot[
  only marks,
  mark=x,
  mark options={draw=gray, line width=1pt},
] coordinates {(-5.0,-5.0)};
\addlegendentry{w/o text};

\AddMiouDropCurve
  {}{appleteal}
  { (0.00,44.11) (0.10,41.90) (0.30,37.85) (0.50,32.53) (0.70,30.59) (0.90,27.92) };

\AddMiouDropCurve
  {}{appleteal}
  { (0.00,44.11) (0.10,37.48) (0.30,26.13) (0.50,13.92) (0.70,6.83) (0.90,0.97) }[mark=o, mark size=2.2pt, mark options={
  solid, 
  line cap=round, 
  draw=appleteal, 
  line width=1pt, 
  double=black, 
  double distance=0.55pt
}];

\AddMiouDropCurve
  {}{appleteal}
  { (0.00,40.21) (0.10,34.32) (0.30,24.35) (0.50,13.80) (0.70,6.90) (0.90,1.07) }[mark=x, mark size=2.2pt, mark options={
  solid, 
  line cap=round, 
  draw=appleteal, 
  line width=1pt, 
  double=black, 
  double distance=0.55pt
}];

\AddMiouDropCurve
  {}{applemustard}
  { (0.00,38.99) (0.10,37.52) (0.30,33.34) (0.50,29.23) (0.70,27.10) (0.90,23.97) };

\AddMiouDropCurve
  {}{applepurple}
  { (0.00,37.73) (0.10,34.20) (0.30,28.72) (0.50,21.95) (0.70,17.38) (0.90,11.31) };

\AddMiouBaseline{}{applebrown}{29.73}
\end{DropFracAxis}

%% file: plots/supp/partial_visual_support_mem/context.tex
\begin{DropFracAxis}[
  title={Context},
  width=0.42\linewidth, height=4.5cm,
  scale only axis,
  xtick={0.0,0.1,0.3,0.5,0.7,0.9},
  xmin=-0.02, xmax=0.92,
  ymin=1.0, ymax=47
]
\addplot[color=appleteal, line width=1.2pt] coordinates {(0,0)(1,1)}; \addlegendentry{\ours};
\addplot[color=applemustard, line width=1.2pt] coordinates {(0,0)(1,1)}; \addlegendentry{\knnclip};
\addplot[color=applepurple, line width=1.2pt] coordinates {(0,0)(1,1)}; \addlegendentry{\freeda};
\addplot[color=applebrown, line width=1.2pt] coordinates {(0,0)(1,1)}; \addlegendentry{Zero-shot};
\addplot[only marks, mark=*, mark options={fill=gray, draw=gray}] coordinates {(0,0)}; \addlegendentry{with text};
\addplot[only marks, mark=o, mark options={draw=gray, line width=1pt}] coordinates {(0,0)}; \addlegendentry{w/o pseudo class vectors};
\addplot[only marks, mark=x, mark options={draw=gray, line width=1pt}] coordinates {(0,0)}; \addlegendentry{w/o text};

\AddMiouDropCurve
  {}{appleteal}
  { (0.00,44.20) (0.10,42.75) (0.30,40.10) (0.50,37.28) (0.70,35.72) (0.90,34.98) };

\AddMiouDropCurve
  {}{appleteal}
  { (0.00,44.20) (0.10,39.01) (0.30,28.20) (0.50,18.00) (0.70,7.34) (0.90,1.51) }[mark=o, mark size=2.2pt, mark options={
  solid, 
  line cap=round, 
  draw=appleteal, 
  line width=1pt, 
  double=black, 
  double distance=0.55pt
}];

\AddMiouDropCurve
  {}{appleteal}
  { (0.00,40.60) (0.10,35.88) (0.30,25.92) (0.50,16.95) (0.70,7.14) (0.90,1.54) }[mark=x, mark size=2.2pt, mark options={
  solid, 
  line cap=round, 
  draw=appleteal, 
  line width=1pt, 
  double=black, 
  double distance=0.55pt
}];

\AddMiouDropCurve
  {}{applemustard}
  { (0.00,41.36) (0.10,39.80) (0.30,37.18) (0.50,33.71) (0.70,30.73) (0.90,29.66) };

\AddMiouDropCurve
  {}{applepurple}
  { (0.00,41.70) (0.10,38.01) (0.30,29.17) (0.50,23.08) (0.70,17.49) (0.90,13.29) };

\AddMiouBaseline{}{applebrown}{34.61}
\end{DropFracAxis}

%% file: plots/supp/partial_textual_support_mem/voc.tex
\begin{DropFracTextAxis}[
  title={VOC},
  width=0.42\linewidth, height=4.5cm,
  scale only axis,
  xtick={0.0,0.1,0.3,0.5,0.7,0.9},
  xmin=-0.02, xmax=0.92,
  ymin=33, ymax=70.5,
]

\addplot[color=appleteal, line width=1.2pt] coordinates {(0,0)(1,1)};
\addlegendentry{\ours};

\addplot[color=applemustard, line width=1.2pt] coordinates {(0,0)(1,1)};
\addlegendentry{\knnclip};

\addplot[color=applepurple, line width=1.2pt] coordinates {(0,0)(1,1)};
\addlegendentry{\freeda};

\addplot[color=applebrown, line width=1.2pt] coordinates {(0,0)(1,1)};
\addlegendentry{Zero-shot};

\addplot[only marks, mark=*, mark options={fill=gray, draw=gray}] coordinates {(0,0)};
\addlegendentry{with text};

\addplot[only marks, mark=x, mark options={draw=gray, line width=1pt}] coordinates {(0,0)};
\addlegendentry{w/o text};

\AddMiouDropCurve{}{appleteal}{
  (0.00,68.61)
  (0.10,67.34)
  (0.30,66.00)
  (0.50,64.61)
  (0.70,62.50)
  (0.90,60.44)
};

\AddMiouDropCurve{}{applemustard}{
  (0.00,66.40)
  (0.10,62.06)
  (0.30,55.37)
  (0.50,48.75)
  (0.70,43.66)
  (0.90,37.76)
};

\AddMiouDropCurve{}{applepurple}{
  (0.00,60.50)
  (0.10,60.07)
  (0.30,58.97)
  (0.50,58.08)
  (0.70,57.32)
  (0.90,56.32)
};

\AddMiouDropCurve{}{appleteal}{
  (0.0,54.10) (0.1,54.10) (0.3,54.10) (0.5,54.10) (0.7,54.10) (0.9,54.10)
}[mark=x, mark size=2.2pt,
  mark options={solid, line cap=round, draw=appleteal,
                line width=1pt, double=black, double distance=0.55pt}];

\AddMiouDropCurve{}{applemustard}{
  (0.0,35.10) (0.1,35.10) (0.3,35.10) (0.5,35.10) (0.7,35.10) (0.9,35.10)
}[mark=x, mark size=2.2pt,
  mark options={solid, line cap=round, draw=applemustard,
                line width=1pt, double=black, double distance=0.55pt}];

\AddMiouDropCurve{}{applepurple}{
  (0.0,55.91) (0.1,55.91) (0.3,55.91) (0.5,55.91) (0.7,55.91) (0.9,55.91)
}[mark=x, mark size=2.2pt,
  mark options={solid, line cap=round, draw=applepurple,
                line width=1pt, double=black, double distance=0.55pt}];

\end{DropFracTextAxis}

%% file: plots/supp/partial_textual_support_mem/ade.tex
\begin{DropFracTextAxis}[
  title={ADE},
  width=0.42\linewidth, height=4.5cm,
  scale only axis,
  xtick={0.0,0.1,0.3,0.5,0.7,0.9},
  xmin=-0.02, xmax=0.92,
  ymin=21.5, ymax=29.5,
]

\addplot[color=appleteal, line width=1.2pt] coordinates {(0,0)(1,1)}; \addlegendentry{\ours};
\addplot[color=applemustard, line width=1.2pt] coordinates {(0,0)(1,1)}; \addlegendentry{\knnclip};
\addplot[color=applepurple, line width=1.2pt] coordinates {(0,0)(1,1)}; \addlegendentry{\freeda};
\addplot[color=applebrown, line width=1.2pt] coordinates {(0,0)(1,1)}; \addlegendentry{Zero-shot};
\addplot[only marks, mark=*, mark options={fill=gray, draw=gray}] coordinates {(0,0)}; \addlegendentry{with text};
\addplot[only marks, mark=x, mark options={draw=gray, line width=1pt}] coordinates {(0,0)}; \addlegendentry{w/o text};

\AddMiouDropCurve{}{appleteal}{
  (0.00,29.22)
  (0.10,28.96)
  (0.30,28.24)
  (0.50,27.48)
  (0.70,27.00)
  (0.90,26.72)
};

\AddMiouDropCurve{}{applemustard}{
  (0.00,27.02)
  (0.10,26.66)
  (0.30,25.59)
  (0.50,24.21)
  (0.70,22.87)
  (0.90,22.45)
};

\AddMiouDropCurve{}{applepurple}{
  (0.00,25.95)
  (0.10,25.79)
  (0.30,25.53)
  (0.50,25.28)
  (0.70,25.14)
  (0.90,25.06)
};

\AddMiouDropCurve{}{appleteal}{
  (0.0,24.34) (0.1,24.34) (0.3,24.34) (0.5,24.34) (0.7,24.34) (0.9,24.34)
}[mark=x, mark size=2.2pt,
  mark options={solid, line cap=round, draw=appleteal,
                line width=1pt, double=black, double distance=0.55pt}];

\AddMiouDropCurve{}{applemustard}{
  (0.0,21.84) (0.1,21.84) (0.3,21.84) (0.5,21.84) (0.7,21.84) (0.9,21.84)
}[mark=x, mark size=2.2pt,
  mark options={solid, line cap=round, draw=applemustard,
                line width=1pt, double=black, double distance=0.55pt}];

\AddMiouDropCurve{}{applepurple}{
  (0.0,24.83) (0.1,24.83) (0.3,24.83) (0.5,24.83) (0.7,24.83) (0.9,24.83)
}[mark=x, mark size=2.2pt,
  mark options={solid, line cap=round, draw=applepurple,
                line width=1pt, double=black, double distance=0.55pt}];

\end{DropFracTextAxis}

%% file: plots/supp/partial_textual_support_mem/city.tex
\begin{DropFracTextAxis}[
  title={City},
  width=0.42\linewidth, height=4.5cm,
  scale only axis,
  xtick={0.0,0.1,0.3,0.5,0.7,0.9},
  xmin=-0.02, xmax=0.92,
  ymin=38, ymax=46.2,
]

\addplot[color=appleteal, line width=1.2pt] coordinates {(0,0)(1,1)}; \addlegendentry{\ours};
\addplot[color=applemustard, line width=1.2pt] coordinates {(0,0)(1,1)}; \addlegendentry{\knnclip};
\addplot[color=applepurple, line width=1.2pt] coordinates {(0,0)(1,1)}; \addlegendentry{\freeda};
\addplot[color=applebrown, line width=1.2pt] coordinates {(0,0)(1,1)}; \addlegendentry{Zero-shot};
\addplot[only marks, mark=*, mark options={fill=gray, draw=gray}] coordinates {(0,0)}; \addlegendentry{with text};
\addplot[only marks, mark=x, mark options={draw=gray, line width=1pt}] coordinates {(0,0)}; \addlegendentry{w/o text};

\AddMiouDropCurve{}{appleteal}{
  (0.00,45.79)
  (0.10,45.59)
  (0.30,44.81)
  (0.50,44.03)
  (0.70,42.99)
  (0.90,41.83)
};

\AddMiouDropCurve{}{applemustard}{
  (0.00,45.67)
  (0.10,44.86)
  (0.30,42.12)
  (0.50,40.42)
  (0.70,39.32)
  (0.90,38.69)
};

\AddMiouDropCurve{}{applepurple}{
  (0.00,39.92)
  (0.10,39.76)
  (0.30,40.04)
  (0.50,40.61)
  (0.70,41.00)
  (0.90,41.46)
};

\AddMiouDropCurve{}{appleteal}{
  (0.0,41.78) (0.1,41.78) (0.3,41.78) (0.5,41.78) (0.7,41.78) (0.9,41.78)
}[mark=x, mark size=2.2pt,
  mark options={solid, line cap=round, draw=appleteal,
                line width=1pt, double=black, double distance=0.55pt}];

\AddMiouDropCurve{}{applemustard}{
  (0.0,39.06) (0.1,39.06) (0.3,39.06) (0.5,39.06) (0.7,39.06) (0.9,39.06)
}[mark=x, mark size=2.2pt,
  mark options={solid, line cap=round, draw=applemustard,
                line width=1pt, double=black, double distance=0.55pt}];

\AddMiouDropCurve{}{applepurple}{
  (0.0,42.29) (0.1,42.29) (0.3,42.29) (0.5,42.29) (0.7,42.29) (0.9,42.29)
}[mark=x, mark size=2.2pt,
  mark options={solid, line cap=round, draw=applepurple,
                line width=1pt, double=black, double distance=0.55pt}];

\end{DropFracTextAxis}

%% file: plots/supp/partial_textual_support_mem/stuff.tex
\begin{DropFracTextAxis}[
  title={Stuff},
  width=0.42\linewidth, height=4.5cm,
  scale only axis,
  xtick={0.0,0.1,0.3,0.5,0.7,0.9},
  xmin=-0.02, xmax=0.92,
  ymin=19, ymax=29.5,
]

\addplot[color=appleteal, line width=1.2pt] coordinates {(0,0)(1,1)}; \addlegendentry{\ours};
\addplot[color=applemustard, line width=1.2pt] coordinates {(0,0)(1,1)}; \addlegendentry{\knnclip};
\addplot[color=applepurple, line width=1.2pt] coordinates {(0,0)(1,1)}; \addlegendentry{\freeda};
\addplot[color=applebrown, line width=1.2pt] coordinates {(0,0)(1,1)}; \addlegendentry{Zero-shot};
\addplot[only marks, mark=*, mark options={fill=gray, draw=gray}] coordinates {(0,0)}; \addlegendentry{with text};
\addplot[only marks, mark=x, mark options={draw=gray, line width=1pt}] coordinates {(0,0)}; \addlegendentry{w/o text};

\AddMiouDropCurve{}{appleteal}{
  (0.00,29.10)
  (0.10,28.19)
  (0.30,27.46)
  (0.50,26.68)
  (0.70,26.01)
  (0.90,25.39)
};

\AddMiouDropCurve{}{applemustard}{
  (0.00,26.84)
  (0.10,25.97)
  (0.30,24.42)
  (0.50,22.39)
  (0.70,21.20)
  (0.90,20.57)
};

\AddMiouDropCurve{}{applepurple}{
  (0.00,24.74)
  (0.10,24.48)
  (0.30,24.22)
  (0.50,24.01)
  (0.70,23.96)
  (0.90,24.07)
};

\AddMiouDropCurve{}{appleteal}{
  (0.0,24.23) (0.1,24.23) (0.3,24.23) (0.5,24.23) (0.7,24.23) (0.9,24.23)
}[mark=x, mark size=2.2pt,
  mark options={solid, line cap=round, draw=appleteal,
                line width=1pt, double=black, double distance=0.55pt}];

\AddMiouDropCurve{}{applemustard}{
  (0.0,20.56) (0.1,20.56) (0.3,20.56) (0.5,20.56) (0.7,20.56) (0.9,20.56)
}[mark=x, mark size=2.2pt,
  mark options={solid, line cap=round, draw=applemustard,
                line width=1pt, double=black, double distance=0.55pt}];

\AddMiouDropCurve{}{applepurple}{
  (0.0,24.54) (0.1,24.54) (0.3,24.54) (0.5,24.54) (0.7,24.54) (0.9,24.54)
}[mark=x, mark size=2.2pt,
  mark options={solid, line cap=round, draw=applepurple,
                line width=1pt, double=black, double distance=0.55pt}];

\end{DropFracTextAxis}

%% file: plots/supp/partial_textual_support_mem/object.tex
\begin{DropFracTextAxis}[
  title={Object},
  width=0.42\linewidth, height=4.5cm,
  scale only axis,
  xtick={0.0,0.1,0.3,0.5,0.7,0.9},
  xmin=-0.02, xmax=0.92,
  ymin=21.2, ymax=39.5,
]

\addplot[color=appleteal, line width=1.2pt] coordinates {(0,0)(1,1)}; \addlegendentry{\ours};
\addplot[color=applemustard, line width=1.2pt] coordinates {(0,0)(1,1)}; \addlegendentry{\knnclip};
\addplot[color=applepurple, line width=1.2pt] coordinates {(0,0)(1,1)}; \addlegendentry{\freeda};
\addplot[color=applebrown, line width=1.2pt] coordinates {(0,0)(1,1)}; \addlegendentry{Zero-shot};
\addplot[only marks, mark=*, mark options={fill=gray, draw=gray}] coordinates {(0,0)}; \addlegendentry{with text};
\addplot[only marks, mark=x, mark options={draw=gray, line width=1pt}] coordinates {(0,0)}; \addlegendentry{w/o text};

\AddMiouDropCurve{}{appleteal}{
  (0.00,38.44)
  (0.10,36.99)
  (0.30,35.73)
  (0.50,34.79)
  (0.70,33.32)
  (0.90,31.96)
};

\AddMiouDropCurve{}{applemustard}{
  (0.00,37.37)
  (0.10,34.96)
  (0.30,31.00)
  (0.50,27.74)
  (0.70,25.56)
  (0.90,23.29)
};

\AddMiouDropCurve{}{applepurple}{
  (0.00,31.12)
  (0.10,30.61)
  (0.30,30.03)
  (0.50,29.10)
  (0.70,28.84)
  (0.90,28.05)
};

\AddMiouDropCurve{}{appleteal}{
  (0.0,29.68) (0.1,29.68) (0.3,29.68) (0.5,29.68) (0.7,29.68) (0.9,29.68)
}[mark=x, mark size=2.2pt,
  mark options={solid, line cap=round, draw=appleteal,
                line width=1pt, double=black, double distance=0.55pt}];

\AddMiouDropCurve{}{applemustard}{
  (0.0,22.91) (0.1,22.91) (0.3,22.91) (0.5,22.91) (0.7,22.91) (0.9,22.91)
}[mark=x, mark size=2.2pt,
  mark options={solid, line cap=round, draw=applemustard,
                line width=1pt, double=black, double distance=0.55pt}];

\AddMiouDropCurve{}{applepurple}{
  (0.0,28.14) (0.1,28.14) (0.3,28.14) (0.5,28.14) (0.7,28.14) (0.9,28.14)
}[mark=x, mark size=2.2pt,
  mark options={solid, line cap=round, draw=applepurple,
                line width=1pt, double=black, double distance=0.55pt}];

\end{DropFracTextAxis}

%% file: plots/supp/partial_textual_support_mem/context.tex
\begin{DropFracTextAxis}[
  title={Context},
  width=0.42\linewidth, height=4.5cm,
  scale only axis,
  xtick={0.0,0.1,0.3,0.5,0.7,0.9},
  xmin=-0.02, xmax=0.92,
  ymin=24, ymax=40,
]

\addplot[color=appleteal, line width=1.2pt] coordinates {(0,0)(1,1)}; \addlegendentry{\ours};
\addplot[color=applemustard, line width=1.2pt] coordinates {(0,0)(1,1)}; \addlegendentry{\knnclip};
\addplot[color=applepurple, line width=1.2pt] coordinates {(0,0)(1,1)}; \addlegendentry{\freeda};
\addplot[color=applebrown, line width=1.2pt] coordinates {(0,0)(1,1)}; \addlegendentry{Zero-shot};
\addplot[only marks, mark=*, mark options={fill=gray, draw=gray}] coordinates {(0,0)}; \addlegendentry{with text};
\addplot[only marks, mark=x, mark options={draw=gray, line width=1pt}] coordinates {(0,0)}; \addlegendentry{w/o text};

\AddMiouDropCurve{}{appleteal}{
  (0.00,39.18)
  (0.10,38.61)
  (0.30,37.48)
  (0.50,36.49)
  (0.70,35.52)
  (0.90,34.50)
};

\AddMiouDropCurve{}{applemustard}{
  (0.00,38.64)
  (0.10,37.39)
  (0.30,33.71)
  (0.50,30.77)
  (0.70,28.21)
  (0.90,26.83)
};

\AddMiouDropCurve{}{applepurple}{
  (0.00,36.03)
  (0.10,35.94)
  (0.30,35.34)
  (0.50,35.23)
  (0.70,35.37)
  (0.90,35.22)
};

\AddMiouDropCurve{}{appleteal}{
  (0.0,30.51) (0.1,30.51) (0.3,30.51) (0.5,30.51) (0.7,30.51) (0.9,30.51)
}[mark=x, mark size=2.2pt,
  mark options={solid, line cap=round, draw=appleteal,
                line width=1pt, double=black, double distance=0.55pt}];

\AddMiouDropCurve{}{applemustard}{
  (0.0,25.34) (0.1,25.34) (0.3,25.34) (0.5,25.34) (0.7,25.34) (0.9,25.34)
}[mark=x, mark size=2.2pt,
  mark options={solid, line cap=round, draw=applemustard,
                line width=1pt, double=black, double distance=0.55pt}];

\AddMiouDropCurve{}{applepurple}{
  (0.0,34.67) (0.1,34.67) (0.3,34.67) (0.5,34.67) (0.7,34.67) (0.9,34.67)
}[mark=x, mark size=2.2pt,
  mark options={solid, line cap=round, draw=applepurple,
                line width=1pt, double=black, double distance=0.55pt}];

\end{DropFracTextAxis}